\documentclass[12pt]{article}
\usepackage{amsmath}
\usepackage{graphicx}
\usepackage{enumerate}
\usepackage{xcolor}
\usepackage{natbib}
\usepackage{multirow}
\usepackage{multicol}
\usepackage{booktabs}
\usepackage{url} 

\newcommand{\blind}{0}
\def\spacingset#1{\renewcommand{\baselinestretch}%
{#1}\small\normalsize} \spacingset{1}

\usepackage{amsmath,amsfonts,bm}

\def\1{\bm{1}}

\def\mL{{\bm{L}}}

\DeclareMathAlphabet{\mathsfit}{\encodingdefault}{\sfdefault}{m}{sl}
\SetMathAlphabet{\mathsfit}{bold}{\encodingdefault}{\sfdefault}{bx}{n}

\newcommand{\E}{\mathbb{E}}

\newcommand{\R}{\mathbb{R}}

\newcommand{\diag}{\mathrm{diag}}

\usepackage{amsfonts}
\usepackage{amsmath,amssymb,amsthm}
\usepackage{physics}

\def \P {\mathbb{P}}

\def \I {\mathbb{I}}
\def \bA {\mathbf{A}}

\def \E {\mathbb{E}}
\def \F {\mathrm{F}}
\def \R {\mathbb{R}}

\def \v {\mathbf{v}}

\def \beps{{\boldsymbol{\epsilon}}}

\def \bSigma{{\boldsymbol{\Sigma}}}

\def\calW{{\cal W}}

\def\Bb{\mathbf{B}}

\def\ZERO{\mathbf{O}}
\def\Bb{\mathbf{B}}
\def\Pb{\mathbf{P}}
\def\Vbhat{\mathbf{\widehat V}}
\def\Vb{\mathbf{V}}

\def\Ubhat{\mathbf{\widehat U}}

\newcommand{\CC}{\mathbf{C}}
\newcommand{\DD}{\mathbf{D}}
\newcommand{\bc}{\bm{c}}
\newcommand{\Ub}{\mathbf{U}}

\newcommand{\Ib}{\mathbf{I}}
\newcommand{\be}{\bm{e}}

\newcommand{\bLambda}{\bm{\Lambda}}

\newcommand{\zero}{\bm{0}}

\def\Hb{\mathbf{H}}

\def\mL{\mathcal{L}}

\def\PMIbb{\mathbb{PMI}}
\def\PMIbbhat{\widehat{\PMIbb}}

\def\bLambdahat{\widehat{\bLambda}}
\def\CLAIME{\textnormal{CLAIME}}
\def\CL{\textnormal{CL}}
\def\PCA{\textnormal{Con}}

\newtheorem{thm}{Theorem}[section]
\newtheorem{prop}{Proposition}[section]
\newtheorem{lem}{Lemma}[section]
\newtheorem{cor}{Corollary}[section]
\theoremstyle{definition}

\newtheorem{rem}{Remark}[section] 
\newtheorem{asm}{Assumption}[section]
\numberwithin{equation}{section}

\renewcommand{\top}{{\scriptscriptstyle \sf T}}

\makeatletter

\usepackage{tikz}
\usetikzlibrary{arrows.meta, positioning, calc}
\usetikzlibrary{shapes.geometric}
\allowdisplaybreaks[4]

\date{}

\begin{document}

\if0\blind
\title{Contrastive Learning on Multimodal Analysis of Electronic Health Records}
{
\author{ 
Tianxi Cai$^{1,2\star}$, Feiqing Huang$^{1\star}$, Ryumei Nakada$^{2,3\star}$, \\ Linjun Zhang$^{3\star}$,  Doudou Zhou$^{4\star}$\\
\footnotesize{$^1$Department of Biostatistics, Harvard T.H. Chan School of Public Health, Boston, MA}\\
\footnotesize{$^2$Department of Biomedical Informatics, Harvard Medical School, Boston, MA}\\
\footnotesize{$^3$Department of Statistics, Rutgers University, Piscataway, NJ}\\
\footnotesize{$^4$Department of Statistics and Data Science, National University of Singapore, Singapore}\\
\footnotesize{$^\star$alphabetical order}
}
  \maketitle
}
\fi

\if1\blind
{
  \bigskip
  \bigskip
  \bigskip
  \begin{center}
    {\LARGE\bf Contrastive Learning on Multimodal Analysis of Electronic Health Records}
\end{center}
  \medskip
} \fi

\begin{abstract}
Electronic health record (EHR) systems capture a wealth of multimodal clinical data, encompassing both structured clinical codes and unstructured clinical notes. Yet, many EHR-focused studies have traditionally examined these modalities in isolation or combined them using simplistic methods, overlooking the intrinsic synergy between them. In reality, these modalities are deeply interconnected, each containing clinically relevant and complementary information that, when integrated effectively, can provide a more comprehensive understanding of patient health. Despite the success of multimodal contrastive learning in vision-language applications, its potential remains under-explored in multimodal EHR, particularly in terms of theoretical understanding. To support statistical analysis of multimodal EHR data, we propose a multimodal feature embedding generative model and design a multimodal contrastive loss to learn EHR feature representations. Our theoretical analysis demonstrates the effectiveness of multimodal learning over single-modality learning and connects the solution of the loss function to the singular value decomposition of a pointwise mutual information matrix. This connection leads to a privacy-preserving algorithm tailored for multimodal EHR representation learning. Simulation studies show that the proposed algorithm performs well under a variety of configurations. We further validate its clinical utility using real-world EHR data.
\end{abstract}


\newpage
\spacingset{1.9} 

\section{Introduction}

The growing accessibility of Electronic Health Record (EHR) data presents numerous opportunities for clinical research, ranging from patient profiling to predicting medical events. However, the complexity increases with the multimodal nature of EHR data, which encompasses diverse data from patient demographics and genetic information to unstructured textual data like clinical notes, as well as structured data such as diagnostic and procedure codes, medication orders, and lab results.

A key challenge in EHR-focused research lies in effectively merging these different data types and ensuring that their clinical aspects are meaningfully and coherently represented. Research has shown the benefits of integrating structured and unstructured data for tasks like automated clinical code assignment \citep{scheurwegs2016data}, managing chronic diseases \citep{sheikhalishahi2019natural}, and pharmacovigilance \citep{stang2010advancing}. While these different modalities serve as complementary data sources, there is significant overlap and correlation among these modalities \citep{qiao2019mnn}. Joint representation of both structured and narrative data into a more manageable low-dimensional space where similar features are grouped closely can significantly improve the utility of both data types. This representation learning technique has gained popularity for its ability to capture and represent the intricate relationships among various EHR features.

Despite extensive research on EHR feature representation, most studies have focused on datasets with a single modality, including structured \citep{choi2016multi,kartchner2017code2vec,hong2021clinical,zhou2022multiview} and unstructured data \citep{de2014medical,choi2016learning,beam2019clinical,alsentzer2019publicly,huang-etal-2020-clinical,lehman2023clinical}. For example, \cite{alsentzer2019publicly} adapted BERT \citep{devlin2018bert} to the clinical domain by pretraining on MIMIC-III notes \citep{johnson2016mimic}, a process that was highly resource-intensive. \cite{de2014medical} aligned free-text with the UMLS Concept Unique Identifier (CUI) space \citep{mcinnes2007using} and applied skip-gram \citep{mikolov2013efficient} to learn embeddings, while \cite{choi2016learning}, \cite{beam2019clinical, hong2021clinical, levy2014neural,arora2016latent} used singular value decomposition (SVD) on pointwise mutual information (PMI) matrices.

Building on these unimodal advances, recent work has explored multimodal EHR representations that integrate both structured and unstructured information \citep{khadanga-etal-2019-using,bardak2021improving,Gan2023}. Joint representation is preferable to treating all data as a single modality because, although structured codes and unstructured text encode overlapping clinical concepts, they possess fundamentally different statistical forms. Ignoring these cross-modality differences can obscure modality-specific information, discard complementary signal, and amplify systematic bias (see further discussion in Section~\ref{sec:theory:comparison}). By contrast, multimodal integration can exploit the complementary strengths of distinct modalities, yielding representations that are not only richer but also less biased than those obtained from unimodal analyses. For instance, information that is noisy or missing in one modality may be corroborated by the other, thereby stabilizing estimation and enhancing generalization. In this paper, we develop a theoretical framework (Section~\ref{sec: theory}) and provide empirical evidence (Sections~\ref{simulation} and \ref{real}) to demonstrate that multimodal analysis addresses these challenges and produces more reliable and interpretable representations of EHR data.

To address these limitations, \citet{liu2022multimodal} proposed a multimodal pre-trained language model that incorporates a cross-attention mechanism to enrich EHR representations across structured and unstructured domains. More recently, multimodal contrastive learning strategies have also been explored \citep{li2022multi,yin2023decision,wang2023hierarchical}. Inspired by the success of vision–language models such as CLIP \citep{radford2021learning}, these approaches aim to learn unified representations of heterogeneous data modalities. From a statistical perspective, contrastive learning can be interpreted through the lens of noise-contrastive estimation \citep{gutmann2010noise}, in which a parametric score function distinguishes positive pairs drawn from the joint distribution from negative pairs drawn from the product of marginals. The popular InfoNCE loss \citep{oord2018representation} maximizes a lower bound on mutual information and is closely related to low-rank approximations of the PMI matrix \citep{levy2014neural}. Despite their promise, these methods remain limited by their reliance on deep neural networks, which are often criticized for their ``black-box'' nature. Practical challenges include the absence of rigorous theoretical guarantees, high computational cost, and privacy concerns due to the need for patient-level data. Together, these issues constrain the applicability of existing multimodal contrastive learning frameworks in the context of EHR data.

While some theoretical analyses of multimodal learning exist, their applicability to sparse discrete EHR data has been limited. Groundbreaking studies like \cite{huang2021makes} have illustrated the benefits of multimodal learning, showing that learning across multiple modalities can reduce population risk compared to single-modality methods. Recently, \citet{chen2023understanding} theoretically proved the zero-shot transfer ability of CLIP. Furthermore, \cite{nakada2023understanding} explored multimodal contrastive learning's performance under a spiked covariance model. However, these studies do not directly apply to the unique discrete feature structure of EHR data, leaving an unaddressed theoretical gap in understanding multimodal contrastive learning's application in healthcare. Bridging this gap is vital, as it lays a foundation for multimodal contrastive learning's development and implementation in healthcare, maximizing its potential to improve  care and drive medical research forward.

Complementary lines of research have attempted to address related challenges from different angles. For instance, exponential family principal component analysis
 (PCA) \citep{collins2001generalization} and nonlinear PCA \citep{hinton2006reducing}  extend classical PCA to non-Gaussian and nonlinear settings, while kernel PCA (KPCA) \citep{gupta2019improving} provides additional flexibility. Yet, handling multimodal, sparse, and privacy-sensitive data remains an open challenge. In parallel, the statistical literature on integrative dimension reduction, such as JIVE \citep{lock2013joint}, AJIVE \citep{feng2018angle}, and other multi-view factor models \citep[e.g.,]{zhou2015group,yang2016non,yi2023hierarchical}, provides tools for modeling shared and modality-specific structures. Our framework shares this goal but is tailored to discrete EHR features and prioritizes privacy-preserving learning via co-occurrence data.

To overcome these limitations, we introduce the \textbf{C}ontrastive \textbf{L}earning \textbf{A}lgorithm for \textbf{I}ntegrated \textbf{M}ultimodal \textbf{E}lectronic health records (CLAIME). Our findings confirm that CLAIME is not only effective for deriving multimodal EHR feature representations but also respects privacy by requiring only aggregated data. Additionally, we propose a novel multimodal feature embedding generative model \eqref{model} in Section~\ref{model}, designed to enhance statistical analysis of multimodal EHR data. This model is notable for its interpretability and accurate portrayal of EHR data generation. It distinguishes itself from earlier word vector generative models \citep{arora2016latent,arora2018linear,junwei,xu2023inference} by (1) enabling the integration of multimodal EHR features, and (2) allowing patient heterogeneity by incorporating error terms specific to patients, thus increasing the model's robustness. Within this generative framework, we validate the consistency of the CLAIME algorithm and clarify the relationship between multimodal feature embeddings and a multimodal PMI matrix. The proposed algorithm is also privacy-preserving since it only requires summary-level data, opening doors for collaboration across multiple institutions. Our research also fills a theoretical void in the analysis of multimodal contrastive learning for EHR data.

The rest of the paper is organized as follows. Section~\ref{sec: method} introduces the proposed method, and Section~\ref{sec: theory} presents its theoretical properties. Sections~\ref{simulation} and~\ref{real} provide simulation and real-world evaluations, respectively. Section~\ref{discussion} concludes with a discussion.

\section{Method}\label{sec: method}

{  We begin by defining the notation.}
For any matrix $\bA$, let $\|\bA\|$, $\|\bA\|_{\max}$, and $\|\bA\|_\F$ denote its operator norm, entrywise maximum norm, and Frobenius norm, respectively. Let $\Pb_p(\bA)$ be the top-$p$ right singular vectors of $\bA$, chosen arbitrarily if non-unique, and let $s_j(\bA)$ be its $j$-th largest singular value. Define $\mathcal{O}_{d,p}$ ($d \geq p$) as the set of $d \times p$ orthonormal matrices. For sequences $\{a_n\}, \{b_n\} > 0$, write $a_n \lesssim b_n$, $a_n = O(b_n)$, $b_n \gtrsim a_n$, or $b_n = \Omega(a_n)$ if $a_n \leq C b_n$ for some $C > 0$ and all $n$; write $a_n \ll b_n$ or $a_n = o(b_n)$ if $a_n / b_n \to 0$. Let $[I] = \{1, 2, \dots, I\}$ for any positive integer $I$, and write $a \vee b$ and $a \wedge b$ for $\max(a, b)$ and $\min(a, b)$, respectively. Denote by $\be_j$ the $j$-th unit vector in $\R^d$.

\subsection{Model Assumptions}
\label{sec:model}

\begin{figure}[h]
\centering
\begingroup
\def\twolines#1#2{%
  \begin{tabular}[c]{@{}c@{}}
  #1\\[-1.6pt]
  #2
  \end{tabular}%
}
\begin{tikzpicture}[
    code/.style={
        draw,
        thick,
        rounded corners,
        fill=teal!10,
        minimum width=2.8cm,
        inner xsep=4pt,
        inner ysep=2.2pt,
        align=center
    },
    cui/.style={
        draw,
        thick,
        rounded corners,
        fill=purple!10,
        minimum width=2.8cm,
        inner xsep=4pt,
        inner ysep=2.2pt,
        align=center
    },
    latent/.style={
        draw,
        thick,
        fill=gray!10,
        minimum width=3.8cm,
        inner xsep=5pt,
        inner ysep=2.2pt,
        align=center
    },
    note/.style={
        draw,
        thick,
        fill=yellow!10,
        minimum width=7.2cm,
        inner xsep=6pt,
        inner ysep=4pt,
        align=center
    },
    arrow/.style={-Stealth, thick, shorten >=1pt, shorten <=1pt},
    dashedarrow/.style={-Stealth, thick, dashed, shorten >=1pt, shorten <=1pt},
    every node/.style={font=\scriptsize\linespread{0.93}\selectfont}
]

\node[latent] (latent) at (0,0) {\shortstack{Latent Clinical State $\bc_i$}};

\node[code] (code1) at (-3.2, 1.10)
{\twolines{PheCode 714.1}{(Rheumatoid arthritis)}};

\node[code] (code2) at (0, 1.10)
{\twolines{RxNorm: 6851}{(Methotrexate)}};

\node[code] (code3) at (3.0, 1.10)
{\twolines{CCS: 152}{(Arthroplasty knee)}};

\node[cui] (cui1) at (-3.2, -1.10)
{\twolines{CUI: C0457086}{(Morning stiffness)}};

\node[cui] (cui2) at (0, -1.10)
{\twolines{CUI: C0152031}{(Swollen joints)}};

\node[cui] (cui3) at (3.0, -1.10)
{\twolines{CUI: C0003873}{(RA)}};

\node[note] (note) at (0, -2.65)
{Clinical Note: \textit{``Patient reports \underline{morning stiffness}, \underline{swollen joints}, and was diagnosed with \underline{RA} $\ldots$''}};

\draw[arrow] (latent.north) -- (code1.south);
\draw[arrow] (latent.north) -- (code2.south);
\draw[arrow] (latent.north) -- (code3.south);

\draw[arrow] (latent.south) -- (cui1.north);
\draw[arrow] (latent.south) -- (cui2.north);
\draw[arrow] (latent.south) -- (cui3.north);

\draw[dashedarrow] (note.north) -- (cui1.south);
\draw[dashedarrow] (note.north) -- (cui2.south);
\draw[dashedarrow] (note.north) -- (cui3.south);

\node[
    draw,
    thick,
    shape=ellipse,
    fill=white,
    inner sep=2pt,
    font=\scriptsize\linespread{0.93}\selectfont
] at (4.65, -1.85) {Named Entity Recognition};

\node[align=left, font=\scriptsize\bfseries] at (-6.4, 1.10) {Structured Codes};
\node[align=left, font=\scriptsize\bfseries] at (-6.2, -1.10) {CUIs from NLP};

\end{tikzpicture}
\endgroup
\caption{ Illustration of multimodal EHR generation: A shared latent state $\bc_i$ generates both structured codes (top) and unstructured CUIs (bottom), motivating  \eqref{model}.}
\label{fig:multimodal-ehr}
\end{figure}

Our approach is motivated by real-world EHRs, where structured codes (e.g., diagnoses, medications) and unstructured concepts extracted from clinical notes (e.g., CUIs) offer complementary views of a patient’s health (Figure~\ref{fig:multimodal-ehr}). During a hospital visit, patient $i$ generates two types of data: structured codes and CUIs, which are typically collected by different subsystems and subject to distinct noise processes. For example, structured codes often reflect billing practices, while CUIs are derived from free-text notes via Named Entity Recognition (NER) tools. This data processing pipeline aligns with standard practice for mapping text to UMLS concepts \citep{choi2016learning, beam2019clinical}. We primarily focus on the two-modality case but also discuss extensions to more general multi-modality settings involving three or more modalities below.

Accordingly, for each patient $i$, we observe two sets of features: $\{ w^{(1)}_{i,t} \}_{t = 1}^{T_i^{(1)}}$ for structured codes and $\{ w^{(2)}_{i,t} \}_{t = 1}^{T_i^{(2)}}$ for CUIs, where $w^{(1)}_{i,t} \in \calW^{(1)}$ and $w^{(2)}_{i,t} \in \calW^{(2)}$. Let $\calW^{(1)} := [d_1]$ and $\calW^{(2)} := [d] \setminus [d_1]$, where $d = d_1 + d_2$ denotes the total number of distinct features across both modalities. 
To model the generative process of these features, we adopt the following latent variable model, which captures the shared structure between modalities while allowing for modality-specific noise:
\begin{equation}
\label{model}
\begin{aligned}
    \P(w^{(1)}_{i,t} = w \mid \bc_i, \beps_i^{(1)}) &= \frac{\exp(\langle \v_w^{\star}, \bc_i \rangle + \epsilon_{i,w}^{(1)} )}{\sum_{w' \in \calW^{(1)}} \exp(\langle \v_{w'}^{\star}, \bc_i \rangle + \epsilon_{i,w'}^{(1)} )}, \quad w \in \calW^{(1)},\ t \in [T_i^{(1)}], \\
    \P(w^{(2)}_{i,t} = w \mid \bc_i, \beps_i^{(2)}) &= \frac{\exp(\langle \v_w^{\star}, \bc_i \rangle + \epsilon_{i,w}^{(2)} )}{\sum_{w' \in \calW^{(2)}} \exp(\langle \v_{w'}^{\star}, \bc_i \rangle + \epsilon_{i,w'}^{(2)} )}, \quad w \in \calW^{(2)},\ t \in [T_i^{(2)}].
\end{aligned}
\end{equation}
Here, $\bc_i \sim \mathcal{N}(\mathbf{0}, \Ib_p)$ represents the latent clinical state of patient $i$, and $\v_w^\star \in \mathbb{R}^p$ is the ground-truth embedding of feature $w$. The modality-specific noise vectors $\beps^{(M)}_i = (\epsilon_{i,w}^{(M)})_{w \in \calW^{(M)}} \sim \mathcal{N}(\mathbf{0}, \bSigma_M)$, for $M \in \{1, 2\}$, capture variability not explained by the latent state, such as documentation inconsistencies or NLP extraction errors.

This formulation reflects our core assumption: while structured and unstructured features are conditionally generated from a shared latent representation $\bc_i$, they are distorted independently by modality-specific noise. The model thus supports:   (1) integration of heterogeneous EHR data sources,   (2) robustness to differing noise structures across modalities, and  (3) inference of informative embeddings from only summary-level statistics.

\begin{rem}
The assumption $\bc_i \sim N(\mathbf{0}, \Ib_p)$ is used to support theoretical analysis, not as a requirement for the CLAIME algorithm itself. Our method does not rely on estimating individual $\bc_i$  vectors. Instead, the Gaussian assumption enables analytical tractability in characterizing the expected co-occurrence patterns of EHR features, which in turn justify the use of SVD-based estimators in later sections. 
Similar analytical approximations have been derived under alternative latent structures, such as  unit sphere  random walks \citep{arora2016latent, arora2017simple} or scaled-Gaussian priors \citep{xu2023inference}. Finally, due to the rotational invariance of our model, the identity covariance assumption on $\bc_i$ can be relaxed without affecting the core methodology.
\end{rem}

\subsection{The CLAIME Algorithm}\label{sec: CLAIME}

We define code and CUI embedding matrices as $\Vb_1^{\star} = (\v_1^{\star},\ldots,\v_{d_1}^{\star} )^\top \in \R^{d_1 \times p}$ and $\Vb_2^{\star} = (\v_{d_1+1}^{\star},\ldots, \v_{d_1+d_2}^{\star} )^\top \in \R^{d_2 \times p}$, respectively, and aim to infer $\Vb^{\star} = \big( \Vb_1^{\star \top}, \Vb_2^{\star \top}\big)^\top \in \R^{d \times p}$. The embeddings are intended to reflect clinical semantics, meaning that highly similar (e.g., rheumatoid arthritis and juvenile rheumatoid arthritis) or related (e.g., fasting glucose and type II diabetes) EHR entities should be embedded close to each other.

{ To motivate our algorithm design, recall that model \eqref{model} assumes EHR features from both modalities are generated from a shared latent patient embedding $\bc_i$. This implies that features frequently co-occurring within the same patient are likely conditionally dependent through $\bc_i$, and should have similar embeddings, whereas features from different patients are marginally independent and their embeddings should remain unaligned.}

Before introducing our algorithm, we first define aggregate co-occurrence matrices $\CC^{(M, M)}$ and $\DD^{(M,M')}$ for $M, M' \in \{1, 2\}$ across different modalities as:
\begin{equation*}
 \CC^{(M, M)}(w, w') = \sum_{i=1}^n  \CC^{(M, M)}_{i,i}(w, w'),\ \  
 \DD^{(M, M')}(w, w') = \sum_{i=1}^n  \DD^{(M, M')}_{i,i}(w, w')
\end{equation*}
where $\CC^{(M, M)}_{i,j}(w, w') =  \big|\{(t,s) \in [T_i^{(M)}] \times [T_j^{(M)}]: t \neq s, w^{(M)}_{i,t} = w, w^{(M')}_{j,s} = w'\} \big|$ and $\DD^{(M, M')}_{i,j}(w, w') =  \big|\{(t,s) \in [T_i^{(M)}] \times [T_j^{(M')}]: w^{(M)}_{i,t} = w, w^{(M')}_{j,s} = w'\} \big|$ for $i, j \in [n]$. Further, we define the marginal occurrence count of $w \in \calW^{(M)}$ as:
\begin{equation}
\gamma_{w}^{(M)} = \CC^{(M,M)}(w, \cdot) = \sum_{w' \in \calW^{(M)}} \CC^{(M,M)}(w, w').
\label{gamma}
\end{equation}

{  To derive clinically meaningful embeddings that reflect modality-specific co-occurrence patterns, we propose a contrastive learning algorithm aligned with model~\eqref{model}. Contrastive learning encourages embeddings of positive pairs to be similar and those of negative pairs to be dissimilar. In our setting, code-CUI pairs from the same patient are treated as positive, while pairs from different patients as negative. This naturally aligns with our generative assumptions and enables learning from aggregated co-occurrence data. Specifically, we define the multimodal contrastive learning loss as} 
\begin{equation}
   \begin{aligned}
        &  \mL_{\CLAIME} (\Vb_1, \Vb_2) =   \frac{\sum_{i,j:i \neq j} s_{ij}}{ n ( nS_1^{(1)}  S_1^{(2)} - S_2^{(1,2) 2})   }   - \frac{\sum_{i=1}^n s_{ii}}{ n  S_2^{(1,2) 2}}   + \frac{\lambda}{2} \|\Vb_1 \Vb_2^\top\|_\F^2, 
     \end{aligned}  
\label{loss: MMCL linear}
\end{equation}
where $s_{ij} = \sum_{t \in [T_i^{(1)}]}  \sum_{s \in [T_j^{(2)}]}  \frac{\langle \v_{w^{(1)}_{i,t}}, \v_{w^{(2)}_{j,s}} \rangle}{\gamma_{w_{i,t}^{(1)}}^{(1)} \gamma_{w_{j,s}^{(2)}}^{(2)}}$. 
Here, $\Vb_1 = (\v_1, \ldots, \v_{d_1})^\top \in \R^{d_1 \times p}$ and $\Vb_2 = (\v_{d_1+1}, \ldots, \v_{d_1+d_2})^\top \in \R^{d_2 \times p}$, 
$\lambda > 0$ serves as a regularization coefficient, $\gamma_{w}^{(M)}, M =1,2$ are weights chosen based on the frequency of $w$, as defined in \eqref{gamma}. { The quantities  $S_q^{(M)} = ( n^{-1} \sum_{i=1}^n (T_i^{(M)})^q )^{1/q}$ and  $S_q^{(1, 2)} = ( n^{-1} \sum_{i \in [n]} (T_i^{(1)})^{q/2} (T_i^{(2)})^{q/2})^{1/q}$  for $q \geq 1$ are empirical moments and cross-moments of $T_i^{(M)}$, and correspond to the number of terms in the two summation components of the loss.} Our analysis in Section~\ref{sec: theory} motivates the choice of $\gamma_{w}^{(M)}$ to guide the minimizer towards $\Vb^{\star}$.

 This loss encourages code-CUI pairs that co-occur within the same patient (second term) to be closely embedded, while penalizing similarity between pairs from different patients (the first term). The regularization term $\frac{\lambda}{2} \|\Vb_1 \Vb_2^\top\|_\F^2$ ensures numerical stability and controls the embedding magnitude.  By optimizing this loss, the model learns to align embeddings based on co-occurrence patterns that signal shared clinical relevance.

To efficiently obtain $\Vbhat = (\Vbhat_1^\top, \Vbhat_2^\top)^\top = \arg \min \mL_{\CLAIME} (\Vb_1, \Vb_2)$, { we define normalized count vectors for each patient $i$ and modality $M \in \{1,2\}$: $\mathbf{w}_i^{(M)} \in \mathbb{R}^{d_M}$ with $\mathbf{w}_{i,w}^{(M)} = \frac{1}{\gamma_w^{(M)}} \sum_{t \in [T_i^{(M)}]} \mathbb{I}(w_{i,t}^{(M)} = w)$. The patient-level representation for each modality becomes $\Vb_1^\top \mathbf{w}_i^{(1)}$ and $\Vb_2^\top \mathbf{w}_i^{(2)} \in \mathbb{R}^p$. Using this, the loss \eqref{loss: MMCL linear} (excluding regularization) can be expressed by pair-wise co-occurrences of concepts as}: 
\begin{equation*}
   \begin{aligned}
        & \frac{1}{ n ( nS_1^{(1)}  S_1^{(2)} - S_2^{(1,2) 2})   } \sum_{w =  1}^{d_1} \sum_{w' = d_1+1}^{d_1+d_2}  \frac{\langle \v_{w}, \v_{w'} \rangle}{\gamma_{w}^{(1)} \gamma_{w'}^{(2)}  } \sum_{i=1}^n \sum_{j:j \neq i}^n \sum_{t \in [T_i^{(1)}]}  \sum_{s \in [T_j^{(2)}]}  \I( w^{(1)}_{i,t} = w) \I( w^{(2)}_{j,s} = w')  \\
        & \quad -  \frac{1}{ n  S_2^{(1,2) 2}} \sum_{w =  1}^{d_1} \sum_{w' = d_1+1}^{d_1+d_2}   \frac{\langle \v_{w}, \v_{w'} \rangle}{\gamma_{w} ^{(1)} \gamma_{w'}^{(2)}  } \sum_{i=1}^n \sum_{t \in [T_i^{(1)}]}  \sum_{s \in [T_i^{(2)}]} \I( w^{(1)}_{i,t} = w) \I( w^{(2)}_{i,s} = w') \\
        & = \frac{1}{ n ( nS_1^{(1)}  S_1^{(2)} - S_2^{(1,2) 2})}  \sum_{i=1}^n \sum_{j:j \neq i}^n (\Vb_1^\top \mathbf{w}_i^{(1)})^\top (\Vb_2^\top \mathbf{w}_j^{(2)}) -  \frac{1}{ n S_2^{(1,2) 2}} \sum_{i=1}^n (\Vb_1^\top \mathbf{w}_i^{(1)})^\top (\Vb_2^\top \mathbf{w}_i^{(2)}).
    \end{aligned}
\end{equation*}
{ The first term captures the average similarity of negative (cross-patient) pairs, while the second reflects positive (same-patient) pairs. Minimizing this loss encourages within-patient alignment while discouraging cross-patient alignment.} Subsequently, via arguments given in Supplementary~S2, we have the following proposition. 
\begin{prop}\label{prop: SVD}
    We have 
    \begin{align*}
        \mL_\CLAIME(\Vb_1, \Vb_2) &= \frac{\lambda}{2} \Big\| \Vb_1 \Vb_2^\top - \frac{1}{\lambda} \PMIbbhat_\CLAIME\Big\|_\F^2 + (\textnormal{constant}),
    \end{align*}
    where $\PMIbbhat_\CLAIME = \{ \PMIbbhat_\CLAIME(w,w') \}_{w \in \calW^{(1)},w' \in \calW^{(2)}}$ with 
    \begin{align*}
        \PMIbbhat_\CLAIME(w,w') := \frac{\CC^{(1,1)}(\cdot, \cdot) \CC^{(2,2)}(\cdot, \cdot)}{\CC^{(1,1)}(w, \cdot) \CC^{(2,2)}(w', \cdot)} \qty(\frac{\DD^{(1,2)}(w, w')}{\DD^{(1,2)}(\cdot, \cdot)} - \frac{ \CC^{(c)}(w, w')}{n ( nS_1^{(1)}  S_1^{(2)} - S_2^{(1,2) 2}) }),
    \end{align*}
    and $\CC^{(c)}(w, w') = \sum_{i=1}^n  \sum_{j:j \neq i}^n  \DD^{(1,2)}_{i,j}(w, w').$
\end{prop}
Proposition~\ref{prop: SVD} shows that $\mL_\CLAIME(\Vb_1, \Vb_2)$ reduces to a rank-constrained approximation of an empirical association matrix $\PMIbbhat_\CLAIME$, where each entry $ \PMIbbhat_\CLAIME(w,w')$ quantifies the association between features $w$ and $w'$. Section~\ref{sec: theory} establishes that this matrix converges to the population-level PMI matrix under mild assumptions.  

To estimate the embeddings, we compute the rank-$p$ SVD of $\PMIbbhat_\CLAIME$, denoted as $\widehat \Ub_1 \widehat \bLambda \widehat \Ub_2^{\top}$, where $\widehat \Ub_1 \in \R^{d_1 \times p}$ and $\widehat \Ub_2 \in \R^{d_2 \times p}$ are the matrices of left and right singular vectors, respectively, and $\widehat \bLambda \in \R^{p \times p}$ is a diagonal matrix with its diagonal elements being the top $p$ singular values. Then, we set $\Vbhat_1 = \widehat \Ub_1 \widehat \bLambda^{1/2}$ and  $\Vbhat_2 = \widehat \Ub_2 \widehat \bLambda^{1/2}$. { From Proposition~\ref{prop: SVD}, the scalar $\lambda$ in \eqref{loss: MMCL linear} controls the scale of the embeddings via the magnitude of singular values, but not their subspace. Since downstream similarity metrics like cosine similarity are scale-invariant, we may fix $\lambda = 1$ for simplicity.}

Our formulation is inspired by contrastive learning, which promotes high similarity among positive pairs and low similarity among negatives. Unlike vision-language models such as CLIP \citep{radford2021learning}, it operates on discrete, high-dimensional EHR features rather than continuous image or text embeddings, and relies only on aggregated co-occurrence statistics instead of raw patient-level data.

\begin{rem}[Nonlinear extension]
  The CLAIME framework can be naturally extended to a softmax-based contrastive loss similar to that used in CLIP:
    \begin{footnotesize}
    \begin{equation}
            \begin{aligned}
        \mL_{\CLAIME}'(\Vb_1, \Vb_2) &= - \frac{1}{ n (S_2^{(1,2)})^2} \sum_{i=1}^n T_i^{(1)} T_i^{(2)} \log \frac{\exp(s_{ii}/(T_i^{(1)} T_i^{(2)} \eta))}{\sum_{j:j\neq i} T_i^{(1)} T_j^{(2)} \exp(\alpha_i s_{ij}/(T_i^{(1)} T_j^{(2)} \eta))}  + R'(\Vb_1, \Vb_2),\label{loss: MMCL nonlinear}
    \end{aligned}
    \end{equation}
    \end{footnotesize}\noindent
    where $R'$ is a smooth regularizer, $\eta > 0$ is the temperature, and $\alpha_i$ depends on $(T_i^{(M)})_{i \in [n], M=1,2}$. In the high-temperature limit ($\eta \to \infty$), the loss reduces to the linear CLAIME loss \eqref{loss: MMCL linear}, up to a constant and regularization. Details are provided in Supplementary~S2.2. 
\label{rm2.1}
\end{rem}

\begin{rem}[Extending CLAIME to more than two modalities] EHR data often consist of multiple heterogeneous streams. For instance, codified data alone encompass diagnoses, procedures, medications, and laboratory results, each of which can be treated as a distinct modality. Similarly, NLP-derived CUIs can be partitioned by semantic types such as \emph{Disease or Syndrome} and \emph{Diagnostic Procedure}. To accommodate such settings, we generalize CLAIME beyond two modalities in Supplementary~S6 by estimating cross-modal PMI matrices for every pair of modalities and jointly learning embeddings over all sources. Further extensions to continuous-valued sources such as genomics, imaging, or clinical tests will require suitable generative models for continuous data.
\label{rm:extension} \end{rem}

\subsection{Comparison between CLAIME and Baseline Approaches}\label{sec: connection CL}

We next contrast CLAIME with the baseline approaches of ignoring between-modality differences. Dealing with multimodal data often presents difficulties, leading to conventional methods that overlook the differences between various modalities. A basic strategy commonly adopted is to simply merge the two modalities through direct concatenation, after which algorithms initially intended for unimodal data are applied. However, this rudimentary treatment of multimodal data may lead to substantial bias due to the inherent heterogeneity between different modalities.

To illustrate, consider the concatenated data for the $i$-th patient represented as $\{ w_{i,t} \}_{t \in [T_i]} = \Big( w_{i,1}^{(1)}, \dots, w_{i,T_i^{(1)}}^{(1)}, w_{i,1}^{(2)}, \dots, w_{i,T_i^{(2)}}^{(2)} \Big), \text{ where } T_i = T_i^{(1)} + T_i^{(2)} \,.$
A popular method to handle such data is the SVD-PMI algorithm, as referenced in \cite{levy2014neural,Gan2023}. In this context, we establish the co-occurrence matrices for the concatenated dataset as follows: 
$$
\CC =  \begin{bmatrix}
	\CC^{(1,1)} & \DD^{(1,2)} \\
	\DD^{(2,1)} & \CC^{(2,2)}
\end{bmatrix}.
$$
Subsequently, the empirical concatenated PMI matrix, denoted as $\PMIbbhat = \{  \PMIbbhat(w,w')\}_{w,w' \in [d]}$, is formulated as $\PMIbbhat(w,w') = \log \frac{\CC(w, w') \CC(\cdot, \cdot) }{\CC(w, \cdot) \CC(w', \cdot)}$, where $\CC(w, \cdot) = \sum_{w'=1}^d \CC(w, w')$ and $\CC(\cdot, \cdot) = \sum_{w=1}^d \CC(w, \cdot)$. Following this, we conduct a rank-$p$ eigen-decomposition of $\PMIbbhat$, represented as $\Ubhat_\PCA \bLambdahat_\PCA \Ubhat_\PCA^{\top}$. The estimator of $\Vb^{\star}$ is then achieved by setting $\Vbhat_\PCA = \Ubhat_\PCA \bLambdahat_\PCA^{1/2}$. We refer to this method as ``Concate''.

The second prevalent technique is contrastive learning (CL), applied directly to the concatenated dataset. Specifically, we define the contrastive loss for $\Vb \in \R^{d \times p}$ as follows:
\begin{equation}
   \begin{aligned}
   &  \mL_{\CL} (\Vb) =  - \frac{1}{\sum_{i=1}^n T_i (T_i-1) } \sum_{i=1}^n \sum_{t \in [T_i]}  \sum_{s \in [T_i] \setminus \{t\} }  \frac{\langle \v_{w_{i,t}}, \v_{w_{i,s}} \rangle}{\gamma_{w_{i,t}} \gamma_{w_{i,s}}}\\
    & \quad \quad + \frac{1}{\sum_{i=1}^n \sum_{j:j\neq i}^n T_i T_j } \sum_{i,j:i \neq j}  \sum_{t \in [T_i]}  \sum_{s \in [T_j]}  \frac{\langle \v_{w_{i,t}}, \v_{w_{j,s}} \rangle}{\gamma_{w_{i,t}} \gamma_{w_{j,s}}} + \frac{\lambda}{2} \|\Vb \Vb^\top\|_\F^2,\label{loss: CL linear}
    \end{aligned}         
\end{equation}
where $\gamma_w = \CC(w, \cdot)$. Let $\widehat \Vb_\CL = \arg \min \mL_{\CL} (\Vb)$. Similar to Proposition~\ref{prop: SVD}, we have the following proposition. Here,  we define
$$
\DD =  \begin{bmatrix}
	\CC^{(1)} & \CC^{(c)} \\
	\CC^{(c) \top} & \CC^{(2)}
\end{bmatrix} \text{ with } \CC^{(M)}(w, w') = \sum_{i=1}^n  \sum_{j:j \neq i}^n  \CC^{(M,M)}_{i,j}(w, w') \text{ for } w,w' \in \calW^{(M)}. 
$$

\begin{prop}\label{prop: PCA CL}
    We have $\mL_\CL(\Vb) = \frac{\lambda}{2} \Big\| \Vb \Vb^\top - \frac{1}{\lambda} \PMIbbhat_\CL\Big\|_\F^2 + (\textnormal{constant})$,
    where $\PMIbbhat_\CL = \{ \PMIbbhat_\CL(w,w') \}_{w ,w' \in [d]}$ with 
    \begin{align*}
                \PMIbbhat_\CL(w,w') := \frac{\CC(\cdot, \cdot) \CC(\cdot, \cdot)}{\CC(w, \cdot) \CC(w', \cdot)} \qty(\frac{\CC(w, w')}{\CC(\cdot, \cdot)} - \frac{ \DD(w, w')}{ n(n  S_1^{(1)}  S_1^{(2)} - S_2^{(1,2) 2} )}).
    \end{align*}
\end{prop}
Proposition~\ref{prop: PCA CL} reveals that obtaining $\Vbhat_\CL$ by minimizing the loss $\mL_\CL$ is essentially equivalent to applying PCA on the matrix $\PMIbbhat_\CL$. The proof for Proposition~\ref{prop: PCA CL} can be found in Supplementary~S4.4. This method is referred to as CL. Nevertheless, as we will demonstrate in Section~\ref{sec: theory}, both $\Vbhat_\PCA$ and $\Vbhat_\CL$ are not optimal solutions.

\section{Theoretical Analysis}\label{sec: theory}

In this section, we establish the theoretical properties of the proposed CLAIME algorithm and provide a comparative analysis with the two representative baselines discussed in Section~\ref{sec: connection CL}: the simple concatenation approach (Concate) and conventional contrastive learning (CL). Our goal is to articulate the statistical principles that underlie CLAIME, thereby clarifying both why and how it achieves performance gains relative to these alternatives.  

\subsection{Theoretical Properties of CLAIME}
First, we introduce some basic definitions. For $w \in \calW^{(M)}$ and $t \in [T_i^{(M)}]$, define $X_{i,w}^{(M)}(t) = \I\{w^{(M)}_{i,t} = w\}$. According to model~\eqref{model}, conditioned on $\bc_i$ and $\beps_{i}^{(M)} = (\epsilon_{i,w}^{(M)} )_{w \in \calW^{(M)}}$, the variable $X_{i,w}^{(M)}(t)$ follows a Bernoulli distribution with mean
\begin{align}
    p_{i,w}^{(M)} = \E[X_{i,w}^{(M)}(t) \mid \bc_i, \beps_{i}^{(M)}] = \frac{\exp(\langle \v_w^\star, \bc_i \rangle + \epsilon_{i,w}^{(M)})}{\sum_{w' \in \calW^{(M)}} \exp(\langle \v_{w'}^\star, \bc_i \rangle + \epsilon_{i,w'}^{(M)})}. \label{model: log linear}
\end{align}
The probability $p_{i,w}^{(M)}$ depends on the discourse vector $\bc_i$ and the noise vector $\beps_{i}^{(M)}$, and is independent of $t$. Let $p_w^{(M)} := \E[p_{i,w}^{(M)}]$.

Next, define the co-occurrence of $w \in \calW^{(M)}$ at time $t$ for patient $i$ and $w' \in \calW^{(M')}$ at time $s$ for patient $j$ as $X_{i,j,w,w'}^{(M,M')}(t,s) = \I\{w^{(M)}_{i,t} = w, w^{(M')}_{j,s} = w'\}$ for $t \ne s$, with mean $p_{i,j,w,w'}^{(M,M')}$. When $i = j$, define $p_{w,w'}^{(M,M')} := \E[p_{i,i,w,w'}^{(M,M')}]$. For $i \ne j$, we have $\E[p_{i,j,w,w'}^{(M,M')}] = p_w^{(M)} p_{w'}^{(M')}$. We then define the population PMI matrix $\PMIbb$ as
\begin{align}
    \PMIbb := \begin{pmatrix}
        \PMIbb^{(1,1)} & \PMIbb^{(1,2)}\\
        \PMIbb^{(2,1)} & \PMIbb^{(2,2)}
    \end{pmatrix}, \quad \PMIbb^{(M,M')}(w,w') = \log \frac{p_{w,w'}^{(M,M')}}{p_w^{(M)} p_{w'}^{(M')}} \label{q:PMI}
\end{align}
for $w \in \calW^{(M)}$, $w' \in \calW^{(M')}$, and $M, M' \in \{1,2\}$.

Before proceeding, we clarify our identifiability assumptions. From \eqref{model}, only the left singular space and singular values of $\Vb_1^{\star}$ and $\Vb_2^{\star}$ are identifiable. Hence, without loss of generality, we assume $\Vb_M^\star = \Ub_M^\star \bLambda_M^\star$ for $M \in \{1,2\}$, where $\Ub_M^\star \in \mathcal{O}_{d_M, p}$ contains the left singular vectors and $\bLambda_M^\star \in \R^{p \times p}$ is diagonal. Model~\eqref{model} is also invariant under the shift $\v_w^\star \gets \v_w^\star - (1/d_M) \sum_{w' \in \calW^{(M)}} \v_{w'}^\star$ for all $w \in \calW^{(M)}$, and $\bSigma_M \gets (\Ib - (1/d_M) \1_{d_M} \1_{d_M}^\top)\bSigma_M$. Hence, we may also assume $\Vb_M^{\star \top} \1_{d_M} = \zero$ and $\bSigma_M^\top \1_{d_M} = \zero$.

\begin{asm}\label{asm: signal covariance}
    For $M \in \{1, 2\}$, assume that $ \|\Ub_M^{\star} \|_{2,\infty} \lesssim \sqrt{p/d_M}$,  $\|\bLambda_M^{\star}\|/s_p(\bLambda_M^{\star}) \lesssim 1$,
    and $\|\bLambda_M^\star\| \ll 1$.
\end{asm}
Assumption~\ref{asm: signal covariance} is widely known as the incoherence constant condition in the literature appearing in matrix completion \citep{candes2009exact}, PCA \citep{zhang2022heteroskedastic}, and the analysis of representation learning algorithm \citep{ji2021power}.

\begin{asm}\label{asm: regime}
    Assume that $d_1 \wedge d_2 \gg p \log^2 (n + d)$ and $ (d_1 \wedge d_2)^2 \gg p^4 (d_1 \vee d_2) \log (n + d)$, $n \gg p^2 (d_1^2 \vee d_2^2) \log^2(n + d)$, and $n (S_1^{(1)} \wedge S_1^{(2)}) \gg p^2 (d_1^6 \vee d_2^6) \log^2 (n + d)$.
\end{asm}

Assumption~\ref{asm: regime} imposes a condition on the regime of unique features $d_1$ and $d_2$, and the number of patients $n$. We will discuss the sufficiency of the condition in Remark~\ref{rem: regime assumption sufficiency}.
This assumption is technically necessary in bounding the normalizing constant of the model (Lemma~S5.2). We introduce the notation $\bSigma = \diag( \bSigma_1, \bSigma_2)$.
\begin{asm}\label{asm: noise covariance}
    Assume that 
    $\|\diag(\bSigma_M)\|_{\max} \lesssim p/d_M$ for $M \in \{1, 2\}$ and $1/p \ll s_p(\bSigma) \leq \|\bSigma_1\| \vee \|\bSigma_2\| \lesssim 1$.
\end{asm}
Assumption~\ref{asm: noise covariance} implies that the signal-to-noise ratio is bounded above and below. 
Note that the signal, measured by the smallest positive singular value of $(1/p) \Vb_M^\star \Vb_{M'}^{\star \top}$ for $M, M' \in \{1, 2\}$ is of order $1/p$ from Assumption~\ref{asm: signal covariance}.
\begin{thm}\label{thm: PMI decomposition}
    Under Assumptions~\ref{asm: signal covariance}-\ref{asm: noise covariance}, we have     
    \begin{align*}
        \Big \| \PMIbb^{(M,M)} - \qty( \frac{1}{p} \Vb_M^{\star} \Vb_M^{\star\top} + \bSigma_M )  \Big\|_\F &\lesssim \frac{p^2}{d_M} \log(n+d), \ \ M \in \{1, 2\}\\
        \Big \|\PMIbb^{(1,2)} -  \frac{1}{p} \Vb_1^{\star} \Vb_2^{\star\top} \Big\|_\F &\lesssim \frac{p^2 (d_1 \vee d_2)}{(d_1 \wedge d_2)^2} \log(n+d),  \quad \text{     and hence } \\
        \Big \| \PMIbb - \qty(\frac{1}{p} \Vb^{\star} \Vb^{\star\top} + \bSigma )  \Big\|_\F     
        &\lesssim \frac{p^2 (d_1 \vee d_2)}{(d_1 \wedge d_2)^2} \log(n+d). 
    \end{align*}
\end{thm}

The proof of Theorem~\ref{thm: PMI decomposition} is presented in Supplementary~S4.1. It shows that the PMI matrix for each modality can be closely approximated by the product of embeddings plus the noise covariance matrix $\bSigma_M$. When the noise level $\bSigma_M$ is substantial, the PMI value $\PMIbb^{(M,M)}(w,w')$ for a single modality may significantly deviate from $\langle \v_w^{\star}, \v_{w'}^{\star}\rangle/p$. On the other hand, the cross-modal PMI matrix $\PMIbb^{(1,2)}$ can be directly approximated by $\Vb_1^{\star} \Vb_2^{\star\top}/p$. The next theorem affirms that $\PMIbbhat_\CLAIME$ is an effective estimator for $\PMIbb^{(1,2)}$.

\begin{asm}\label{asm: T moments}
    For $M = 1,2$, assume that $S_2^{(M) 2} \gg S_1^{(M)}$, 
    $S_1^{(M)} \gtrsim S_2^{(M)} \gtrsim S_4^{(M)}$ and $S_2^{(1,2)} \gtrsim S_4^{(1,2)}$.
    Also assume the non-negative correlation between $(T_i^{(1)})_{i \in [n]}$ and $(T_i^{(2)})_{i \in [n]}$, i.e., $S_2^{(1,2) 2} \geq S_1^{(1)} S_1^{(2)}$.
\end{asm}
Assumption~\ref{asm: T moments} requires that the empirical variance of $T_i^{(M)}$ dominates the mean of $T_i^{(M)}$, and their moments are comparable to each other. For the same length setting $T_i^{(M)} \equiv T^{(M)}$ for all $i$, the assumption is satisfied if $T^{(M)} \to \infty$. 
Assumption~\ref{asm: T moments} is satisfied for exponential distribution and gamma distribution with constant order shape parameter.

\begin{thm}\label{thm: PMI estimation mmcl} 
    Under Assumptions~\ref{asm: signal covariance}-\ref{asm: T moments}, with probability $1 - \exp(-\Omega(\log^2(n+d)))$, 
    \begin{align*}
         \norm{\PMIbbhat_\CLAIME - \PMIbb^{(1,2)}}_\F &\lesssim \left\{ \frac{d_1^{3/2} d_2^{3/2} }{\sqrt{n} (S_1^{(1) 1/2} \wedge S_1^{(2) 1/2})} +  \frac{d_1^{1/2} d_2^{1/2}}{\sqrt{n}}
         + \frac{d_1^{1/2} d_2^{1/2} p^2}{(d_1 \wedge d_2)^2} \right \} \log(n+d).
    \end{align*}
\end{thm}

The proof of Theorem~\ref{thm: PMI estimation mmcl} is given in Supplementary~S4.3. Under Assumption~\ref{asm: regime}, $\PMIbbhat_\CLAIME$ is an asymptotically unbiased estimate of $\PMIbb^{(1,2)}$. Combining Theorems~\ref{thm: PMI decomposition} and \ref{thm: PMI estimation mmcl} yields the convergence rate of CLAIME. We also analyze the rate of learning joint representations via CLAIME. Specifically, we  concatenate the left singular vectors of $\Vbhat_1$ and $\Vbhat_2$. Since $\Pb_p(\Vbhat_1^\top), \Pb_p(\Vbhat_2^\top)$ are only identifiable up to rotation, the subspace spanned by $\Pb_p([\Pb_p^\top(\Vbhat_1^\top), \Pb_p^\top(\Vbhat_2^\top)]$ depends on the choice of rotations. To address this, we introduce a matrix $\Hb$ and define the estimated joint representation by $\Ubhat_\Hb := \Pb_p([\Hb \Pb_p^\top(\Vbhat_1^\top), \Pb_p^\top(\Vbhat_2^\top)])$. 
We then evaluate performance via $\min_{\Hb \in \mathcal{O}_{p,p}}\|\sin\Theta(\Ubhat_\Hb, \Ub^\star)\|_\F$, where $\Ub^\star \in \mathcal{O}_{d,p}$ is the concatenated embedding vectors defined as $\Ub^\star := (1/\sqrt{2}) [\Ub_1^{\star \top}, \Ub_2^{\star \top}]$. The proof of Theorem~\ref{thm: MMCL convergence} is presented in Supplementary~S4.3.

\begin{thm}\label{thm: MMCL convergence}
    Under Assumptions~\ref{asm: signal covariance}-\ref{asm: T moments}, with probability $1 - \exp(-\Omega(\log^2(n+d)))$,  
    \begin{align}
        &\|\sin\Theta(\Pb_p( \Vbhat_1^\top ), \Ub_1^{\star} )\|_\F \vee \|\sin\Theta(\Pb_p(  \Vbhat_2^\top ), \Ub_2^{\star})\|_\F \lesssim \frac{p^3 (d_1 \vee d_2) \log(n+d)}{(d_1 \wedge d_2)^2}\nonumber\\
        &\quad\quad+ \frac{p d_1^{3/2} d_2^{3/2} \log(n+d)}{\sqrt{n} (S_1^{(1) 1/2} \wedge S_1^{(2) 1/2})} + \frac{p d_1^{1/2} d_2^{1/2}\log(n+d)} {\sqrt{n}}
        \quad \text{     and }
        \label{eq: bound MMCL convergence}
    \end{align}
    \begin{align*}
        \min_{\Hb \in \mathcal{O}_{p,p}} \|\sin\Theta(\Ubhat_\Hb, \Ub^\star)\|_\F &\lesssim \left \{ \frac{p^3 (d_1 \vee d_2)}{(d_1 \wedge d_2)^2} + \frac{p d_1^{3/2} d_2^{3/2} }{\sqrt{n} (S_1^{(1) 1/2} \wedge S_1^{(2) 1/2})}+ \frac{p d_1^{1/2} d_2^{1/2}}{\sqrt{n}} \right \} \log(n+d).
    \end{align*}
\end{thm}

\begin{rem}\label{rem: regime assumption sufficiency}
The bound in~\eqref{eq: bound MMCL convergence} consists of a bias term from the nonlinearity of the data-generating process (first term), and two variance terms (last two terms).
    The first variance term arises from estimating per-record co-occurrence frequencies. Each patient contributes $(T_i^{(1)}, T_i^{(2)})$-length multimodal data, so the variance of empirical co-occurrence frequencies is $O(1/(n(S_1^{(1)} \wedge S_1^{(2)})))$. Normalizing to obtain PMI entries scales this variance by $d_1 d_2$, and converting to Frobenius norm adds an extra factor of $d_1^{1/2} d_2^{1/2}$ via the inequality $\|A\|_\F \le d_1^{1/2} d_2^{1/2} \|A\|_{\max}$. Applying the Davis–Kahan theorem then gives a contribution of order $O(p d_1^{3/2} d_2^{3/2}/\sqrt{n(S_1^{(1)} \wedge S_1^{(2)}}))$. The second variance term reflects the averaging across $n$ patients, yielding a $1/\sqrt{n}$ rate, which again is scaled by $d_1^{1/2} d_2^{1/2}$ when converted to Frobenius norm. As a result, to make the overall bound vanish and ensure consistency, Assumption~\ref{asm: regime} is required.
\end{rem}

\subsection{Comparison with the Baseline Methods}\label{sec:theory:comparison}
We now compare CLAIME  with the two baseline approaches introduced in Section~\ref{sec: connection CL}. As discussed there, both baselines can be interpreted as applying SVD to different empirical PMI matrices: the first is SVD on the concatenated PMI matrix $\PMIbbhat$, while the second is based on the estimator $\PMIbbhat_\CL$ derived from the CL loss \eqref{loss: CL linear}. We show in Theorem~\ref{thm: PMI estimation cl} that both $\PMIbbhat$ and $\PMIbbhat_\CL$ are biased estimators of the population matrix $\PMIbb$. Building on this result, Theorem~\ref{thm: CL and PCA convergence} establishes the convergence rates of the corresponding estimators $\widehat \Vb_\CL$ and $\Vbhat_\PCA$. The proofs of both theorems are provided in Supplementary~S4.4.

\begin{thm}\label{thm: PMI estimation cl}
    Under Assumptions~\ref{asm: signal covariance}-\ref{asm: T moments}, with probability $1 - \exp(-\Omega(\log^2(n+d)))$,
    \begin{align*}
        \norm{\PMIbbhat - (\PMIbb + \Bb_\PCA)}_\F &\lesssim \frac{(d_1 \vee d_2) d_1^2 \log(n+d)}{\sqrt{n} S_1^{(1) 1/2}} + \frac{(d_1 \vee d_2) d_2^2 \log(n+d)}{\sqrt{n} S_1^{(2) 1/2}}\\
        &\quad+ \frac{(d_1 \vee d_2) d_1 d_2 \log(n+d)}{\sqrt{n} (S_1^{(1) 1/2} \wedge S_1^{(2) 1/2})} + \frac{(d_1 \vee d_2) \log(n+d)}{\sqrt{n}}, \quad \text{    and}
    \end{align*}
    \begin{align*}
         &\norm{\PMIbbhat_\CL - (\PMIbb + \Bb_\CL)}_\F \lesssim \frac{(d_1 \vee d_2) d_1^2 \log(n+d)}{\sqrt{n} S_1^{(1) 1/2}} + \frac{(d_1 \vee d_2) d_2^2 \log(n+d)}{\sqrt{n} S_1^{(2) 1/2}}\\
         &\quad\quad+ \frac{(d_1 \vee d_2) d_1 d_2 \log(n+d)}{\sqrt{n} (S_1^{(1) 1/2} \wedge S_1^{(2) 1/2})} + \frac{\log(n+d) (d_1 \vee d_2)}{\sqrt{n}} + \frac{p^2 (d_1 \vee d_2) \log(n+d)}{(d_1 \wedge d_2)^2}.
    \end{align*}
    Here, $\Bb_\PCA$ and $\Bb_\CL$ are bias matrices defined in Supplementary~S4.2 and S4.4.
\end{thm}

\begin{rem}
\label{rem:debias}
As shown in Theorems~S4.2 (Concate) and S4.5 (CL), the bias matrix $\mathbf{B}_{\CL}$ includes a correctable term based on observed $T_i^{(M)}$ and an uncorrectable term involving the population-level PMI. In contrast, $\mathbf{B}_{\PCA}$ depends only on $T_i^{(M)}$ and is fully correctable. To evaluate the effect of this correction, we conducted additional experiments by removing the bias term associated with $T_i^{(M)}$. The results, presented in Supplementary~S7.3.3, show that the debiased estimators offer little to no improvement over the original ones, suggesting that the impact of the removable bias component is limited in practice.
\end{rem}

\begin{thm}\label{thm: CL and PCA convergence} 
    Suppose that $\|\Bb_\PCA\| \vee \|\Bb_\CL\| \ll 1/p$. 
    Under Assumptions~\ref{asm: signal covariance}-\ref{asm: T moments}, with probability $1 - \exp(-\Omega(\log^2(n+d)))$,
    \begin{align*}
        &\norm{\sin\Theta\qty(\Pb_p(\Vbhat_\PCA^\top), \Pb_p\qty(\frac{1}{p} \Vb^\star \Vb^{\star \top} + \bSigma + \Bb_\PCA))}_\F\\
        &\quad\lesssim \frac{p (d_1 \vee d_2) d_1^2 \log(n+d)}{\sqrt{n} S_1^{(1) 1/2}} + \frac{p (d_1 \vee d_2) d_2^2 \log(n+d)}{\sqrt{n} S_1^{(2) 1/2}}\\
        &\quad\quad+ \frac{p (d_1 \vee d_2) d_1 d_2 \log(n+d)}{\sqrt{n} (S_1^{(1) 1/2} \wedge S_1^{(2) 1/2})} + \frac{p \log(n+d) (d_1 \vee d_2)}{\sqrt{n}} + \frac{p^3 (d_1 \vee d_2) \log(n+d)}{(d_1 \wedge d_2)^2},\\
        &\norm{\sin\Theta\qty(\Pb_p(\widehat \Vb_\CL^\top), \Pb_p\qty(\frac{1}{p} \Vb^\star \Vb^{\star \top} + \bSigma + \Bb_\CL ))}_\F\\
        &\quad\lesssim \frac{p (d_1 \vee d_2) d_1^2 \log(n+d)}{\sqrt{n} S_1^{(1) 1/2}} + \frac{p (d_1 \vee d_2) d_2^2 \log(n+d)}{\sqrt{n} S_1^{(2) 1/2}}\\
        &\quad\quad+ \frac{p (d_1 \vee d_2) d_1 d_2 \log(n+d)}{\sqrt{n} (S_1^{(1) 1/2} \wedge S_1^{(2) 1/2})} + \frac{p \log(n+d) (d_1 \vee d_2)}{\sqrt{n}} + \frac{p^3 (d_1 \vee d_2) \log(n+d)}{(d_1 \wedge d_2)^2}.
    \end{align*}
\end{thm}

Theorem~\ref{thm: CL and PCA convergence} shows that both Concate and CL will yield biased estimators for the embeddings $\Vb^{\star}$ due to the inherent bias terms and covariance matrix of the noise. While the bias terms $\Bb_\PCA$ and $\Bb_\CL$ can be  partially removed, the influence of $\bSigma$ remains irreducible due to its unobserved nature. Comparing Theorems~\ref{thm: MMCL convergence} and \ref{thm: CL and PCA convergence}, we find that CLAIME achieves a sharper rate than Concate and CL even when $\Bb_\CL$, $\Bb_\PCA$ and $\bSigma$ go to zero. As a result, CLAIME is a better estimator than  Concate and CL. For the bias terms $\Bb_\PCA$ and $\Bb_\CL$, we have the following lemma. The proof of Lemma~\ref{lem:bias} is given in Supplementary~S4.7. 

\begin{lem}\label{lem:bias}
Suppose Assumption~\ref{asm: T moments} holds and $T_i^{(1)} = T_i^{(2)}$ for $i \in [n]$. Then
$\|\Bb_\PCA\| \vee \|\Bb_\CL\| \lesssim (d_1 \vee d_2) S_1^{(1)}/S_2^{(1)2}$,
which reduces to $(d_1 \vee d_2)/T$ if $T_i^{(1)} = T_i^{(2)} \equiv T >0$.
\end{lem}

We then characterize the effect of $\bSigma$ and show that when $\bSigma$ satisfies suitable conditions, the two existing methods Concate and CL will not provide ideal estimators of $\Vb^{\star}$.

\begin{asm}\label{asm: noise covariance eigengap} 
     Assume that $(s_p(\bSigma) - s_{p+1}(\bSigma))/s_p(\bSigma) \gtrsim 1$.
\end{asm}

Assumption~\ref{asm: noise covariance eigengap} is known as the eigengap condition for the top-$p$ singular vectors of $\bSigma$. A similar assumption has been used in the analysis of contrastive learning \citep{ji2021power}.
This enables us to identify the top-$p$ singular vectors of $(1/p) \Vb^\star \Vb^{\star \top} + \bSigma$.

\begin{asm}\label{asm: noise covariance incoherence 1}
    Assume $\|\sin\Theta(\Ub_1^\star, \Pb_p(\bSigma_1))\|_\F \wedge \|\sin\Theta(\Ub_2^\star, \Pb_p(\bSigma_2))\|_\F \geq \sqrt{(2/3) p}$.
\end{asm}

Assumption~\ref{asm: noise covariance incoherence 1} holds if the column space of $\Pb_p(\bSigma_1)$ is nearly orthogonal to the column space of $\Ub^\star_1$
and that of $\Ub^\star_2$ nearly orthogonal to that of $\Pb_p(\bSigma_2)$. Roughly speaking, this assumption requires that we can well separate the features and noises based on their directions in each modality.

\begin{thm}\label{thm: PCA DDPCA MMCL}
    Suppose $\|\Bb_\PCA\| \vee \|\Bb_\CL\| \ll 1/p$.
    Under Assumptions~\ref{asm: signal covariance}-\ref{asm: noise covariance incoherence 1}, with probability $1 - \exp(-\Omega(\log^2(n+d)))$, $\|\sin\Theta(\Ubhat_\PCA, \Ub^\star)\|_\F \gtrsim \sqrt{p}$ and $\|\sin\Theta(\Ubhat_\CL, \Ub^\star)\|_\F \gtrsim \sqrt{p}$.  
\end{thm}
The formal statement is available in Theorem~S4.8. Theorem~\ref{thm: PCA DDPCA MMCL} shows that there is a huge deviation from the estimators of Concate and CL to the true parameters as $\|\sin\Theta(\Ubhat_\PCA, \Ub^\star)\|_\F$ and $\|\sin\Theta(\Ubhat_\CL, \Ub^\star)\|_\F$ are both lower bounded by the order of $\sqrt p$. Since for any $\Ubhat, \Ub^\star\in \mathcal{O}_{d,p}$, $\|\sin\Theta(\Ubhat, \Ub^\star)\|_\F$ can be trivially upper bounded by $\sqrt p$, this result shows that $\Ubhat_\PCA$ and $\Ubhat_\CL$ can hardly capture any signal in $\Ub^\star$.

We present an example construction of the error covariance matrix where we expect to see the performance difference among CLAIME, Concate, and CL.

\begin{cor}\label{cor: low-rankness in noise covariance matrix}
    Choose $\Pb^{(1)} \in \mathcal{O}_{d_1,p}$ satisfying $[\Ub_1^\star; \Pb^{(1)}] \in \mathcal{O}_{d_1,2p}$ and $\|[\Ub_1^\star; \Pb^{(1)}]\|_{2,\infty}^2 \lesssim p/d_1$.
    Similarly choose $\Pb^{(2)} \in \mathcal{O}_{d_2,p}$ such that $[\Ub_2^\star; \Pb^{(2)}] \in \mathcal{O}_{d_2,2p}$ and $\|[\Ub_2^\star; \Pb^{(2)}]\|_{2,\infty}^2 \lesssim p/d_2$.
    Let $\bSigma_1 := \Pb^{(1)} \Pb^{(1) \top}$, $\bSigma_2 := (1/2) \Pb^{(2)} \Pb^{(2) \top}$, $\Vb^\star_1 = \Ub^\star_1$ and $\Vb^\star_2 = \Ub^\star_2$. For these noise covariance matrices, Assumptions~\ref{asm: signal covariance}, \ref{asm: noise covariance} and \ref{asm: noise covariance eigengap} hold. In addition, Assumption~\ref{asm: noise covariance incoherence 1} holds since $\Pb^{(1) \top} \Vb_1^\star = \Pb^{(2) \top} \Vb_2^\star = \ZERO_{p \times p}$.
    Under additional assumptions on $(T_i^{(M)})_{i \in [n]}$, $d_1$ and $d_2$ (Assumptions~\ref{asm: regime} and \ref{asm: T moments}), with probability $1 - \exp(-\Omega(\log^2(n+d)))$,
    \begin{align*}
        \|\sin\Theta(\Ubhat_\PCA, \Ub^\star)\|_\F &\gtrsim \sqrt{p}, \ \  \|\sin\Theta(\Ubhat_\CL, \Ub^\star)\|_\F \gtrsim \sqrt{p} \ \ \text{ and }\\
        \min_{\Hb \in \mathcal{O}_{p,p}} \|\sin\Theta(\Ubhat_\Hb, \Ub^\star)\|_\F &\lesssim \left \{ \frac{p^3 (d_1 \vee d_2)}{(d_1 \wedge d_2)^2} + \frac{p d_1^{3/2} d_2^{3/2} }{\sqrt{n} (S_1^{(1) 1/2} \wedge S_1^{(2) 1/2})}+ \frac{p d_1^{1/2} d_2^{1/2}}{\sqrt{n}} \right \} \log(n+d).
    \end{align*}
\end{cor}
Corollary~\ref{cor: low-rankness in noise covariance matrix} directly follows from Theorems~\ref{thm: MMCL convergence} and \ref{thm: PCA DDPCA MMCL}. Note that the factor $1/2$ in the definition of $\bSigma_2$ guarantees the eigengap condition at the $p$-th largest singular value of the joint noise covariance matrix $\bSigma$.
The noise covariance in Corollary~\ref{cor: low-rankness in noise covariance matrix} implies that the subspace spanned by the noise is orthogonal to that of the signal, so that noisy fluctuations in the feature frequency distribution occur simultaneously across many features, in a way separable  from the signal. Under this condition, Concate and CL  tend to learn the eigenspace corresponding to noise covariance matrix, as they  are vulnerable to strong noise. In contrast, CLAIME effectively escapes from the noise and recovers the core representations when $n$ and $(d_1 \vee d_2) / (d_1 \wedge d_2)^2$ are sufficiently large.

\section{Simulation Studies}
\label{simulation}

In this section, we evaluate the performance of CLAIME and compare it to CL, Concate, and Kernel PCA (KPCA) through simulation studies.  KPCA is a nonlinear PCA method that operates in a high-dimensional feature space induced by a kernel function. Following  \citet{gupta2019improving}, we apply KPCA with Gaussian RBF kernel directly to the co-occurrence matrix $\DD$ given that the co-occurrence matrix can be viewed as a generalized $n$-gram similarity matrix.
When patient-level data is unavailable, the first three methods can still be applied by performing SVD on various aggregated PMI matrices, as described in Section~\ref{sec: method}. Since patient-level data is also available in the synthesized datasets, we can additionally apply the gradient descent-based variants: CLAIME-GD and CL-GD, which optimize the objective functions defined in \eqref{loss: MMCL linear} and \eqref{loss: CL linear}, respectively. The suffix ``-GD'' is added to distinguish them from their SVD-based counterparts. To enhance training efficiency, only $10$ negative samples per patient are stochastically sampled. The penalty parameter $\lambda$ is set to $1$, while the learning rate starts at $10^{-4}$ and decays by a factor of $10$ every $10$ epochs to ensure convergence. Both gradient-based algorithms converge when their loss functions change by less than $10^{-6}$.

We generate a patient embedding vector $\mathbf{c}_i \in \mathbb{R}^p$ for each patient $i$, independently drawn from a multivariate normal distribution $N(\mathbf{0}, \mathbf{I}_p)$. To construct the code and CUI embedding matrices, we first generate $\Vb^\star = \mathbf{R}\mathbf{M}$, where $\mathbf{M} \in \mathbb{R}^{d \times p}$ has i.i.d. standard normal entries, and $\mathbf{R}\in\mathcal{O}_{d,d}$ is an orthonormal matrix obtained via QR decomposition of a random matrix. The rows of $\Vb^\star$ are mean-centered (i.e., each row has zero mean) and rescaled so that its largest singular value is $1$. We then partition $\Vb^\star$ into $\Vb_1^\star$ (first $d_1$ rows) and $\Vb_2^\star$ (remaining $d_2$ rows). Performing SVD on $\Vb_1^\star$ and $\Vb_2^\star$ separately yields the column basis matrices $\Ub_1^\star$ and $\Ub_2^\star$. Next, we generate the data according to model~\eqref{model}, where the patient $i$ is assigned $T_i^{(1)}$ codes and $T_i^{(2)}$ CUIs, with $T_i^{(1)}, T_i^{(2)} \sim \text{Poisson}(50)$ truncated to exclude zeros. We consider two different cases for the error covariance structure.

For Case 1, we set $\mathbf{\Sigma}^{(M)} = \text{diag}(\mathbf{I}_{p/2},\mathbf{0})/c\in\mathbb{R}^{d_M\times d_M}$ for $M \in \{1,2\}$, where  $c>0$  controls the signal-to-noise ratio (SNR)  of the data generating process. For Case 2, we generated $\sigma_w \overset{i.i.d.}{\sim} \text{Unif}(0, 1)$ and set $\bSigma_{w,w'}^{(M)} = \rho^{|w-w'|} \sigma_w \sigma_{w'}/2$ for $w,w'\in [d_M]$ and $M \in \{1, 2\}$, where $\rho \in (0, 1)$ controls both the structure and the SNR. Specifically, larger $\rho$ values yield higher SNR in the synthesized data. 

For both cases, we set $d_1 = d_2 = d/2$ and conduct three experiments to compare the performance of CLAIME, CLAIME-GD, CL, CL-GD, Concate and KPCA under varying sample sizes $n$, feature sizes $d$, and SNR. Performance is evaluated using the metric $\text{Err}(\widehat{\mathbf{U}},\mathbf{U}^\star) = \max_{M=1,2}\|\widehat{\mathbf{U}}_M\widehat{\mathbf{U}}_M^{ \top} - \Ub_M^{\star} \Ub_M^{\star\top}\|_\F$, which measures the subspace distance between estimated and true embeddings. We first fix the SNR by setting $c = 0.2$ for Case 1 and $\rho=0.8$ for Case 2. In the first experiment, we vary the sample size $n$ in $\{2,4,6\} \times 10^4$, while fixing $d=200$ and $p=4$. In the second experiment, we vary $d \in \{100,200,300,400\}$, while fixing $n=2\times10^4$ and $p=4$. For the third experiment, we first fix $n=2\times10^4$, $d=200$ and $p=4$. Then, we vary $c$ equally spaced between $0.2$ and $0.8$ for Case 1, and we vary $\rho$ equally spaced between $0.2$ and $0.8$ for Case 2.

\begin{figure}[h!]
\includegraphics[width=0.75\linewidth]{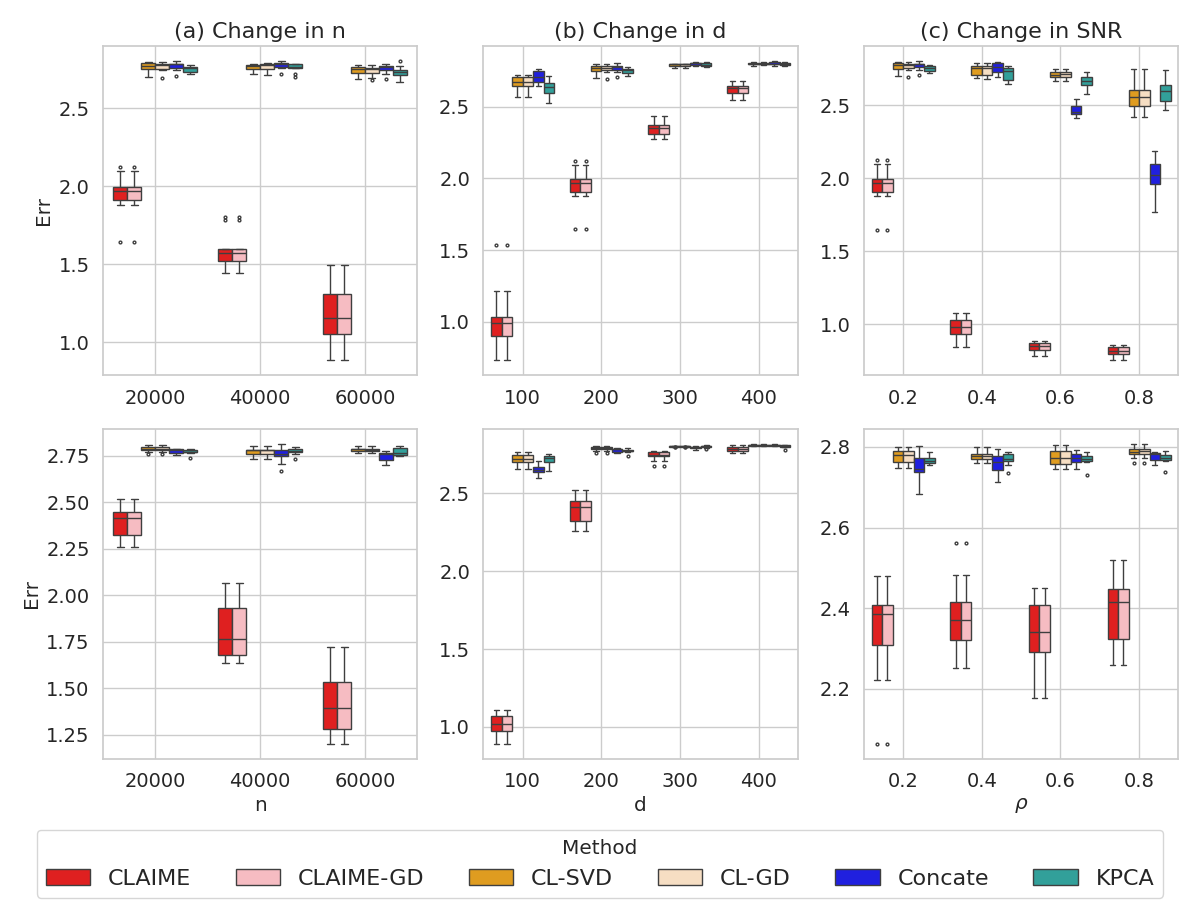}
\centering
\caption{In the top panel, the error metric is plotted against varying $n$, $d$ and $c$ for Case 1. In the bottom panel, the error metric is plotted against varying $n$, $d$ and $\rho$ for Case 2.}
\label{fig:simulation}
\end{figure}

Boxplots of the error metrics $\text{Err}(\widehat{\mathbf{U}},\mathbf{U}^\star)$, averaged over $100$ repetitions,  are presented in Figure~\ref{fig:simulation}. The results for CL and CL-GD, CLAIME and CLAIME-GD are very close to each other, as indicated by the proximity of their lines in Figures~\ref{fig:simulation} (a)--(c). This suggests that employing gradient-based methods on patient-level data (i.e., CL-GD and CLAIME-GD) is comparable to utilizing SVD on summary-level PMI matrices (i.e., CL and CLAIME). This observation aligns with the theoretical findings in Propositions~\ref{prop: SVD} and \ref{prop: PCA CL}. 
On one hand, in Figures~\ref{fig:simulation}(a) and (b), the errors for CLAIME and CLAIME-GD decrease with an increase in $n$ or a decrease in $d$. This behavior is anticipated and can be theoretically inferred from the upper bounds provided in Theorems~\ref{thm: PMI decomposition} -- \ref{thm: MMCL convergence}.
In contrast, the errors for Concate, CL, and CL-GD remain unchanged. This can be theoretically justified by their lower bounds being irrelevant to both $n$ and $d$, as shown in Theorem~\ref{thm: PCA DDPCA MMCL}. 
Moreover, it can be observed that KPCA behaves similarly to CL and Concate.
From the top panel of Figure~\ref{fig:simulation}(c), we observe that all methods incur larger errors as $c$ decreases—that is, as the SNR worsens. Compared to the other methods, CLAIME and CLAIME-GD consistently demonstrate superior performance. This finding supports our initial motivation: the proposed method is more robust in noisier data settings. Similarly, in the bottom panel of Figure~\ref{fig:simulation}(c), while other methods exhibit only a slight decline in performance as $\rho$ decreases, CLAIME and CLAIME-GD again significantly outperform the alternatives across all values of $\rho$. 

We conducted additional simulation experiments in Supplementary~S7.3.1 and~S7.3.2 to empirically examine larger $d$ and account for model misspecification. Similar patterns were observed, demonstrating the robustness of our methods across broader settings. Furthermore, in Supplementary~S7.3.3, we compared our approach against the Concate and CL methods with debiasing (Remark~\ref{rem:debias}), and our method continued to outperform them.

\section{Application to Electronic Health Records Studies}
\label{real}

We train a joint representation of codified and narrative CUI features using summary EHR data of patients with at least 1 diagnostic code of Rheumatoid Arthritis (RA) at Mass General Brigham (MGB). The MGB RA EHR cohort consists of 53,716 patients whose longitudinal EHR data have been summarized as co-occurrence counts of EHR concepts within a $30$-day window. We include all EHR diagnosis, medication, and procedure codes which have been rolled up to higher concept levels: diagnostic codes to PheCodes\footnote{https://phewascatalog.org/phecodes}, procedure codes to clinical classification system (CCS)\footnote{https://www.hcup-us.ahrq.gov/toolssoftware/ccs\_svcsproc/ccssvcproc.jsp}, medication codes to ingredient level RxNorm codes \citep{liu2005rxnorm}. We include all EHR codified features along with CUIs that have occurred more than $15$ times in the cohort. This results in $d=4,668$ features, including  $d_1 = 3,477$ codified features with $1,776$ PheCodes for diseases, $238$ CCS codes for procedures, and $1,463$ RxNorm codes for medications, and $d_2=1,048$ CUI features.

Based on the summary-level co-occurrence matrix, we can derive estimators for CLAIME, CL, Concate and KPCA, respectively. To assess the quality of the obtained embeddings, we utilize a benchmark previously introduced in \cite{Gan2023}. For multimodal learning, we are interested in our method's ability to capture cross-modality relationships. Specifically, we first assess the similarity between mapped code-CUI pairs, leveraging UMLS to map codified concepts to CUIs. For relatedness, we first curate UMLS relationships for CUI-CUI pairs, considering major classes like ``may treat or may prevent'', ``classifies'', ``differential diagnosis", ``method of'', and ``causative''. Then, we map disease CUIs to PheCodes, drugs to RxNorm, and procedures to CCS categories to obtain related code-CUI pairs. These mapped code pairs are then utilized to evaluate the system's ability to detect relatedness between codified and NLP features. 

To assess different relationship types, we compute cosine similarities between embedding vectors of known and randomly selected pairs. This enables calculation of the Area Under the Curve (AUC), quantifying how well the embeddings distinguish known pairs from random ones. Random pairs are matched on semantic type; for instance, when evaluating the ``may treat or may prevent'' relationship, we consider only disease–drug pairs. This strategy allows us to compare embedding performance across relationship types.

\begin{table}[h!]
    \centering
    \footnotesize
    \begin{tabular}{c|c|llll|c}
        \hline
        Type &  Group & CLAIME & CL & Concate & KPCA & Number of known pairs \\
        \hline
        \multirow{2}*{Similar} & CUI-PheCode & 0.910 & 0.911 & 0.911 & \textbf{0.916}  & 345\\
        & CUI-RxNorm & 0.944 & 0.898  & 0.900 & \textbf{0.986} & 49\\
        \cline{1-7}
        \multirow{4}*{Related} & May Treat (Prevent) & \textbf{0.775} & 0.753 & 0.753 & 0.711 &  1302\\
        & Classifies & \textbf{0.905} & 0.895 & 0.895 & 0.887 & 744\\
        & ddx & \textbf{0.773} & 0.766 &  0.765 & 0.670 & 1169\\
        & Causative & \textbf{0.775} & {0.742} & {0.741} & 0.695 & 524\\
          \hline
    \end{tabular}
\caption{AUCs of between-vector cosine similarity in detecting known similar or related pairs of codes vs CUIs for embeddings trained by different methods.}
\label{tab: union embed eval}
\end{table}

The results in Table~\ref{tab: union embed eval} show that all methods are more effective at identifying similar pairs than related ones, as indicated by consistently higher AUCs in the ``Similar'' category. Among the methods, KPCA performs best for detecting similarity, achieving the highest AUCs for both CUI-PheCode and CUI-RxNorm pairs. However, when it comes to the more challenging task of detecting relatedness, the proposed CLAIME method consistently outperforms the baselines, especially KPCA. Notably, CLAIME shows a substantial advantage in identifying causative relationships, with an approximate 3.3\% AUC improvement over CL and Concate, and an 8\% gain over KPCA. These results underscore CLAIME's robustness in capturing complex and nuanced associations between medical concepts.

Since the data come from the RA cohort, it is of interest for us to further perform case studies for RA-related codified and CUI concepts. We select $7$ key concepts for detailed analysis: rheumatoid arthritis (RA), C-reactive proteins (CP), stiffness of joint (SoJ), Reiter's disease (RD), swollen joint (SJ), pneumococcal vaccine (PV), and leflunomide (LF). For each of the $7$ concepts, we calculate its cosine similarity with all remaining concepts using the embeddings generated by CLAIME, Concate, and CL, respectively. Concepts are ranked by similarity, and a subset comprising the top-$100$ rankings from any one of the three methods is chosen as positive control. Then, we randomly select an equal number of other concepts as negative controls.  This process produces an evaluation set for each of the $7$ concepts. Concordance between similarity scores (from CLAIME, Concate, CL) and relevance scores (from GPT3.5 \citep{openai2023optimizing}, GPT4 \citep{achiam2023gpt} ) is assessed using Kendall's tau rank correlation. Relevance scores are obtained by prompting GPT3.5 and GPT4 to rate concept relevance on a scale of $0$ to $1$ between the given concept and its respective evaluation set. The detailed prompt is given in Supplementary~S7.1.

\begin{figure}[ht!]
\includegraphics[width=0.75\linewidth]{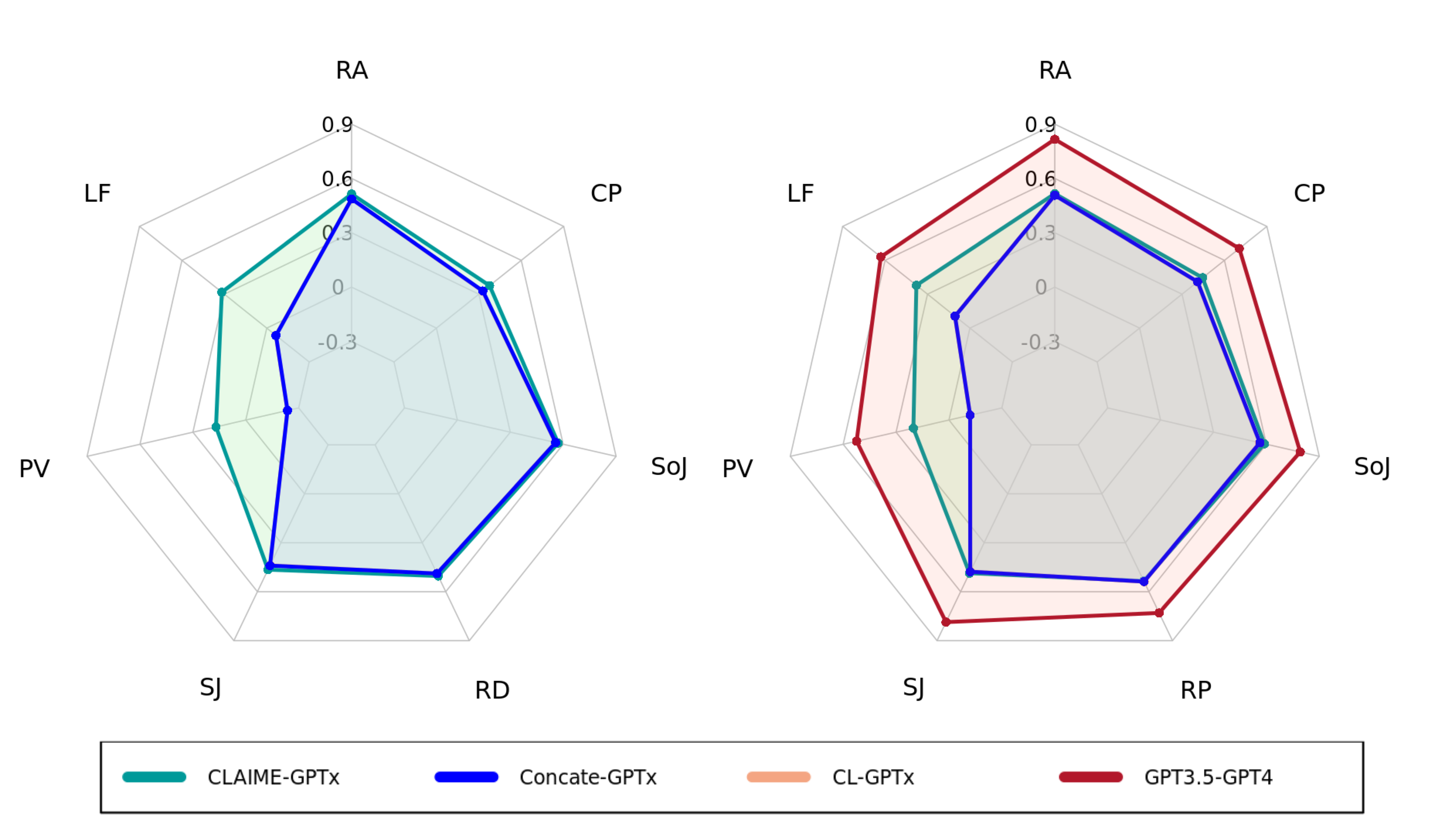}
\centering
\caption{Feature-specific Kendall’s rank correlations between CLAIME, Concate, and CL with GPT-3.5 and GPT-4 scores for 7 concepts. ‘GPTx’ denotes GPT-3.5 (left) and GPT-4 (right); note that CL-GPTx overlaps with Concate-GPTx and is not visible in the figure.}
\label{fig:real}
\end{figure}

The results, depicted in Figure~\ref{fig:real}, reveal that Concate and CL exhibit remarkably similar behavior, with the ``CL-GPTx'' lines obscured beneath ``Concate-GPTx''. CLAIME can effectively encode and preserve relations between these medical concepts and their associated features. Notably, negative rank correlations can be observed for CL-GPT3.5/4 and Concate-GPT3.5/4 regarding ``leflunomide'' and ``pneumococcal vaccine'' related codes and CUIs. To gain more insights, some detailed examples are provided in Table~S1 in Supplementary S7.1.

As shown in Supplementary~S6, CLAIME achieved strong performance when partitioning data elements into five modalities: diagnoses, procedures, medications, lab results, and CUIs. Its advantage is most evident in evaluating related relationships, with particularly pronounced improvements for ‘causative,’ ‘classifies,’ and ‘method of’ associations.

\section{Discussion and Conclusion}
\label{discussion}

In this paper, we propose a noisy multimodal log-linear generative model with an inner-product structure for analyzing multimodal EHR data and present the privacy-preserving algorithm CLAIME for estimating multimodal feature embeddings with theoretical justification.  Our algorithm is readily applicable to federated learning when multiple healthcare systems wish to co-train their models without sharing patient-level data. It is also of great interest to consider settings where different systems have overlapping but non-identical feature sets, in which case block-wise matrix completion techniques may be useful \citep{zhou2021multi}.

While our main development focuses on the two-modality setting, CLAIME is conceptually extensible to more than two modalities, as illustrated in Supplementary~S6. Theoretical justification, however, remains challenging. A central insight of our work is that, in the two-modality case, the loss function admits a closed-form global minimizer directly connected to the SVD of the PMI matrix. 
This reduction is crucial: it allows us to analyze the performance of the CLAIME representation through the top-$p$ singular space of $\PMIbbhat_\CLAIME$. By contrast, the straightforward multimodal extension lacks such an explicit form, making the global minimizer analytically intractable. 

This limitation reflects a broader challenge in the theory of contrastive learning, where rigorous analyses remain largely confined to the two-view case \citep{nakada2023understanding,dufumier2024align}. Moreover, our current extension assumes access only to pairwise co-occurrence matrices \citep{finlayson2014building,beam2020clinical}. If patient-level data were available, one could, in principle, construct higher-order co-occurrence tensors that capture joint dependencies across three or more modalities. Tensor SVD or related decompositions could then provide a natural way to generalize CLAIME embeddings with accompanying theoretical guarantees. Such a tensor-based framework could reveal interactions that are invisible to pairwise models, for example uncovering clinical pathways that involve coordinated use of lab tests, diagnoses, and treatments. Developing the statistical theory and scalable algorithms for such tensor-based generalizations remains an important direction for future work.

Beyond the linear loss considered here, exploring nonlinear losses such as those in Remark~\ref{rm2.1} is a natural next step. While a linear inner-product model captures only additive associations between modalities, nonlinear mappings (e.g., replacing the bilinear form $\Vb_1 \Vb_2^\top$ with a neural network) could model richer dependencies. However, ensuring identifiability and statistical guarantees would require new theory beyond classical low-rank approximation, making the balance between flexibility, interpretability, and tractability a key open challenge. Another promising direction is the estimation of patient embeddings $\bc_i$, which could enable clustering into clinically meaningful subgroups, modeling disease trajectories, and powering personalized applications such as “patients like me” recommendations. Such embeddings may be derived through joint factorization of patient–feature co-occurrence matrices or by incorporating temporal information from longitudinal data, with a theoretical framework for their statistical properties—consistency, stability, and generalization—being essential to ensure reliability in practice.

\paragraph{Data Availability.}

We validate the efficacy of our algorithm using EHR data from MGB. These data are protected under federal privacy regulations and cannot be shared publicly. The authors do not have authority to distribute the underlying MGB EHR data.

\paragraph{Acknowledgement.}

The authors thank the editor, the associate editors, and the referees for their constructive comments and suggestions. Tianxi Cai was supported by NIH grants R01 LM013614 and R01 NS098023.  Linjun Zhang was supported by NSF CAREER DMS-2340241 and Renaissance Philanthropy ``AI for Math'' Fund. Doudou Zhou was supported by the MOE AcRF Tier 1 Grant A-8003569-00-00 and the NUS Startup Grant A-0009985-00-00.

\paragraph{Conflict of Interest.} The authors report there are no competing interests to declare.

{\footnotesize 
\bibliographystyle{apalike}
\bibliography{ref}
}
\end{document}


\begin{center}
\textit{\large Supplementary Material to}
\end{center}

\begin{center}
{\LARGE Contrastive Learning on Multimodal Analysis of Electronic Health Records}
\vskip10pt
\end{center}

\setcounter{section}{0}
\renewcommand{\thesection}{S.\arabic{section}}
\setcounter{equation}{0}
\counterwithout{equation}{section}
\renewcommand{\theequation}{S.\arabic{equation}}
\setcounter{theorem}{0}
\counterwithout{theorem}{section}
\renewcommand{\thetheorem}{S.\arabic{theorem}}
\renewcommand{\thelemma}{S.\arabic{theorem}}

\renewcommand{\thesection}{S\arabic{section}}  
\renewcommand{\thetable}{S\arabic{table}}  
\renewcommand{\thefigure}{S\arabic{figure}}
\renewcommand{\theequation}{S\arabic{equation}}

\section{Notation}\label{subsec:notation}

For any square matrix $\bA \in \R^{d\times d}$, let $D(\bA) \triangleq \diag(A_{1,1}, \dots, A_{d,d})$ be the diagonal matrix consisting of diagonal entries of $\bA$. Also let $\Delta(\bA) \triangleq \bA - D(\bA)$ be the matrix with off-diagonal entries of $\bA$. 
Define $\tau \triangleq \log(n+d_1+d_2)$. We say an event $\cF$ occurs with high probability when $\P(\cF^c) = \exp(-\Omega(\tau^2))$.
We write $\exp(\bA)$ and $\log(\bA)$ for matrices whose entries are applied $x \mapsto \exp(x)$ and $x \mapsto \log x$ elementwise, respectively.
Let $\SVD_p(\bA)$ be the rank-$p$ approximation of $\bA$. When the rank-$p$ approximation of $\bA$ is not unique, we choose an arbitrary rank-$p$ approximation.
We write $\bSigma = \diag(\bSigma_1, \bSigma_2) = (\sigma_{w,w'})_{w,w'}$.

\paragraph{Concatenated Dataset.}

Recall that we concatenate two modalities into a concatenated dataset; for the $i$-th patient, we have
$$\{ w_{i,t} \}_{t \in [T_i]} = \Big( w_{i,1}^{(1)}, \dots, w_{i,T_i^{(1)}}^{(1)}, w_{i,1}^{(2)}, \dots, w_{i,T_i^{(2)}}^{(2)} \Big)\,.$$
For $w \in \calW$ and $t \in [T_i]$, let $X_{i,w}(t) = \I\{w_{i,t} = w\}$ be the occurrence of word $w \in \calW$ at time $t$. Also define the co-occurrence of words $w \in \calW$ at time $t$ and $w' \in \calW$ at time  $s$ as $X_{i,w,w'}(t, s) = \I\{w_{i,t} = w, w_{i,s} = w'\}$. For notational simplicity, we further define $\beps_i = (\beps_i^{(1)}, \beps_i^{(2)}) = (\epsilon_{i,w})_{w=1,\ldots,d}$.

Given a discourse vector $\bc_i$ and noise $\epsilon_{i,w}$, $X_{i,w}(t)$ follows a Bernoulli distribution with occurrence probability  
\begin{align*}
    p_{i,w}(t) &= \E[X_{i,w}(t)|\bc_i, \beps_i]\\
    &= \begin{cases}
        \frac{\exp(\langle \v_w^\star, \bc_i\rangle + \epsilon_{i,w})}{\sum_{w' \in \calW^{(M_w)}}\exp(\langle \v_{w'}^\star, \bc_i \rangle + \epsilon_{i,w'})} & \text{ if $M_w = 1$ and $t \leq T_i^{(1)}$, or $M_w = 2$ and $t > T_i^{(1)}$},\\
        0 & \text{otherwise},
    \end{cases}
\end{align*}
where $M_w \in \{1, 2\}$ is the modality of $w$.
Note that $p_{i,w}(t)$ is a function of discourse vector $\bc_i$ and noise $\epsilon_{i,w}$.

\paragraph{Co-occurrence Matrix.}

Note that we use the full window size for calculating the co-occurrence and cross-co-occurrence.
Let $M, M' \in \{1, 2\}$.
For $i, j \in[n]$, we define the co-occurrence of two words for concatenated data and for single source data, and define the cross-co-occurrence of two words between $i$-th and $j$-th patients as follows:
\begin{align}
    \CC_{i,j}(w, w') &:= |\{(t,s) \in [T_i] \times [T_j]: t \neq s, w_{i,t} = w, \ \ w_{j,s} = w'\}|, \ \ w, w' \in \calW,\nonumber\\
    \DD_{i,j}(w, w') &:= |\{(t,s) \in [T_i] \times [T_j]: w_{i,t} = w, \ \ w_{j,s} = w'\}|, \ \ w, w' \in \calW,\nonumber\\
    \CC^{(M, M)}_{i,j}(w, w') &:= |\{(t,s) \in [T_i^{(M)}] \times [T_j^{(M)}]: t \neq s, w^{(M)}_{i,t} = w, \ \ w^{(M)}_{j,s} = w'\}|, \ \ w \in \calW^{(M)}, w' \in \calW^{(M)}\nonumber\\
    \DD^{(M, M')}_{i,j}(w, w') &:= |\{(t,s) \in [T_i^{(M)}] \times [T_j^{(M')}]: w^{(M)}_{i,t} = w, \ \ w^{(M')}_{j,s} = w'\}|, \ \ w \in \calW^{(M)}, w' \in \calW^{(M')}.\label{eq: DD def}
\end{align} 
We notice that for $w, w' \in \calW^{(M)}$, $\CC_{i,i}(w, w') = \CC^{(M,M)}_{i,i}(w, w')$ holds and for $w \in \calW^{(1)}, w' \in \calW^{(2)}$, $\CC_{i,i}(w,w') = \DD_{i,i}^{(1,2)}(w,w')$ and $\DD_{i,j}(w, w') = \DD^{(1,2)}_{i,j}(w, w')$ holds for any $i, j \in [n]$.
Define $\CC(w, w') = \sum_{i=1}^n \CC_{i,i}(w, w')$, $\CC^{(M,M)}(w, w') = \sum_{i=1}^n \CC^{(M,M)}_{i,i}(w, w')$, and $\DD^{(1,2)}(w, w') = \sum_{i=1}^n \DD^{(1,2)}_{i,i}(w, w')$.
Let $\CC^{(M,M)}(w, \cdot) = \sum_{w' \in \calW^{(M)}} \CC^{(M,M)}(w, w')$ be the marginal co-occurrence of $w \in \calW^{(M)}$. Let $\CC^{(M,M)}(\cdot, \cdot) = \sum_{w \in \calW^{(M)}} \CC^{(M,M)}(w, \cdot)$ and\\
$\DD^{(M,M')}(\cdot, \cdot) = \sum_{w \in \calW^{(M)},w'\in\calW^{(M')}} \DD^{(M,M')}(w, w')$.
We define the marginal frequency $\Rb = \diag((\CC(w, \cdot)/\CC(\cdot, \cdot))_{w \in \calW})$ and $\Rb^{(M)} = \diag((\CC^{(M,M)}(w, \cdot)/\CC^{(M,M)}(\cdot, \cdot))_{w \in \calW^{(M)}})$.

\section{Proof of Section \ref{sec: CLAIME}}\label{sec: proof of sec: connection MMCL}

\subsection{Loss Reformulation}

\begin{prop}[Restatement of Proposition~\ref{prop: SVD}]\label{prop: SVD ap}
    We have 
    \begin{align*}
        \mL_\CLAIME(\Vb_1, \Vb_2) &= \frac{\lambda}{2} \Big\| \Vb_1 \Vb_2^\top - \frac{1}{\lambda} \PMIbbhat_\CLAIME\Big\|_\F^2 + (\textnormal{constant}),
    \end{align*}
    where $\PMIbbhat_\CLAIME = \{ \PMIbbhat_\CLAIME(w,w') \}_{w \in \calW^{(1)},w' \in \calW^{(2)}}$ with 
    \begin{align*}
        \PMIbbhat_\CLAIME(w,w') &\triangleq \frac{\CC^{(1,1)}(\cdot, \cdot) \CC^{(2,2)}(\cdot, \cdot)}{\CC^{(1,1)}(w, \cdot) \CC^{(2,2)}(w', \cdot)} \qty(\frac{\DD^{(1,2)}(w, w')}{\DD^{(1,2)}(\cdot, \cdot)} - \frac{\sum_{i,j: i \neq j} \DD_{i,j}^{(1,2)}(w, w')}{ \sum_{i,j: i \neq j} T_i^{(1)} T_j^{(2)} }).
    \end{align*}
\end{prop}

\begin{proof}[Proof of Proposition~\ref{prop: SVD ap}]
    Note that the loss \eqref{loss: MMCL linear} is defined as
    \begin{equation}
        \begin{aligned}
            &  \mL_{\CLAIME} (\Vb_1, \Vb_2) =  - \frac{1}{\sum_{i=1}^n  T_i^{(1)} T_i^{(2)} } \sum_{i=1}^n \sum_{t \in [T_i^{(1)}]}  \sum_{s \in [T_i^{(2)}]}  \frac{\langle \v_{w^{(1)}_{i,t}}, \v_{w^{(2)}_{i,s}} \rangle}{\gamma_{w_{i,t}}^{(1)} \gamma_{w_{i,s}}^{(2)}}\\
            & \quad \quad + \frac{1}{\sum_{i=1}^n \sum_{j:j\neq i}^n T_i^{(1)} T_j^{(2)} } \sum_{i=1}^n \sum_{j:j \neq i}^n \sum_{t \in [T_i^{(1)}]}  \sum_{s \in [T_j^{(2)}]}  \frac{\langle \v_{w^{(1)}_{i,t}}, \v_{w^{(2)}_{j,s}} \rangle}{\gamma_{w_{i,t}}^{(1)} \gamma_{w_{j,s}}^{(2)}} + \frac{\lambda}{2} \|\Vb_1 \Vb_2^\top\|_\F^2.\label{eq: claime loss}
        \end{aligned}         
    \end{equation}
    Note that
    \begin{align*}
        &\frac{1}{\sum_{i=1}^n  T_i^{(1)} T_i^{(2)} } \sum_{i=1}^n \sum_{t \in [T_i^{(1)}]} \sum_{s \in [T_i^{(2)}]} \frac{1}{\gamma_{w_{i,t}^{(1)}}^{(1)} \gamma_{w_{i,s}^{(2)}}^{(2)}} \langle \v_{w_{i,t}^{(1)}}, \v_{w_{i,s}^{(2)}} \rangle\\
        &\quad= \frac{1}{\sum_{i=1}^n T_i^{(1)} T_i^{(2)} } \sum_{i=1}^n \sum_{w \in \calW^{(1)}, w' \in \calW^{(2)}} \sum_{t \in [T_i^{(1)}]} \sum_{s \in [T_i^{(2)}]} \frac{1}{\gamma_w \gamma_{w'}} \v_{w'}^\top \v_w \I\{w_{i,t}^{(1)} = w, w_{i,s}^{(2)} = w'\}\\
        &\quad= \frac{1}{\sum_{i=1}^n  T_i^{(1)} T_i^{(2)} } \sum_{i=1}^n \sum_{w \in \calW^{(1)}, w' \in \calW^{(2)}} \frac{1}{\gamma_w \gamma_{w'}} \DD^{(1,2)}_{i,i}(w, w') \v_{w'}^\top \v_w\\
        &\quad= \tr\qty(\frac{1}{\sum_{i=1}^n  T_i^{(1)} T_i^{(2)} } \sum_{i=1}^n \Rb^{(1) -1} \DD^{(1,2)}_{i,i} \Rb^{(2) -1} \Vb_2 \Vb_1^\top).
    \end{align*}
    By a similar argument, we can rewrite the second term in \eqref {eq: claime loss} to obtain
    \begin{align*}
        \mL_\CLAIME(\Vb_1, \Vb_2) &= - \tr\qty(\frac{1}{\sum_{i=1}^n  T_i^{(1)} T_i^{(2)} } \sum_{i=1}^n \Rb^{(1) -1} \DD^{(1,2)}_{i,i} \Rb^{(2) -1} \Vb_2 \Vb_1^\top)\\
        &\quad+ \tr\qty(\frac{1}{\sum_{i,j:i \neq j}  T_i^{(1)} T_j^{(2)} } \sum_{i,j:i\neq j} \Rb^{(1) -1} \DD^{(1,2)}_{i,j} \Rb^{(2) -1} \Vb_2 \Vb_1^\top) + \Pi(\Vb_1, \Vb_2)\\
        &= - \tr(\PMIbbhat_\CLAIME \Vb_2 \Vb_1^\top) + \Pi(\Vb_1, \Vb_2),
    \end{align*}
    where we used $\CC^{(M,M)}(\cdot, \cdot) = \sum_i T_i^{(M)} (T_i^{(M)} - 1)$ for $M \in \{1, 2\}$ and $\DD^{(1,2)}(\cdot, \cdot) = \sum_i T_i^{(1)} T_i^{(2)}$.
    Since $\Pi(\Vb_1, \Vb_2) = (\lambda/2)\|\Vb_1 \Vb_2^\top\|_\F^2$, we can further rewrite $\mL_\CLAIME(\Vb_1, \Vb_2)$ as
    \begin{align*}
        \mL_\CLAIME(\Vb_1, \Vb_2) &= - \tr(\PMIbbhat_\CLAIME \Vb_2 \Vb_1^\top) + \Pi(\Vb_1, \Vb_2)\\
        &= \frac{\lambda}{2} \norm{\Vb_1 \Vb_2^\top - \frac{1}{\lambda} \PMIbbhat_\CLAIME}_\F^2 + (\textnormal{constant}).
    \end{align*}
\end{proof}

\subsection{Extension to Nonlinear Loss}\label{sec: nonlinear app}

\begin{prop}\label{prop: nonlinear}
    Let $\mL_{\CLAIME}$ and $\mL_{\CLAIME}'$ be the linear and nonlinear loss functions defined in \eqref{loss: MMCL linear} and \eqref{loss: MMCL nonlinear}, respectively. Then,
    \begin{align*}
        &\lim_{\eta\to\infty}\eta \qty[\mL_{\CLAIME}'(\Vb_1,\Vb_2) - \frac{1}{\sum_{i=1}^n T_i^{(1)} T_i^{(2)}} \sum_{i=1}^n T_i^{(1)} T_i^{(2)} \log\sum_{j: j \neq i} T_i^{(1)} T_j^{(2)} -R'(\Vb_1,\Vb_2)]\\
        &\quad= \mL_{\CLAIME}(\Vb_1,\Vb_2)-\frac{\lambda}{2}\|\Vb_1 \Vb_2^\top\|_\F^2.
    \end{align*}
\end{prop}

\begin{proof}
    Recall that
    \begin{align}
        \mL_{\CLAIME}'(\Vb_1,\Vb_2)
        &=-
        \frac{1}{\sum_{i=1}^{n}T_i^{(1)}T_i^{(2)}}
          \sum_{i=1}^{n} T_i^{(1)} T_i^{(2)}
              \log\frac{\exp\bigl(m_{ii}/\eta\bigr)}
                       {\sum_{j:j\neq i}^{n}T_i^{(1)}T_j^{(2)}
                             \exp\bigl(\alpha_i m_{ij}/\eta\bigr)}
          +R'(\Vb_1,\Vb_2),
          \label{eq: claime nonlinear}
    \end{align}
    where $R'$ is any regularizer, $\eta > 0$ is the temperature parameter, and 
    \begin{align}
        m_{ij} := \frac{1}{T_i^{(1)} T_j^{(2)}} \sum_{t \in [T_i^{(1)}]} \sum_{s \in [T_j^{(2)}]} \frac{\langle \v_{w_{i,t}^{(1)}}, \v_{w_{j,s}^{(2)}} \rangle}{\gamma_{w_{i,t}^{(1)}}^{(1)} \gamma_{w_{j,s}^{(2)}}^{(2)}},
        \ \ \alpha_i := \frac{\sum_{j: j\neq i} T_i^{(1)} T_j^{(2)}}{T_i^{(1)} T_i^{(2)}} \cdot \frac{\sum_{i=1}^n T_i^{(1)} T_i^{(2)}}{\sum_{i \neq j} T_i^{(1)} T_j^{(2)}}.\label{eq: alpha i}
    \end{align}
    To compare objectives, first subtract a constant term and drop the additive regularizer:
    \begin{align}
        &\mL_{\CLAIME}'(\Vb_1,\Vb_2) - \frac{1}{\sum_{i=1}^n T_i^{(1)} T_i^{(2)}} \sum_{i=1}^n T_i^{(1)} T_i^{(2)} \log\sum_{j: j \neq i} T_i^{(1)} T_j^{(2)} - R'(\Vb_1,\Vb_2)\nonumber\\
      &\quad=
      -
      \frac{1}{\sum_{i=1}^{n}T_i^{(1)}T_i^{(2)}}
      \sum_{i=1}^{n} T_i^{(1)}T_i^{(2)} \qty{
          \frac{m_{ii}}{\eta}
          -\log
           \frac{
                 \sum_{j:j\neq i} T_i^{(1)}T_j^{(2)}
                 \exp\bigl(\alpha_i m_{ij}/\eta\bigr)
               }{
                 \sum_{j:j\neq i} T_i^{(1)} T_j^{(2)}
               }
         }.\label{eq: nonlinear loss reformulation}
    \end{align}
    By Taylor approximation we have that for any fixed $x$,
    \[
      \exp\bigl(x/\eta\bigr)
      =1+\frac{x}{\eta}+O\bigl(x^{2}/\eta^{2}\bigr),
      \qquad
      \text{and}\qquad
      \lim_{\eta\to\infty}\eta\log\bigl(1+x/\eta\bigr)=x.
    \]
    This gives
    \begin{align*}
        &\eta \qty[\mL_{\CLAIME}'(\Vb_1,\Vb_2) - \frac{1}{\sum_{i=1}^n T_i^{(1)} T_i^{(2)}} \sum_{i=1}^n T_i^{(1)} T_i^{(2)} \log\sum_{j: j \neq i} T_i^{(1)} T_j^{(2)} - R'(\Vb_1,\Vb_2)]\\
      &\quad=
      -
      \frac{1}{\sum_{i=1}^{n}T_i^{(1)}T_i^{(2)}}
      \sum_{i=1}^{n} T_i^{(1)}T_i^{(2)} \qty{
          m_{ii}
          -\eta \log
           \qty(1 + \frac{
                 \sum_{j:j\neq i} T_i^{(1)}T_j^{(2)}
                 \bigl(\alpha_i m_{ij} + O(\alpha_i^2 m_{ij}^2/\eta)\bigr)
               }{
                 \eta \sum_{j:j\neq i} T_i^{(1)} T_j^{(2)}
               })
         }.
    \end{align*}
    Sending \(\eta\to\infty\) yields
    \begin{align*}
        &\lim_{\eta\to\infty}\eta \qty[\mL_{\CLAIME}'(\Vb_1,\Vb_2) - \frac{1}{\sum_{i=1}^n T_i^{(1)} T_i^{(2)}} \sum_{i=1}^n T_i^{(1)} T_i^{(2)} \log\sum_{j: j \neq i} T_i^{(1)} T_j^{(2)} -R'(\Vb_1,\Vb_2)]\\
        &\quad= -\frac{1}{\sum_{i=1}^{n} T_i^{(1)}T_i^{(2)}} \sum_{i=1}^{n} T_i^{(1)}T_i^{(2)} \qty{
            m_{ii}
              - \frac{
                     \sum_{j:j\neq i} T_i^{(1)}T_j^{(2)}
                     \alpha_i m_{ij}
                   }{
                     \sum_{j:j\neq i} T_i^{(1)} T_j^{(2)}
                   }
             }\\
        &\quad= -\frac{1}{\sum_{i=1}^{n}T_i^{(1)}T_i^{(2)}} \sum_{i=1}^{n} T_i^{(1)}T_i^{(2)} m_{ii} + \frac{1}{\sum_{i \neq j} T_i^{(1)} T_j^{(2)}} \sum_{i,j:i \neq j} T_i^{(1)} T_j^{(2)} m_{ij},
    \end{align*}
    where we used the definition of $\alpha_i$ in \eqref{eq: alpha i}.
    By definition of the linear loss $\mL_{\CLAIME}$, and using $s_{ij} = T_i^{(1)} T_j^{(2)} m_{ij}$, this concludes the proof.
\end{proof}

\section{Proof of Section \ref{sec: connection CL}}\label{sec: proof of sec: connection CL}

\subsection{Loss Reformulation}

\begin{prop}[Restatement of Proposition~\ref{prop: PCA CL}]\label{prop: PCA CL ap}
    We have 
    \begin{align*}
        \mL_\CL(\Vb) &= \frac{\lambda}{2} \Big\| \Vb \Vb^\top - \frac{1}{\lambda} \PMIbbhat_\CL\Big\|_\F^2 + (\textnormal{constant}),
    \end{align*}
    where $\PMIbbhat_\CL = \{ \PMIbbhat_\CL(w,w') \}_{w ,w' \in [d]}$ with 
    \begin{align*}
        \PMIbbhat_\CL(w,w') \triangleq \frac{\CC(\cdot, \cdot) \CC(\cdot, \cdot)}{\CC(w, \cdot) \CC(w', \cdot)} \qty(\frac{\CC(w, w')}{\CC(\cdot, \cdot)} - \frac{\sum_{i,j: i \neq j} \DD_{i,j}(w, w')}{ \sum_{i=1}^n \sum_{j:j \neq i}^n T_i T_j }).
    \end{align*}
\end{prop}

\begin{proof}[Proof of Proposition~\ref{prop: PCA CL ap}]
    The proof is similar to the proof of Proposition~\ref{prop: SVD ap}.
    Recall that the loss \eqref{loss: CL linear} is defined as
    \begin{align*}
        &\mL_{\CL} (\Vb) = -\frac{1}{\sum_{i=1}^n T_i (T_i-1) } \sum_{i=1}^n \sum_{t \in [T_i]}  \sum_{s \in [T_i] \setminus \{t\} }  \frac{\langle \v_{w_{i,t}}, \v_{w_{i,s}} \rangle}{\gamma_{w_{i,t}} \gamma_{w_{i,s}}}\\
        &\quad \quad + \frac{1}{\sum_{i=1}^n \sum_{j:j\neq i}^n T_i T_j } \sum_{i,j: i \neq j}  \sum_{t \in [T_i]}  \sum_{s \in [T_j]}  \frac{\langle \v_{w_{i,t}}, \v_{w_{j,s}} \rangle}{\gamma_{w_{i,t}} \gamma_{w_{j,s}}} + \frac{\lambda}{2} \|\Vb \Vb^\top\|_\F^2.
    \end{align*}
    Note that
    \begin{align*}
        &\frac{1}{\sum_{i=1}^n T_i (T_i-1)} \sum_{i=1}^n \sum_{t \in [T_i]} \sum_{s \in [T_i] \setminus \{t\}} \frac{1}{\gamma_{w_{i,t}} \gamma_{w_{i,s}}} \langle \v_{w_{i,t}}, \v_{w_{i,s}} \rangle\\
        &\quad= \frac{1}{\sum_{i=1}^n T_i (T_i-1)} \sum_{i=1}^n \sum_{w, w' \in \calW} \sum_{t \in [T_i]} \sum_{s \in [T_i] \setminus \{t\}} \frac{1}{\gamma_w \gamma_{w'}} \v_{w'}^{\star \top} \v_w^\star \I\{w_{i,t} = w, w_{i,s} = w'\}\\
        &\quad= \frac{1}{\sum_{i=1}^n T_i (T_i-1)} \sum_{i=1}^n \sum_{w, w' \in \calW} \frac{1}{\gamma_w \gamma_{w'}} \CC_{i,i}(w, w') \v_{w'}^{\star \top} \v_w^\star\\
        &\quad= \tr\qty(\frac{1}{\sum_{i=1}^n T_i (T_i-1)} \sum_{i=1}^n \Rb^{-1} \CC_{i,i} \Rb^{-1} \Vb \Vb^\top).
    \end{align*}
    By a similar argument as above, we can show that
    \begin{align*}
        \mL_\CL(\Vb) &= - \tr\qty(\Rb^{-1} \qty(\frac{1}{\sum_{i=1}^n T_i (T_i-1)} \sum_{i=1}^n \CC_{i,i}) \Rb^{-1} \Vb \Vb^\top)\\
        &\quad+ \tr\qty(\Rb^{-1} \qty(\frac{1}{\sum_{i,j:i\neq j}^n T_i T_j} \sum_{i,j:i\neq j} \DD_{i,j}) \Rb^{-1} \Vb \Vb^\top) + \Pi(\Vb)\\
        &= - \tr(\PMIbbhat_\CL \Vb \Vb^\top) + \Pi(\Vb).
    \end{align*}
    We can further rewrite $\mL_\CL(\Vb)$ as
    \begin{align*}
        \mL_\CL(\Vb) = \frac{\lambda}{2} \norm{\Vb \Vb^\top - \frac{1}{\lambda} \PMIbbhat_\CL}_\F^2 + (\textnormal{constant}).
    \end{align*}
\end{proof}

\section{Proof of Section~\ref{sec: theory}}

\subsection{Low-rank Approximation of PMI Matrix}\label{proof: PMI decomposition}

\begin{thm}[Restatement of Theorem~\ref{thm: PMI decomposition}]\label{thm: PMI decomposition ap}
     Under Assumptions~\ref{asm: signal covariance}, \ref{asm: regime} and \ref{asm: noise covariance}, we have     
    \begin{align}
        \max_{w,w'\in\calW^{(M)}} \Big| \PMIbb^{(M,M)}(w,w') - \qty( \frac{\v_w^{\star \top} \v_{w'}^\star}{p} + (\bSigma_M)_{w,w'} )  \Big| &\lesssim \frac{p^2}{d_M^2} \tau, \ \ M \in \{1, 2\},\label{eq: pmi decomp M max}\\
        \max_{w\in\calW^{(1)},w'\in\calW^{(2)}} \Big |\PMIbb^{(1,2)}(w,w') -  \frac{1}{p} \v_w^{\star \top} \v_{w'}^\star \Big|&\lesssim \frac{p^2}{(d_1 \wedge d_2)^2} \tau,\label{eq: pmi decomp 12 max}\\
        \Big \| \PMIbb^{(M,M)} - \qty( \frac{1}{p} \Vb_M^{\star} \Vb_M^{\star\top} + \bSigma_M ) \Big\|_\F  &\lesssim \frac{p^2}{d_M} \tau, \ \ M \in \{1, 2\},\label{eq: pmi decomp M}\\
        \Big \|\PMIbb^{(1,2)} -  \frac{1}{p} \Vb_1^{\star} \Vb_2^{\star\top} \Big\|_\F &\lesssim \frac{p^2 (d_1 \vee d_2)}{(d_1 \wedge d_2)^2} \tau,\label{eq: pmi decomp 12}
    \end{align}     
    and hence
    \begin{align*}
        \Big \| \PMIbb - \qty(\frac{1}{p} \Vb^{\star} \Vb^{\star\top} + \bSigma )  \Big\|_\F 
        &\lesssim \frac{p^2 (d_1 \vee d_2)}{(d_1 \wedge d_2)^2} \tau. 
    \end{align*}
\end{thm}

\begin{proof}[Proof of Theorem~\ref{thm: PMI decomposition ap}]
    Let $\bar \bv^{(M)}$, $\bar\sigma_w^{(M)}$ and $\bar\sigma^{(M, M')}$ be the quantities defined in Lemma~\ref{lem: A4}.
    Define
    \begin{align*}
        \bar\bsigma^{(M)} \triangleq (\bar\sigma_w^{(M)})_{w \in \calW^{(M)}} \in \R^{d_M}.
    \end{align*}
    For $M, M' \in \{1, 2\}$, $w \in \calW^{(M)}$ and $w' \in \calW^{(M')}$, we have the following result from Lemma~\ref{lem: PMI decomposition in max norm}:
    \begin{small}
    \begin{align}
            &\biggl|\PMIbb^{(M,M')}(w, w') - \biggl(\frac{\v_w^{\star \top}\v_{w'}^\star}{p} - \frac{1}{p} \v_w^{\star \top} \bar \bv^{(M')} - \frac{1}{p} \v_{w'}^{\star \top} \bar \bv^{(M)} + \frac{1}{p} \bar \bv^{(M) \top} \bar \bv^{(M')} + \sigma_{w,w'} - \bar \sigma_w^{(M')} - \bar \sigma_{w'}^{(M)} + \bar \sigma^{(M, M')}\biggr)\biggr|\nonumber\\
            &\quad\lesssim \frac{p^2}{(d_M \wedge d_{M'})^2} \tau.\label{eq: PMI pointwise decomposition}
    \end{align}
    \end{small}\noindent
    
    We bound the terms $\bar \bv^{(M)}$ and $\bar \bv^{(M')}$.
    Since $\bar \bv^{(M)} = \Vb_M^{\star \top} \bq^{(M)}$ with $\bq^{(M)}$ defined in Lemma~\ref{lem: A1} and $\Vb_M^{\star \top} \1_{d_M} = \zero$ by assumption,
    \begin{align}
        \bar \bv^{(M)} = \bar \bv^{(M)} - \frac{1}{d_M} \Vb_M^{\star \top} \1_{d_M} &= \Vb_M^{\star \top} \qty(\bq^{(M)} - \frac{1}{d_M} \1_{d_M}).\label{eq: V v bar 1}
    \end{align}
    From Lemma~\ref{lem: A1}, we have $\norm{\bq^{(M)} - (1/d_M) \1_{d_{M}}} \leq p / d_M^{3/2}$.
    Thus 
    \begin{align*}
        \norm{\bar \bv^{(M)}} &\leq \norm{\Vb_M^\star} \norm{\bq^{(M)} - \frac{1}{d_M} \1_{d_{M}}} \leq \|\bLambda_M^\star\| \frac{p}{d_M^{3/2}} \lesssim \frac{p}{d_M^{3/2}},
    \end{align*}
    where we used Assumption~\ref{asm: signal covariance} in the last inequality.
    We next bound $\bar \bsigma^{(M)}$ and $\bar \sigma^{(M, M')}$.
    Note that $\sigma_{w,w'} = \bar \sigma_w^{(M')} = \bar \sigma_{w'}^{(M)} = \bar \sigma^{(M, M')} = 0$ if $M \neq M'$, hence we assume $M=M'=1$ without loss of generality.
    Since $\bar \sigma_{w'}^{(M)} = \be_{w'}^\top \bSigma (\bq^{(M) \top}, \zero^\top)^\top$ and $\be_{w'}^\top \bSigma (\1_{d_M}^\top, \zero^\top)^\top = 0$, we have
    \begin{align*}
        \max_{w' \in \calW^{(M')}} \abs{\bar \sigma_{w'}^{(M)}} = \max_{w' \in \calW^{(M')}} \abs{\be_{w'}^\top \bSigma ( (\bq^{(M)} - (1/d_M) \1_{d_M})^\top, \zero^\top )} &\leq \frac{p}{d_M^{3/2}} \|\bSigma_{M'}\|_{2,\infty} \lesssim \frac{p^{3/2}}{d_M^{3/2} d_{M'}^{1/2}}.
    \end{align*}
    where we used Assumption~\ref{asm: noise covariance}. 
    A similar argument gives $|\bar \sigma^{(M, M)}| \lesssim p^2 / d_M^{3/2} d_{M'}^{3/2}$. 
    Thus, by \eqref{eq: PMI pointwise decomposition},
    \begin{align}
        &\max_{w\in\calW^{(M)},w'\in\calW^{(M')}} \biggl|\PMIbb^{(M,M')}(w, w') - \frac{\v_w^{\star \top}\v_{w'}^\star}{p} - \sigma_{w,w'}\biggr|\nonumber\\
        &\quad\lesssim \frac{p^2}{(d_M \wedge d_{M'})^2} \tau + \frac{1}{p} \max_{w \in \calW^{(M)}} \|\v_w^\star\| \|\bar \bv^{(M')}\| + \frac{1}{p} \max_{w' \in \calW^{(M')}} \|\v_{w'}^\star\| \|\bar \bv^{(M)}\| + \frac{1}{p} \|\bar \bv^{(M)}\| \|\bar \bv^{(M')}\|\nonumber\\
        &\quad\quad+ \max_{w \in \calW^{(M)}} |\bar \sigma_w^{(M')}| + \max_{w' \in \calW^{(M')}} |\bar \sigma_{w'}^{(M)}| + |\bar \sigma^{(M, M')}|\nonumber\\
        &\quad\lesssim \frac{p^2}{(d_M \wedge d_{M'})^2} \tau + \frac{p^{1/2}}{d_M^{1/2} d_{M'}^{3/2}} + \frac{p^{1/2}}{d_M^{3/2} d_{M'}^{1/2}} + \frac{p}{d_M^{3/2} d_{M'}^{3/2}} + \frac{p^{3/2}}{d_M^{1/2} d_{M'}^{3/2}} + \frac{p^{3/2}}{d_M^{3/2} d_{M'}^{1/2}} + \frac{p^2}{d_M^{3/2} d_{M'}^{3/2}}\nonumber\\
        &\quad\lesssim \frac{p^2}{(d_M \wedge d_{M'})^2} \tau.\label{eq: PMI decomposition in max norm rev}
    \end{align}
    This gives \eqref{eq: pmi decomp M max} and \eqref{eq: pmi decomp 12 max}.
    
    Further rewriting in matrix form combined with $\|\bA\|_\F \leq \sqrt{d_M d_{M'}} \|\bA\|_{\max}$ for any $\bA \in \R^{d_M \times d_{M'}}$ gives \eqref{eq: pmi decomp M} and \eqref{eq: pmi decomp 12}.
\end{proof}

\subsection{Estimation of Population PMI Matrix}\label{sec: PMI estimation ap}

We decompose $\PMIbbhat$ into four blocks:
\begin{align*}
    \PMIbbhat &\triangleq \begin{pmatrix}
        \PMIbbhat^{(1,1)} & \PMIbbhat^{(1,2)}\\
        \PMIbbhat^{(2,1)} & \PMIbbhat^{(2,2)}
    \end{pmatrix}.
\end{align*}
Note that the block \textit{sample} PMI matrix $\PMIbbhat^{(M,M')}$ is defined as a partial matrix of $\PMIbbhat$, while the population PMI matrix $\PMIbb$ is defined as a stacked matrix of $\PMIbb^{(1,1)}$, $\PMIbb^{(1,2)}$, $\PMIbb^{(1,2)\top}$, and $\PMIbb^{(2,2)}$,

We also define a matrix $\PMIbbhat_\CLAIME \in \R^{d_1 \times d_2}$ as
\begin{align*}
    \PMIbbhat_\CLAIME(w,w') &\triangleq \frac{\CC^{(1,1)}(\cdot, \cdot) \CC^{(2,2)}(\cdot, \cdot)}{\CC^{(1,1)}(w, \cdot) \CC^{(2,2)}(w', \cdot)} \qty(\frac{\DD^{(1,2)}(w, w')}{\DD^{(1,2)}(\cdot, \cdot)} - \frac{\sum_{i,j: i \neq j} \DD_{i,j}^{(1,2)}(w, w')}{\sum_{i,j: i \neq j} T_i^{(1)} T_j^{(2)}}).
\end{align*}

\begin{thm}\label{thm: PMI estimation ap}
    Suppose that Assumptions~\ref{asm: signal covariance}, \ref{asm: regime}, \ref{asm: noise covariance} and \ref{asm: T moments} hold.
    Fix $M, M' \in \{1, 2\}$. Then,
    \begin{align*}
        &\norm{\PMIbbhat^{(M,M)} - \PMIbb^{(M,M)} - \Bb_\PCA^{(M,M)}}_{\max}
        \lesssim \frac{d_M^2 \tau}{\sqrt{n} S_1^{(M) 1/2}} + \frac{\tau}{\sqrt{n}},\\
        &\norm{\PMIbbhat^{(1,2)} - \PMIbb^{(1,2)} - \Bb_\PCA^{(1,2)}}_{\max} \lesssim \frac{d_1 d_2 \tau}{\sqrt{n} (S_1^{(1) 1/2} \wedge S_1^{(2) 1/2})} + \frac{\tau}{\sqrt{n}},\\
        &\norm{\PMIbbhat - \PMIbb - \Bb_\PCA}_\F \lesssim \frac{(d_1 \vee d_2) d_1^2 \tau}{\sqrt{n} S_1^{(1) 1/2}} + \frac{(d_1 \vee d_2) d_2^2 \tau}{\sqrt{n} S_1^{(2) 1/2}} + \frac{(d_1 \vee d_2) d_1 d_2 \tau}{\sqrt{n} (S_1^{(1) 1/2} \wedge S_1^{(2) 1/2})} + \frac{(d_1 \vee d_2) \tau}{\sqrt{n}}
    \end{align*}
    hold with probability $1 - \exp(-\Omega(\tau^2))$. Here the matrix $\Bb_\PCA$ reflects the bias term, which is defined as
    \begin{align*}
        \Bb_\PCA \triangleq \begin{pmatrix}
            \Bb_\PCA^{(1,1)} & \Bb_\PCA^{(1,2)}\\
            \Bb_\PCA^{(1,2) \top} & \Bb_\PCA^{(2,2)}
        \end{pmatrix}
    \end{align*}
    with
    \begin{align*}
        \Bb_\PCA^{(M, M)} &= \beta_\PCA^{(M,M)} \1_{d_M} \1_{d_M}^\top,\ \ \Bb_\PCA^{(1,2)} = \beta_\PCA^{(1,2)} \1_{d_1} \1_{d_2}^\top,
    \end{align*}
    where 
    \begin{align*}
        \beta_\PCA^{(M,M)} &= \log \frac{ \qty(\sum_{i=1}^n T_i (T_i-1)) \sum_{i=1}^n T_i^{(M)} (T_i^{(M)}-1)  }{ \big ( \sum_{i=1}^n T_i^{(M)} (T_i -1) \big)^2 }\\
        \beta_\PCA^{(1,2)} &= \log \frac{ \qty(\sum_{i=1}^n T_i (T_i-1)) \sum_{i=1}^n T_i^{(1)} T_i^{(2)} }{ \qty( \sum_{i=1}^n T_i^{(1)} (T_i -1) ) \qty( \sum_{i=1}^n T_i^{(2)} (T_i -1) ) }.
    \end{align*}
\end{thm}

\begin{proof}[Proof of Theorem~\ref{thm: PMI estimation ap}]
    Note that for $w,w' \in \calW^{(M)}$,
    \begin{align*}
        \CC(w, \cdot) &= \sum_{i=1}^n (T_i - 1) \sum_{t \in [T_i]} \I\{w_{i,t} = w\},\ \ \CC(\cdot, \cdot) = \sum_{i=1}^n T_i (T_i - 1).
    \end{align*}
    This yields
    \begin{align*}
        &\log \frac{\CC(w, w') \CC(\cdot, \cdot)}{\CC(w, \cdot) \CC(w', \cdot)} - \log \frac{p_{w,w'}^{(M,M)}}{p_w^{(M)} p_{w'}^{(M)}}\\
        &= \underbrace{\log \frac{\CC^{(M,M)}(w,w')}{\sum_{i=1}^n T_i^{(M)} (T_i^{(M)}-1) p_{w,w'}^{(M,M)}}}_{\triangleq u_1}
        - \underbrace{\log \frac{\CC(w,\cdot)}{\sum_{i=1}^n  T_i^{(M)} (T_i-1) p_w^{(M)}}}_{\triangleq u_2}
        - \underbrace{\log \frac{\CC(w',\cdot)}{\sum_{i=1}^n  T_i^{(M)} (T_i-1) p_{w'}^{(M)}}}_{\triangleq u_2'}\\
        &+ \underbrace{\log \frac{ \qty(\sum_{i=1}^n T_i (T_i-1)) \qty(\sum_{i=1}^n T_i^{(M)} (T_i^{(M)}-1)) }{ \qty( \sum_{i=1}^n T_i^{(M)} (T_i -1) )^2 }}_{= (\Bb_\PCA)_{w,w'}},
    \end{align*}
    where we used $\CC(w,w') = \CC^{(M,M)}(w,w')$ for $w,w' \in \calW^{(M)}$.
    We first bound $u_1$. From Lemmas~\ref{lem: A8} and \ref{lem: A10}, 
    \begin{align*}
        \max_{w,w' \in \calW^{(M)}} |u_1| &\leq \max_{w,w' \in \calW^{(M)}} \abs{\log\qty(1 + \frac{\CC^{(M,M)}(w,w')}{\sum_{i=1}^n T_i^{(M)} (T_i^{(M)}-1) p_{i,i,w,w'}^{(M,M)}} - 1)}\\
        &\quad+ \max_{w,w' \in \calW^{(M)}}\abs{\log\qty(1 + \frac{\sum_{i=1}^n T_i^{(M)} (T_i^{(M)}-1) p_{i,i,w,w'}^{(M,M)}}{\sum_{i=1}^n T_i^{(M)} (T_i^{(M)}-1) p_{w,w'}^{(M,M)}} - 1)}\\
        &\lesssim \frac{d_M^2 \tau}{\sqrt{n} S_1^{(M) 1/2}} + \frac{\tau}{\sqrt{n}}
    \end{align*}
    holds with high probability, where we used $|\log(1 + x)| \leq |x|$ for $|x| < 1/2$ in the second inequality.
    Let $M' \in \{1, 2\} \setminus \{M\}$.
    For the term $u_2$, note that
    \begin{align*}
        &\frac{\CC(w,\cdot)}{\sum_{i=1}^n  T_i^{(M)} (T_i-1) p_w^{(M)}} - 1\\
        &\quad= \frac{\CC^{(M,M)}(w,\cdot) + \DD^{(M,M')}(w,\cdot)}{p_w^{(M)} \sum_{i=1}^n T_i^{(M)} (T_i^{(M)}-1) + p_w^{(M)} \sum_{i=1}^n  T_i^{(M)} T_i^{(M')}} - 1\\
        &\quad= \frac{\sum_{i=1}^n T_i^{(M)} (T_i^{(M)}-1)}{\sum_{i=1}^n T_i^{(M)} (T_i^{(M)}-1) + \sum_{i=1}^n  T_i^{(M)} T_i^{(M')}} \qty(\frac{\CC^{(M,M)}(w, \cdot)}{p_w^{(M)} \sum_{i=1}^n  T_i^{(M)} (T_i^{(M)} - 1)} - 1)\\
        &\quad\quad+ \frac{\sum_{i=1}^n  T_i^{(M)} T_i^{(M')}}{\sum_{i=1}^n T_i^{(M)} (T_i^{(M)}-1) + \sum_{i=1}^n  T_i^{(M)} T_i^{(M')}} \qty(\frac{\DD^{(M,M')}(w, \cdot)}{p_w^{(M)} \sum_{i=1}^n  T_i^{(M)} T_i^{(M')}} - 1),
    \end{align*}
    where we used $\CC(w, \cdot) = \CC^{(M,M)}(w, \cdot) + \DD^{(M,M')}(w, \cdot)$.
    Thus, using Lemmas~\ref{lem: A7} and \ref{lem: A9},
    \begin{align*}
        \max_{w \in \calW^{(M)}} |u_2| &= \max_{w \in \calW^{(M)}} \abs{\log(1 + \frac{\CC(w,\cdot)}{\sum_{i=1}^n  T_i^{(M)} (T_i-1) p_w^{(M)}} - 1)}\\
        &\leq \max_{w \in \calW^{(M)}} \abs{\frac{\CC^{(M,M)}(w, \cdot)}{p_w^{(M)} \sum_{i=1}^n  T_i^{(M)} (T_i^{(M)} - 1)} - 1} + \max_{w \in \calW^{(M)}} \abs{\frac{\DD^{(M,M')}(w, \cdot)}{p_w^{(M)} \sum_{i=1}^n  T_i^{(M)} T_i^{(M')}} - 1}\\
        &\leq \frac{d_M \tau}{\sqrt{n} S_1^{(M) 1/2}} + \frac{\tau}{\sqrt{n}}
    \end{align*}
    holds with high probability, where we used $|\log(1 + x)| \leq |x|$ for $|x| < 1/2$ in the second inequality. 
    Since $\max_{w \in \calW^{(M)}} |u_2| = \max_{w' \in \calW^{(M)}} |u_2'|$, we could also bound $u_2'$.
    In summary,
    \begin{align*}
        \max_{w,w' \in \calW^{(M)}} \abs{\log \frac{\CC(w, w') \CC(\cdot, \cdot)}{\CC(w, \cdot) \CC(w', \cdot)} - \log \frac{p_{w,w'}^{(M,M)}}{p_w^{(M)} p_{w'}^{(M)}} - (\Bb_\PCA)_{w,w'}} &\lesssim \frac{d_M^2 \tau}{\sqrt{n} S_1^{(M) 1/2}} + \frac{\tau}{\sqrt{n}}
    \end{align*}
    holds with high probability. This gives the first claim.

    On the other hand, for $w \in \calW^{(1)}$ and $w' \in \calW^{(2)}$,
    \begin{align*}
        &\log \frac{\CC(w, w') \CC(\cdot, \cdot)}{\CC(w, \cdot) \CC(w', \cdot)} - \log \frac{p_{w,w'}^{(1,2)}}{p_w^{(1)} p_{w'}^{(2)}}\\
        &= \underbrace{\log \frac{\DD^{(1,2)}(w,w')}{\sum_{i=1}^n T_i^{(1)} T_i^{(2)} p_{w,w'}^{(1,2)}}}_{\triangleq u_3}
        - \underbrace{\log \frac{\CC(w,\cdot)}{\sum_{i=1}^n  T_i^{(1)} (T_i-1) p_w^{(1)}}}_{\triangleq u_4}
        - \underbrace{\log \frac{\CC(w',\cdot)}{\sum_{i=1}^n  T_i^{(2)} (T_i-1) p_{w'}^{(2)}}}_{\triangleq u_4'}\\
        &+ \underbrace{\log \frac{ \qty(\sum_{i=1}^n T_i (T_i-1)) \qty(\sum_{i=1}^n T_i^{(1)} T_i^{(2)}) }{ \qty( \sum_{i=1}^n T_i^{(1)} (T_i -1) ) \qty( \sum_{i=1}^n T_i^{(2)} (T_i -1) ) }}_{=(\Bb_\PCA)_{w,w'}},
    \end{align*}
    where we used $\CC(w, w') = \DD^{(1,2)}(w,w')$ for $w \in \calW^{(1)}$ and $w' \in \calW^{(2)}$,
    From Lemmas~\ref{lem: A8} and \ref{lem: A10}, 
    \begin{align*}
        \max_{w \in \calW^{(1)}, w' \in \calW^{(2)}} |u_3| &= \abs{\log\qty(1 + \frac{\DD^{(1,2)}(w,w')}{\sum_{i=1}^n T_i^{(1)} T_i^{(2)} p_{i,i,w,w'}^{(1,2)}} - 1) + \log\qty(1 + \frac{\sum_{i=1}^n T_i^{(1)} T_i^{(2)} p_{i,i,w,w'}^{(1,2)}}{\sum_{i=1}^n T_i^{(1)} T_i^{(2)} p_{w,w'}^{(1,2)}} - 1)}\\
        &\lesssim \frac{d_1 d_2 \tau}{\sqrt{n} (S_1^{(1) 1/2} \wedge S_1^{(2) 1/2})} + \frac{\tau}{\sqrt{n}}
    \end{align*}
    holds with high probability.
    Note that $\max_{w \in \calW^{(1)}, w' \in \calW^{(2)}} (|u_4| \vee |u_4'|) = \max_{M = 1,2} \max_{w,w' \in \calW^{(M)}} (|u_2| \vee |u_2'|)$.
    In summary,
    \begin{align*}
        &\max_{w \in \calW^{(1)},w' \in \calW^{(2)}} \abs{\log \frac{\CC(w, w') \CC(\cdot, \cdot)}{\CC(w, \cdot) \CC(w', \cdot)} - \log \frac{p_{w,w'}^{(1,2)}}{p_w^{(1)} p_{w'}^{(2)}} - (\Bb_\PCA)_{w,w'}}\\
        &\quad\lesssim \frac{d_1 d_2}{\sqrt{n} (S_1^{(1) 1/2} \wedge S_1^{(2) 1/2})} \tau + \frac{d_1}{\sqrt{n} S_1^{(1) 1/2}} \tau + \frac{d_2}{\sqrt{n} S_1^{(2) 1/2}} \tau + \frac{\tau}{\sqrt{n}}\\
        &\quad\lesssim \frac{d_1 d_2 \tau}{\sqrt{n} (S_1^{(1) 1/2} \wedge S_1^{(2) 1/2})} + \frac{\tau}{\sqrt{n}}
    \end{align*}
    holds with high probability.
    This gives the second claim.
    The last claim follows directly from the first and second claims using $\|\bA\|_\F \leq d \|\bA\|_{\max}$ for $\bA \in \R^{d \times d}$.
\end{proof}

\subsection{Multimodal Contrastive Learning on Cross-PMI Matrix}
\label{sec:D4}

\begin{thm}[Restatement of Theorem~\ref{thm: PMI estimation mmcl}]\label{thm: PMI estimation mmcl ap}
    Suppose that Assumptions~\ref{asm: signal covariance}, \ref{asm: regime}, \ref{asm: noise covariance} and \ref{asm: T moments} hold. Then, 
    \begin{align*}
        \norm{\PMIbbhat_\CLAIME - \PMIbb^{(1,2)}}_{\max} &\lesssim \frac{d_1 d_2 \tau}{\sqrt{n} (S_1^{(1) 1/2} \wedge S_1^{(2) 1/2})} + \frac{\tau}{\sqrt{n}} + \frac{p^2 \tau}{(d_1 \wedge d_2)^2},\\
        \norm{\PMIbbhat_\CLAIME - \PMIbb^{(1,2)}}_\F &\lesssim \frac{d_1^{3/2} d_2^{3/2} \tau}{\sqrt{n} (S_1^{(1) 1/2} \wedge S_1^{(2) 1/2})} + \frac{d_1^{1/2} d_2^{1/2}\tau}{\sqrt{n}} + \frac{p^2 d_1^{1/2} d_2^{1/2} \tau}{(d_1 \wedge d_2)^2}
    \end{align*}
    hold with probability $1 - \exp(-\Omega(\tau^2))$.
\end{thm}

\begin{proof}[Proof of Theorem~\ref{thm: PMI estimation mmcl ap}]
    Recall that for $w \in \calW^{(1)}$ and $w' \in \calW^{(2)}$,
    \begin{align*}
        \PMIbbhat_\CLAIME(w,w') \triangleq \frac{\CC^{(1,1)}(\cdot, \cdot) \CC^{(2,2)}(\cdot, \cdot)}{\CC^{(1,1)}(w, \cdot) \CC^{(2,2)}(w', \cdot)} \qty(\frac{\DD^{(1,2)}(w, w')}{\DD^{(1,2)}(\cdot, \cdot)} - \frac{\sum_{i,j: i \neq j} \DD_{i,j}^{(1,2)}(w, w')}{\sum_{i,j: i \neq j} T_i^{(1)} T_j^{(2)}}).
    \end{align*}
    Using Lemmas~\ref{lem: A7}, \ref{lem: A8}, \ref{lem: A9} and \ref{lem: A10}, we obtain
    \begin{align*}
        &\PMIbbhat_\CLAIME(w,w')\\
        &\quad= \qty[\qty{1 + O\qty(\frac{d_1}{\sqrt{n} S_1^{(1) 1/2}} \tau)} \qty{1 + O\qty(\frac{\tau}{\sqrt{n}})} p^{(1)}_w ]^{-1}\\
        &\quad\quad\times \qty[\qty{1 + O\qty(\frac{d_2}{\sqrt{n} S_1^{(2) 1/2}} \tau)} \qty{1 + O\qty(\frac{\tau}{\sqrt{n}})} p^{(2)}_{w'} ]^{-1}\\
        &\quad\quad\times \Biggl[ \qty{1 + O\qty(\frac{d_1 d_2 \tau}{\sqrt{n} (S_2^{(1) 1/2} \wedge S_2^{(2) 1/2})})} \qty{1 + O\qty(\frac{\tau}{\sqrt{n}})} p_{w,w'}^{(1,2)}\\
        &\quad\quad\quad- \qty{1 + O\qty(\frac{d_1 d_2 \tau}{\sqrt{n} (S_2^{(1) 1/2} \wedge S_2^{(2) 1/2})})} \qty{1 + O\qty(\frac{\tau}{\sqrt{n}})} p_w^{(1)} p_{w'}^{(2)}\Biggr].
    \end{align*}
    Thus,
    \begin{small}
    \begin{align*}
        &\PMIbbhat_\CLAIME(w,w')\\
        &\quad= \qty{1 + O\qty(\frac{d_1}{\sqrt{n} S_1^{(1) 1/2}} \tau + \frac{d_2}{\sqrt{n} S_1^{(2) 1/2}} \tau + \frac{\tau}{\sqrt{n}})}\\
        &\quad\quad\times \Biggl[ \qty{1 + O\qty(\frac{d_1 d_2 \tau}{\sqrt{n} (S_1^{(1) 1/2} \wedge S_1^{(2) 1/2})} + \frac{\tau}{\sqrt{n}})} \frac{p_{w,w'}^{(1,2)}}{p_w^{(1)} p_{w'}^{(2)}} - \qty{1 + O\qty(\frac{d_1 d_2 \tau}{\sqrt{n} (S_1^{(1) 1/2} \wedge S_1^{(2) 1/2})} + \frac{\tau}{\sqrt{n}})} \Biggr]
    \end{align*}
    \end{small}\noindent
    holds with high probability. Here the Big O notation does not depend on $w \in \calW^{(1)}$ and $w' \in \calW^{(2)}$.
    Using $1/(d_1 d_2) \lesssim \min_{w \in \calW^{(1)},w' \in \calW^{(2)}} p_{w,w'}^{(1,2)} \leq \max_{w \in \calW^{(1)},w' \in \calW^{(2)}} p_{w,w'}^{(1,2)} \lesssim 1/(d_1 d_2)$ and $1/d_M \lesssim \min_{w \in \calW^{(M)}} p_w^{(M)} \leq \max_{w \in \calW^{(M)}} p_w^{(M)} \lesssim 1/d_M$ from Lemmas~ \ref{lem: A4} and \ref{lem: A5}, we obtain
    \begin{align*}
        \PMIbbhat_\CLAIME(w,w') &= \frac{p_{w,w'}^{(1,2)}}{p_w^{(1)} p_{w'}^{(2)}} - 1 + O\qty(\frac{d_1 d_2 \tau}{\sqrt{n} (S_1^{(1) 1/2} \wedge S_1^{(2) 1/2})} + \frac{\tau}{\sqrt{n}}).
    \end{align*}
    From \eqref{eq: PMI decomposition in max norm rev} combined with $|\log(1+x) - x| \leq |x|$ for $|x| < 1/2$, 
    \begin{align*}
        \max_{w\in\calW^{(1)}, w'\in\calW^{(2)}} \abs{\log\frac{p_{w,w'}^{(1,2)}}{p_w^{(1)} p_{w'}^{(2)}} - \qty(\frac{p_{w,w'}^{(1,2)}}{p_w^{(1)} p_{w'}^{(2)}} - 1)} \lesssim \frac{p^2 \tau}{(d_1 \wedge d_2)^2}.
    \end{align*}
    Therefore,
    \begin{align}
        \max_{w\in\calW^{(1)}, w'\in\calW^{(2)}} \abs{\PMIbbhat_\CLAIME(w,w') - \PMIbb(w,w')} \lesssim \frac{d_1 d_2 \tau}{\sqrt{n} (S_1^{(1) 1/2} \wedge S_1^{(2) 1/2})} + \frac{\tau}{\sqrt{n}} + \frac{p^2 \tau}{(d_1 \wedge d_2)^2}.\label{eq: PMI mmcl 11}
    \end{align}
    The bound for $\|\PMIbbhat_\CLAIME - \PMIbb\|_\F$ follows from the fact that $\|\bA\|_\F \leq \sqrt{d_1 d_2} \|\bA\|_{\max}$ for $\bA \in \R^{d_1 \times d_2}$.
\end{proof}

\begin{thm}[Proof of Theorem~\ref{thm: MMCL convergence}]\label{thm: MMCL convergence ap}
    Suppose that Assumptions~\ref{asm: signal covariance}, \ref{asm: regime}, \ref{asm: noise covariance} and \ref{asm: T moments} hold.
    Let $(\Vbhat_1, \Vbhat_2)$ be the minimizer of the loss \ref{loss: MMCL linear}.
    Then,
    \begin{align*}
        &\|\sin\Theta(\Pb_p(\Vbhat_1^\top), \Pb_p(\Vb^{\star \top}_1))\|_\F \vee \|\sin\Theta(\Pb_p(\Vbhat_2^\top), \Pb_p(\Vb^{\star \top}_2))\|_\F\\
        &\quad\lesssim \frac{p^3 (d_1 \vee d_2) \tau}{(d_1 \wedge d_2)^2} + \frac{p d_1^{3/2} d_2^{3/2} \tau}{\sqrt{n} (S_1^{(1) 1/2} \wedge S_1^{(2) 1/2})} + \frac{p d_1^{1/2} d_2^{1/2}\tau}{\sqrt{n}},\\
        &\min_{\Hb \in \mathcal{O}_{p,p}} \|\sin\Theta(\Ubhat_\Hb, \Ub^\star)\|_\F\\
        &\quad\lesssim \frac{p^3 (d_1 \vee d_2) \tau}{(d_1 \wedge d_2)^2} + \frac{p d_1^{3/2} d_2^{3/2} \tau}{\sqrt{n} (S_1^{(1) 1/2} \wedge S_1^{(2) 1/2})} + \frac{p d_1^{1/2} d_2^{1/2}\tau}{\sqrt{n}}
    \end{align*}
    hold with probability $1 - \exp(-\Omega(\tau^2))$.
\end{thm}

\begin{proof}[Proof of Theorem~\ref{thm: MMCL convergence ap}]
    From Proposition~\ref{prop: SVD ap}, we have that
    \begin{align*}
        \mL_\CLAIME(\Vb_1, \Vb_2) &= \frac{\lambda}{2} \Big\| \Vb_1 \Vb_2^\top - \frac{1}{\lambda} \PMIbbhat_\CLAIME\Big\|_\F^2 + (\textnormal{constant}).
    \end{align*}
    Thus, by Eckart-Young-Mirsky theorem, minimizing $\mL_\CLAIME(\Vb_1, \Vb_2)$ yields a global minimizer $(\Vbhat_1, \Vbhat_2)$ satisfying
    \begin{align*}
        \Vbhat_1 \Vbhat_2^\top = \frac{1}{\lambda} \SVD_p(\PMIbbhat_\CLAIME),
    \end{align*}
    and thus $\Pb_p(\Vbhat_1^\top) = \Pb_p(\PMIbbhat_\CLAIME^\top)$ and $\Pb_p(\Vbhat_2^\top) = \Pb_p(\PMIbbhat_\CLAIME)$.
    Using Theorems~\ref{thm: PMI decomposition ap} and \ref{thm: PMI estimation mmcl ap}, we obtain
    \begin{align*}
        \norm{\PMIbbhat_\CLAIME - \frac{1}{p} \Vb_1^\star \Vb_2^{\star \top}}_\F &\lesssim \frac{p^2 (d_1 \vee d_2) \tau}{(d_1 \wedge d_2)^2} + \frac{d_1^{3/2} d_2^{3/2} \tau}{\sqrt{n} (S_1^{(1) 1/2} \wedge S_1^{(2) 1/2})} + \frac{d_1^{1/2} d_2^{1/2}\tau}{\sqrt{n}}
    \end{align*}
    with high probability. Note that Assumption~\ref{asm: regime} gives $\norm{\PMIbbhat_\CLAIME - \frac{1}{p} \Vb_1^\star \Vb_2^{\star \top}} \ll 1/p$.
    From Assumption~\ref{asm: signal covariance}, we have $1/p \lesssim s_p(\Vb_1^\star \Vb_2^{\star \top}/p)$.
    Using Theorem 4 from \cite{yu2015useful} and Assumption~\ref{asm: signal covariance}, we can further obtain
    \begin{align*}
        &\|\sin\Theta(\Pb_p(\Vbhat_1^\top), \Ub^{\star \top}_1)\|_\F \vee \|\sin\Theta(\Pb_p(\Vbhat_2^\top), \Ub^{\star \top}_2)\|_\F\\
        &\quad\lesssim \frac{1/p}{(1/p)^2} \norm{\PMIbbhat_\CLAIME - \frac{1}{p} \Vb_1^\star \Vb_2^{\star \top}}_\F\\
        &\quad\lesssim \frac{p^3 (d_1 \vee d_2) \tau}{(d_1 \wedge d_2)^2} + \frac{p d_1^{3/2} d_2^{3/2} \tau}{\sqrt{n} (S_1^{(1) 1/2} \wedge S_1^{(2) 1/2})} + \frac{p d_1^{1/2} d_2^{1/2}\tau}{\sqrt{n}}
    \end{align*}
    with high probability. 
    We next bound $\min_{\Hb \in \mathcal{O}_{p,p}} \|\sin\Theta(\Ubhat_\Hb, \Ub^\star)\|_\F$ from above.
    Let $\Hb_1, \Hb_2 \in \mathcal{O}_{p \times p}$ be the orthogonal matrices satisfying $\|\Pb_p(\Vbhat_1^\top) \Hb_1 - \Ub_1^\star\|_\F = \|\sin\Theta(\Pb_p(\Vbhat_1^\top), \Ub_1^\star)\|_\F$ and $\|\Pb_p(\Vbhat_2^\top) \Hb_2 - \Ub_2^\star\|_\F = \|\sin\Theta(\Pb_p(\Vbhat_2^\top), \Ub_2^\star)\|_\F$. 
    Note that for the choice of $\Hb \leftarrow \Hb' := \Hb_2 \Hb_1^\top$, it follows that
    \begin{align*}
        \Ubhat_{\Hb'} = \Pb_p\qty( [\Hb \Pb_p^\top(\Vbhat_1^\top), \Pb_p^\top(\Vbhat_2^\top)]) = \Pb_p\qty( \Hb_2^\top [\Hb \Pb_p^\top(\Vbhat_1^\top), \Pb_p^\top(\Vbhat_2^\top)]) = \Pb_p\qty( [\Hb_1^\top \Pb_p^\top(\Vbhat_1^\top), \Hb_2^\top \Pb_p^\top(\Vbhat_2^\top)]),
    \end{align*}
    since $\Hb_2 \in \mathcal{O}_{p,p}$. Using $\Ub^\star = (1/\sqrt{2}) [\Ub_1^{\star \top}; \Ub_2^{\star \top}]^\top$ combined with Davis-Kahan theorem, we have
    \begin{align*}
        \|\sin\Theta(\Ubhat_{\Hb'}, \Ub^\star)\|_\F \lesssim \norm{\begin{pmatrix}
            \Pb_p(\Vbhat_1^\top) \Hb_1 - \Ub_1^\star\\
            \Pb_p(\Vbhat_2^\top) \Hb_2 - \Ub_2^\star
        \end{pmatrix}}_\F \leq \|\sin\Theta(\Pb_p(\Vbhat_1^\top), \Ub_1^\star)\|_\F + \|\sin\Theta(\Pb_p(\Vbhat_2^\top), \Ub_2^\star)\|_\F
    \end{align*}
    by definition of $\Hb_1$ and $\Hb_2$.
    This concludes the proof.
\end{proof}

\subsection{Contrastive Learning on Joint PMI Matrix}\label{sec: CL ap}

\begin{thm}[Restatement of Theorem~\ref{thm: PMI estimation cl}: Contrastive Learning Part]\label{thm: PMI estimation cl ap}
    Suppose that Assumptions~\ref{asm: signal covariance}, \ref{asm: regime}, \ref{asm: noise covariance} and \ref{asm: T moments} hold. Then,
    \begin{align*}
        &\norm{\PMIbbhat_\CL - (\PMIbb + \Bb_\CL)}_{\max}\\
        &\quad\lesssim \frac{d_1^2 \tau}{\sqrt{n} S_1^{(1) 1/2}} + \frac{d_2^2 \tau}{\sqrt{n} S_1^{(2) 1/2}} + \frac{d_1 d_2 \tau}{\sqrt{n} (S_1^{(1) 1/2} \wedge S_1^{(2) 1/2})} + \frac{\tau}{\sqrt{n}} + \frac{p^2 \tau}{(d_1 \wedge d_2)^2},\\
        &\norm{\PMIbbhat_\CL - (\PMIbb + \Bb_\CL)}_\F\\
        &\quad\lesssim \frac{(d_1 \vee d_2) d_1^2 \tau}{\sqrt{n} S_1^{(1) 1/2}} + \frac{(d_1 \vee d_2) d_2^2 \tau}{\sqrt{n} S_1^{(2) 1/2}} + \frac{(d_1 \vee d_2) d_1 d_2 \tau}{\sqrt{n} (S_1^{(1) 1/2} \wedge S_1^{(2) 1/2})} + \frac{\tau (d_1 \vee d_2)}{\sqrt{n}} + \frac{p^2 (d_1 \vee d_2) \tau}{(d_1 \wedge d_2)^2},
    \end{align*}
    hold with probability $1 - \exp(-\Omega(\tau^2))$. Here the matrix $\Bb_\CL$ reflects the bias term, which is defined as
    \begin{align*}
        \Bb_\CL \triangleq \begin{pmatrix}
            \Bb_\CL^{(1,1)} & \Bb_\CL^{(1,2)}\\
            \Bb_\CL^{(1,2) \top} & \Bb_\CL^{(2,2)}
        \end{pmatrix},
    \end{align*}
    with
    \begin{align*}
        \Bb_\CL^{(M,M)} &= \beta_{\CL,1}^{(M,M)} \1_{d_M} \1_{d_M}^\top + \beta_{\CL,2}^{(M,M)} (\PMIbb^{(M,M)} + \beta_{\CL,1}^{(M,M)} \1_{d_M} \1_{d_M}^\top),\\
        \Bb_\CL^{(1,2)} &= \beta_{\CL,1}^{(1,2)} \1_{d_1} \1_{d_2}^\top + \beta_{\CL,2}^{(1,2)} (\PMIbb^{(1,2)} + \beta_{\CL,1}^{(1,2)} \1_{d_1} \1_{d_2}^\top),
    \end{align*}
    where 
    \begin{align*}
        \beta_{\CL,1}^{(M,M)} &= 1 - \frac{\sum_{i \in [n]} T_i (T_i - 1)}{\sum_{i \in [n]} T_i^{(M)} (T_i^{(M)} - 1)} \frac{\sum_{i,j: i \neq j} T_i^{(M)} T_j^{(M)}}{\sum_{i,j: i \neq j} T_i T_j},\\
        \beta_{\CL,1}^{(1,2)} &= 1 - \frac{\sum_{i \in [n]} T_i (T_i - 1)}{\sum_{i \in [n]} T_i^{(1)} T_i^{(2)}} \frac{\sum_{i,j: i \neq j} T_i^{(1)} T_j^{(2)}}{\sum_{i,j: i \neq j} T_i T_j},\\
        \beta_{\CL,2}^{(M,M)} &= \qty(\frac{\sum_{i \in [n]} T_i^{(M)} (T_i - 1)}{\sum_{i \in [n]} T_i (T_i - 1)})^{-2} \frac{\sum_{i \in [n]} T_i^{(M)} (T_i^{(M)} - 1)}{\sum_{i \in [n]} T_i (T_i - 1)} - 1,\\
        \beta_{\CL,2}^{(1,2)} &= \qty{\frac{\sum_{i \in [n]} T_i^{(1)} (T_i - 1)}{\sum_{i \in [n]} T_i (T_i - 1)}}^{-1} \qty{\frac{\sum_{i \in [n]} T_i^{(2)} (T_i - 1)}{\sum_{i \in [n]} T_i (T_i - 1)}}^{-1} \frac{\sum_{i \in [n]} T_i^{(1)} T_i^{(2)}}{\sum_{i \in [n]} T_i (T_i - 1)} - 1.
    \end{align*}
\end{thm}

\begin{proof}[Proof of Theorem~\ref{thm: PMI estimation cl ap}]
    Recall that
    \begin{align*}
        \PMIbbhat_\CL(w,w') \triangleq \frac{\CC(\cdot, \cdot) \CC(\cdot, \cdot)}{\CC(w, \cdot) \CC(w', \cdot)} \qty(\frac{\CC(w, w')}{\CC(\cdot, \cdot)} - \frac{\sum_{i,j: i \neq j} \DD_{i,j}(w, w')}{\sum_{i,j: i \neq j} T_i T_j}).
    \end{align*}
    We consider $w, w' \in \calW^{(M)}$ first. Let $M = 1$ without loss of generality.
    We first deal with the term inside parenthesis.
    Since $\CC(w, w') = \CC^{(1,1)}(w, w')$, $\CC(w, \cdot) = \CC^{(1,1)}(w, \cdot) + \DD^{(1,2)}(w, \cdot)$ and $\DD_{i,j}(w, w') = \DD_{i,j}^{(1,1)}(w,w')$ hold for any $i, j \in [n]$, we have
    \begin{small}
    \begin{align*}
        &\frac{\CC(w, w')}{\CC(\cdot, \cdot)} - \frac{\sum_{i,j: i \neq j} \DD_{i,j}(w, w')}{\sum_{i,j: i \neq j} T_i T_j}\\
        &\quad= \frac{\sum_{i \in [n]} T_i^{(1)} (T_i^{(1)} - 1)}{\sum_{i \in [n]} T_i (T_i - 1)} \frac{\CC^{(1,1)}(w, w')}{p_{w,w'}^{(1,1)} \sum_{i \in [n]} T_i^{(1)} (T_i^{(1)} - 1)} p_{w,w'}^{(1,1)} - \frac{\sum_{i,j: i \neq j} T_i^{(1)} T_j^{(1)}}{\sum_{i,j: i \neq j} T_i T_j} \frac{\sum_{i,j: i \neq j} \DD_{i,j}^{(1,1)}(w,w')}{p_w^{(1)} p_{w'}^{(1)} \sum_{i,j: i \neq j} T_i^{(1)} T_j^{(1)}} p_w^{(1)} p_{w'}^{(1)}\\
        &\quad= \frac{\sum_{i \in [n]} T_i^{(1)} (T_i^{(1)} - 1)}{\sum_{i \in [n]} T_i (T_i - 1)} p_w^{(1)} p_{w'}^{(1)} \qty(\frac{\CC^{(1,1)}(w, w')}{p_{w,w'}^{(1,1)} \sum_{i \in [n]} T_i^{(1)} (T_i^{(1)} - 1)} \frac{p_{w,w'}^{(1,1)}}{p_w^{(1)} p_{w'}^{(1)}} - (1 - \beta_{\CL,1}^{(1,1)}) \frac{\sum_{i,j: i \neq j} \DD_{i,j}^{(1,1)}(w,w')}{p_w^{(1)} p_{w'}^{(1)} \sum_{i,j: i \neq j} T_i^{(1)} T_j^{(1)}}),
    \end{align*}
    \end{small}\noindent
    where $\beta_{\CL,1}^{(M,M)} \triangleq 1 - \frac{\sum_{i \in [n]} T_i (T_i - 1)}{\sum_{i \in [n]} T_i^{(M)} (T_i^{(M)} - 1)} \frac{\sum_{i,j: i \neq j} T_i^{(M)} T_j^{(M)}}{\sum_{i,j: i \neq j} T_i T_j}$ for $M \in \{1, 2\}$.
    Thus, using Lemmas~\ref{lem: A8} and \ref{lem: A10}, we obtain
    \begin{align*}
        &\frac{\CC(w, w')}{\CC(\cdot, \cdot)} - \frac{\sum_{i,j: i \neq j} \DD_{i,j}(w, w')}{\sum_{i,j: i \neq j} T_i T_j}\\
        &\quad= \frac{\sum_{i \in [n]} T_i^{(1)} (T_i^{(1)} - 1)}{\sum_{i \in [n]} T_i (T_i - 1)} p_w^{(1)} p_{w'}^{(1)} \Biggl\{ \qty(1 + O\qty(\frac{d_1^2}{\sqrt{n} S_1^{(1) 1/2}} \tau + \frac{\tau}{\sqrt{n}})) \frac{p_{w,w'}^{(1,1)}}{p_w^{(1)} p_{w'}^{(1)}}\\
        &\quad\quad- (1 - \beta_{\CL,1}^{(1,1)}) \qty(1 + O\qty(\frac{d_1^2 \tau}{\sqrt{n} S_1^{(1) 1/2}} + \frac{\tau}{\sqrt{n}})) \Biggr\}\\
        &\quad= \frac{\sum_{i \in [n]} T_i^{(1)} (T_i^{(1)} - 1)}{\sum_{i \in [n]} T_i (T_i - 1)} p_w^{(1)} p_{w'}^{(1)}\\
        &\quad\quad\times \Biggl\{ \qty(1 + O\qty(\frac{d_1^2 \tau}{\sqrt{n} S_1^{(1) 1/2}} + \frac{\tau}{\sqrt{n}})) \frac{p_{w,w'}^{(1,1)}}{p_w^{(1)} p_{w'}^{(1)}} - (1 - \beta_{\CL,1}^{(1,1)}) \qty(1 + O\qty(\frac{d_1^2 \tau}{\sqrt{n} S_1^{(1) 1/2}} + \frac{\tau}{\sqrt{n}})) \Biggr\}
    \end{align*}
    with high probability.

    Using $1/d_1^2 \lesssim \min_{w,w' \in \calW^{(1)}} p_{w,w'}^{(1,1)} \leq \max_{w,w' \in \calW^{(1)}} p_{w,w'}^{(1,1)} \lesssim 1/d_1^2$ and $1/d_1 \lesssim \min_{w \in \calW^{(1)}} p_w^{(1)} \leq \max_{w \in \calW^{(1)}} p_w^{(1)} \lesssim 1/d_1$ from Theorem~\ref{thm: PMI decomposition ap}, and $\beta_{\CL,1}^{(1,1)} \lesssim 1$ from Lemma~\ref{lem: bias ap}, we obtain
    \begin{align*}
        &\frac{\CC(w, w')}{\CC(\cdot, \cdot)} - \frac{\sum_{i,j: i \neq j} \DD_{i,j}(w, w')}{\sum_{i,j: i \neq j} T_i T_j}\\
        &\quad= \frac{\sum_{i \in [n]} T_i^{(1)} (T_i^{(1)} - 1)}{\sum_{i \in [n]} T_i (T_i - 1)} p_w^{(1)} p_{w'}^{(1)} \Biggl\{ \frac{p_{w,w'}^{(1,1)}}{p_w^{(1)} p_{w'}^{(1)}} - 1 + \beta_{\CL,1}^{(1,1)} +  O\qty(\frac{d_1^2 \tau}{\sqrt{n} S_1^{(1) 1/2}} + \frac{\tau}{\sqrt{n}}) \Biggr\}
    \end{align*}
    with high probability.

    We next deal with the term $\CC(w,\cdot) / \CC(\cdot, \cdot)$.
    Note that
    \begin{align*}
        \frac{\CC(w, \cdot)}{\CC(\cdot, \cdot)} &= \frac{\CC^{(1,1)}(w, \cdot) + \DD^{(1,2)}(w, \cdot)}{\sum_{i \in [n]} T_i (T_i - 1)}\\
        &= \frac{\sum_{i \in [n]} T_i^{(1)} (T_i^{(1)} - 1)}{\sum_{i \in [n]} T_i (T_i - 1)} \frac{\CC^{(1,1)}(w, \cdot)}{p_w^{(1)} \sum_{i \in [n]} T_i^{(1)} (T_i^{(1)} - 1)} p_w^{(1)}\\
        &\quad+ \frac{\sum_{i \in [n]} T_i^{(1)} T_i^{(2)}}{\sum_{i \in [n]} T_i (T_i - 1)} \frac{\DD^{(1,2)}(w, \cdot)}{p_w^{(1)} \sum_{i \in [n]} T_i^{(1)} T_i^{(2)}} p_w^{(1)}.
    \end{align*}
    Using Lemmas~\ref{lem: A7} and \ref{lem: A9}, we obtain
    \begin{align}
        \frac{\CC(w, \cdot)}{\CC(\cdot, \cdot)} &= \frac{\sum_{i \in [n]} T_i^{(1)} (T_i^{(1)} - 1)}{\sum_{i \in [n]} T_i (T_i - 1)} p_w^{(1)} \qty(1 + O\qty(\frac{d_1 \tau}{\sqrt{n} S_1^{(1) 1/2}} + \frac{\tau}{\sqrt{n}}))\nonumber\\
        &\quad+ \frac{\sum_{i \in [n]} T_i^{(1)} T_i^{(2)}}{\sum_{i \in [n]} T_i (T_i - 1)} p_w^{(1)} \qty(1 + O\qty(\frac{d_1 \tau}{\sqrt{n} S_1^{(1) 1/2}} + \frac{\tau}{\sqrt{n}}))\nonumber\\
        &= \frac{\sum_{i \in [n]} T_i^{(1)} (T_i - 1)}{\sum_{i \in [n]} T_i (T_i - 1)} p_w^{(1)} \qty(1 + O\qty(\frac{d_1 \tau}{\sqrt{n} S_1^{(1) 1/2}} + \frac{\tau}{\sqrt{n}}))\label{eq: C w rate}
    \end{align}
    with high probability. We can similarly obtain the expansion for $\CC(w', \cdot) / \CC(\cdot, \cdot)$. Thus
    \begin{align*}
        &\PMIbbhat_\CL(w,w')\\
        &\quad= \qty[\frac{\sum_{i \in [n]} T_i^{(1)} (T_i - 1)}{\sum_{i \in [n]} T_i (T_i - 1)} p_w^{(1)} \qty(1 + O\qty(\frac{d_1 \tau}{\sqrt{n} S_1^{(1) 1/2}} + \frac{\tau}{\sqrt{n}}))]^{-1}\\
        &\quad\quad\times \qty[\frac{\sum_{i \in [n]} T_i^{(1)} (T_i - 1)}{\sum_{i \in [n]} T_i (T_i - 1)} p_{w'}^{(1)} \qty(1 + O\qty(\frac{d_1 \tau}{\sqrt{n} S_1^{(1) 1/2}} + \frac{\tau}{\sqrt{n}}))]^{-1}\\
        &\quad\quad\times \frac{\sum_{i \in [n]} T_i^{(1)} (T_i^{(1)} - 1)}{\sum_{i \in [n]} T_i (T_i - 1)} p_w^{(1)} p_{w'}^{(1)} \Biggl\{ \frac{p_{w,w'}^{(1,1)}}{p_w^{(1)} p_{w'}^{(1)}} - 1 + \beta_{\CL,1}^{(1,1)} +  O\qty(\frac{d_1^2 \tau}{\sqrt{n} S_1^{(1) 1/2}} + \frac{\tau}{\sqrt{n}}) \Biggr\}\\
        &\quad= \qty{\frac{\sum_{i \in [n]} T_i^{(1)} (T_i - 1)}{\sum_{i \in [n]} T_i (T_i - 1)}}^{-2} \frac{\sum_{i \in [n]} T_i^{(1)} (T_i^{(1)} - 1)}{\sum_{i \in [n]} T_i (T_i - 1)} \Biggl\{ \frac{p_{w,w'}^{(1,1)}}{p_w^{(1)} p_{w'}^{(1)}} - 1 + \beta_{\CL,1}^{(1,1)} +  O\qty(\frac{d_1^2 \tau}{\sqrt{n} S_1^{(1) 1/2}} + \frac{\tau}{\sqrt{n}}) \Biggr\}\\
        &\quad= (1 + \beta_{\CL,2}^{(1,1)}) \Biggl\{ \frac{p_{w,w'}^{(1,1)}}{p_w^{(1)} p_{w'}^{(1)}} - 1 + \beta_{\CL,1}^{(1,1)} +  O\qty(\frac{d_1^2 \tau}{\sqrt{n} S_1^{(1) 1/2}} + \frac{\tau}{\sqrt{n}}) \Biggr\}
    \end{align*}
    holds with high probability, where $\beta_{\CL,2}^{(M,M)} \triangleq \qty(\frac{\sum_{i \in [n]} T_i^{(M)} (T_i - 1)}{\sum_{i \in [n]} T_i (T_i - 1)})^{-2} \frac{\sum_{i \in [n]} T_i^{(M)} (T_i^{(M)} - 1)}{\sum_{i \in [n]} T_i (T_i - 1)} - 1$ and the Big O notation does not depend on $w$ and $w'$.
    Using $\beta_{\CL,2}^{(1,1)} \lesssim 1$ from Lemma~\ref{lem: bias ap}, we obtain
    \begin{align}
        \PMIbbhat_\CL(w,w') = \frac{p_{w,w'}^{(1,1)}}{p_w^{(1)} p_{w'}^{(1)}} - 1 + \beta_{\CL,1}^{(1,1)} + \beta_{\CL,2}^{(1,1)} \qty(\frac{p_{w,w'}^{(1,1)}}{p_w^{(1)} p_{w'}^{(1)}} - 1 + \beta_{\CL,1}^{(1,1)}) + O\qty(\frac{d_1^2 \tau}{\sqrt{n} S_1^{(1) 1/2}} + \frac{\tau}{\sqrt{n}})\label{eq: PMI CL 11}
    \end{align}
    with high probability.
    A symmetric argument yields that for $w, w' \in \calW^{(2)}$,
    \begin{align}
        \PMIbbhat_\CL(w,w') = \frac{p_{w,w'}^{(2,2)}}{p_w^{(2)} p_{w'}^{(2)}} - 1 + \beta_{\CL,1}^{(2,2)} + \beta_{\CL,2}^{(2,2)} \qty(\frac{p_{w,w'}^{(2,2)}}{p_w^{(2)} p_{w'}^{(2)}} - 1 + \beta_{\CL,1}^{(2,2)}) + O\qty(\frac{d_2^2 \tau}{\sqrt{n} S_1^{(2) 1/2}} + \frac{\tau}{\sqrt{n}})\label{eq: PMI CL 22}
    \end{align}
    holds with high probability.
    
    For $w \in \calW^{(1)}$ and $w' \in \calW^{(2)}$, a similar argument combined with Lemma~\ref{lem: A10} and $\beta_{\CL,1}^{(1,2)} \lesssim 1$ from Lemma~\ref{lem: bias ap} yields
    \begin{small}
    \begin{align*}
        &\frac{\CC(w, w')}{\CC(\cdot, \cdot)} - \frac{\sum_{i,j: i \neq j} \DD_{i,j}(w, w')}{\sum_{i,j: i \neq j} T_i T_j}\\
        &\quad= \frac{\sum_{i \in [n]} T_i^{(1)} T_i^{(2)}}{\sum_{i \in [n]} T_i (T_i - 1)} \frac{\DD^{(1,2)}(w, w')}{p_{w,w'}^{(1,2)} \sum_{i \in [n]} T_i^{(1)} T_i^{(2)}} p_{w,w'}^{(1,2)} - \frac{\sum_{i,j: i \neq j} T_i^{(1)} T_j^{(2)}}{\sum_{i,j: i \neq j} T_i T_j} \frac{\sum_{i,j: i \neq j} \DD_{i,j}^{(1,2)}(w,w')}{p_w^{(1)} p_{w'}^{(2)} \sum_{i,j: i \neq j} T_i^{(1)} T_j^{(2)}} p_w^{(1)} p_{w'}^{(2)}\\
        &\quad= \frac{\sum_{i \in [n]} T_i^{(1)} T_i^{(2)}}{\sum_{i \in [n]} T_i (T_i - 1)} p_w^{(1)} p_{w'}^{(2)} \qty(\frac{\DD^{(1,2)}(w, w')}{p_{w,w'}^{(1,2)} \sum_{i \in [n]} T_i^{(1)} T_i^{(2)}} \frac{p_{w,w'}^{(1,2)}}{p_w^{(1)} p_{w'}^{(2)}} - (1 - \beta_{\CL,1}^{(1,2)}) \frac{\sum_{i,j: i \neq j} \DD_{i,j}^{(1,2)}(w,w')}{p_w^{(1)} p_{w'}^{(2)} \sum_{i,j: i \neq j} T_i^{(1)} T_j^{(2)}})\\
        &\quad= \frac{\sum_{i \in [n]} T_i^{(1)} T_i^{(2)}}{\sum_{i \in [n]} T_i (T_i - 1)} p_w^{(1)} p_{w'}^{(2)}\\
        &\quad\quad\times \Biggl\{ \qty(1 + O\qty(\frac{d_1 d_2 \tau}{\sqrt{n} (S_1^{(1) 1/2} \wedge S_1^{(2) 1/2})} + \frac{\tau}{\sqrt{n}})) \frac{p_{w,w'}^{(1,2)}}{p_w^{(1)} p_{w'}^{(2)}}\\
        &\quad\quad\quad- (1 - \beta_{\CL,1}^{(1,2)}) \qty(1 + O\qty(\frac{d_1 d_2 \tau}{\sqrt{n} (S_1^{(1) 1/2} \wedge S_1^{(2) 1/2})} + \frac{\tau}{\sqrt{n}})) \Biggr\}\\
        &\quad= \frac{\sum_{i \in [n]} T_i^{(1)} T_i^{(2)}}{\sum_{i \in [n]} T_i (T_i - 1)} p_w^{(1)} p_{w'}^{(2)} \Biggl\{ \frac{p_{w,w'}^{(1,2)}}{p_w^{(1)} p_{w'}^{(2)}} - 1 + \beta_{\CL,1}^{(1,2)} +  O\qty(\frac{d_1 d_2 \tau}{\sqrt{n} (S_1^{(1) 1/2} \wedge S_1^{(2) 1/2})} + \frac{\tau}{\sqrt{n}}) \Biggr\},
    \end{align*}
    \end{small}\noindent
    where $\beta_{\CL,1}^{(1,2)} \triangleq 1 - \frac{\sum_{i \in [n]} T_i (T_i - 1)}{\sum_{i \in [n]} T_i^{(1)} T_i^{(2)}} \frac{\sum_{i,j: i \neq j} T_i^{(1)} T_j^{(2)}}{\sum_{i,j: i \neq j} T_i T_j}$. Therefore, using \eqref{eq: C w rate}, we have
    \begin{small}
    \begin{align*}
        \PMIbbhat_\CL(w,w') &= \qty[\frac{\sum_{i \in [n]} T_i^{(1)} (T_i - 1)}{\sum_{i \in [n]} T_i (T_i - 1)} p_w^{(1)} \qty(1 + O\qty(\frac{d_1 \tau}{\sqrt{n} S_1^{(1) 1/2}} + \frac{\tau}{\sqrt{n}}))]^{-1}\\
        &\quad\times \qty[\frac{\sum_{i \in [n]} T_i^{(2)} (T_i - 1)}{\sum_{i \in [n]} T_i (T_i - 1)} p_{w'}^{(2)} \qty(1 + O\qty(\frac{d_2 \tau}{\sqrt{n} S_1^{(2) 1/2}} + \frac{\tau}{\sqrt{n}}))]^{-1}\\
        &\quad\times \frac{\sum_{i \in [n]} T_i^{(1)} T_i^{(2)}}{\sum_{i \in [n]} T_i (T_i - 1)} p_w^{(1)} p_{w'}^{(2)} \Biggl\{ \frac{p_{w,w'}^{(1,2)}}{p_w^{(1)} p_{w'}^{(2)}} - 1 + \beta_{\CL,1}^{(1,2)} +  O\qty(\frac{d_1 d_2 \tau}{\sqrt{n} (S_1^{(1) 1/2} \wedge S_1^{(2) 1/2})} + \frac{\tau}{\sqrt{n}}) \Biggr\}\\
        &= \qty{\frac{\sum_{i \in [n]} T_i^{(1)} (T_i - 1)}{\sum_{i \in [n]} T_i (T_i - 1)}}^{-1} \qty{\frac{\sum_{i \in [n]} T_i^{(2)} (T_i - 1)}{\sum_{i \in [n]} T_i (T_i - 1)}}^{-1} \frac{\sum_{i \in [n]} T_i^{(1)} T_i^{(2)}}{\sum_{i \in [n]} T_i (T_i - 1)}\\
        &\quad\times \Biggl\{ \frac{p_{w,w'}^{(1,2)}}{p_w^{(1)} p_{w'}^{(2)}} - 1 + \beta_{\CL,1}^{(1,2)} +  O\qty(\frac{d_1 d_2 \tau}{\sqrt{n} (S_1^{(1) 1/2} \wedge S_1^{(2) 1/2})} + \frac{\tau}{\sqrt{n}}) \Biggr\}\\
        &= (1 + \beta_{\CL,2}^{(1,2)}) \Biggl\{ \frac{p_{w,w'}^{(1,2)}}{p_w^{(1)} p_{w'}^{(2)}} - 1 + \beta_{\CL,1}^{(1,2)} + O\qty(\frac{d_1 d_2 \tau}{\sqrt{n} (S_1^{(1) 1/2} \wedge S_1^{(2) 1/2})} + \frac{\tau}{\sqrt{n}}) \Biggr\}
    \end{align*}
    \end{small}\noindent
    holds with high probability, where 
    $$
    \beta_{\CL,2}^{(1,2)} \triangleq \qty{\frac{\sum_{i \in [n]} T_i^{(1)} (T_i - 1)}{\sum_{i \in [n]} T_i (T_i - 1)}}^{-1} \qty{\frac{\sum_{i \in [n]} T_i^{(2)} (T_i - 1)}{\sum_{i \in [n]} T_i (T_i - 1)}}^{-1} \frac{\sum_{i \in [n]} T_i^{(1)} T_i^{(2)}}{\sum_{i \in [n]} T_i (T_i - 1)} - 1.
    $$
    The fact that $\beta_{\CL,2}^{(1,2)} \lesssim 1$ from Lemma~\ref{lem: bias ap} gives
    \begin{align}
        \PMIbbhat_\CL(w,w') = \frac{p_{w,w'}^{(1,2)}}{p_w^{(1)} p_{w'}^{(2)}} - 1 + \beta_{\CL,1}^{(1,2)} + \beta_{\CL,2}^{(1,2)} \qty(\frac{p_{w,w'}^{(1,2)}}{p_w^{(1)} p_{w'}^{(2)}} - 1 + \beta_{\CL,1}^{(1,2)}) + O\qty(\frac{d_1 d_2 \tau}{\sqrt{n} (S_1^{(1) 1/2} \wedge S_1^{(2) 1/2})} + \frac{\tau}{\sqrt{n}}).\label{eq: PMI CL 12}
    \end{align}
    From Lemma~\ref{lem: PMI decomposition in max norm},
    \begin{align*}
        \max_{w\in\calW^{(M)}, w'\in\calW^{(M')}} \abs{\PMIbb(w,w') - \qty(\frac{p_{w,w'}^{(M,M')}}{p_w^{(M)} p_{w'}^{(M')}} - 1)} \lesssim \frac{p^2 \tau}{(d_M \wedge d_{M'})^2}.
    \end{align*}
    Therefore, \eqref{eq: PMI CL 11}, \eqref{eq: PMI CL 22} and \eqref{eq: PMI CL 12} again with $\beta_{\CL,2}^{(1,1)} \vee \beta_{\CL,2}^{(1,2)} \vee \beta_{\CL,2}^{(2,2)} \lesssim 1$ yield
    \begin{align*}
        &\max_{w,w' \in \calW^{(1)}} \abs{\PMIbbhat_\CL(w,w') - \qty{\PMIbb(w,w') + \beta_{\CL,1}^{(1,1)} + \beta_{\CL,2}^{(1,1)} \qty(\PMIbb(w,w') + \beta_{\CL,1}^{(1,1)})}}\\
        &\quad\lesssim \frac{d_1^2 \tau}{\sqrt{n} S_1^{(1) 1/2}} + \frac{\tau}{\sqrt{n}} + \frac{p^2 \tau}{d_1^2},\\
        &\max_{w,w' \in \calW^{(2)}} \abs{\PMIbbhat_\CL(w,w') - \qty{\PMIbb(w,w') + \beta_{\CL,1}^{(2,2)} + \beta_{\CL,2}^{(2,2)} \qty(\PMIbb(w,w') + \beta_{\CL,1}^{(2,2)})}}\\
        &\quad\lesssim \frac{d_2^2 \tau}{\sqrt{n} S_1^{(2) 1/2}} + \frac{\tau}{\sqrt{n}} + \frac{p^2 \tau}{d_2^2},\\
        &\max_{w \in \calW^{(1)}, w' \in \calW^{(2)}} \abs{\PMIbbhat_\CL(w,w') - \qty{\PMIbb(w,w') + \beta_{\CL,1}^{(1,2)} + \beta_{\CL,2}^{(1,2)} \qty(\PMIbb(w,w') + \beta_{\CL,1}^{(1,2)})}}\\
        &\quad\lesssim \frac{d_1 d_2 \tau}{\sqrt{n} (S_1^{(1) 1/2} \wedge S_1^{(2) 1/2})} + \frac{\tau}{\sqrt{n}} + \frac{p^2 \tau}{(d_1 \wedge d_2)^2}
    \end{align*}
    with high probability. By definition of $\Bb_\CL$, this gives the first claim. The second claim follows from $\|\bA\|_\F \leq \sqrt{d} \|\bA\|_{\max}$ for any $\bA \in \R^{d \times d}$.
\end{proof}

\begin{thm}[Restatement of Theorem~\ref{thm: CL and PCA convergence}: CL Part]\label{thm: CL convergence ap}
    Suppose that Assumptions~\ref{asm: signal covariance}-\ref{asm: T moments} hold.
    Let $\Vbhat_\CL$ be the minimizer of the loss \ref{loss: CL linear}.
    If $\|\Bb_\CL\| \ll 1/p$, then,
    \begin{align*}
        &\norm{\sin\Theta\qty(\Pb_p(\Vbhat_\CL^\top), \Pb_p\qty(\frac{1}{p} \Vb^\star \Vb^{\star \top} + \bSigma + \Bb_\CL))}_\F\\
        &\quad\lesssim \frac{p (d_1 \vee d_2) d_1^2 \tau}{\sqrt{n} S_1^{(1) 1/2}} + \frac{p (d_1 \vee d_2) d_2^2 \tau}{\sqrt{n} S_1^{(2) 1/2}} + \frac{p (d_1 \vee d_2) d_1 d_2 \tau}{\sqrt{n} (S_1^{(1) 1/2} \wedge S_1^{(2) 1/2})} + \frac{p \tau (d_1 \vee d_2)}{\sqrt{n}} + \frac{p^3 (d_1 \vee d_2) \tau}{(d_1 \wedge d_2)^2}
    \end{align*}
    holds with probability $1 - \exp(-\Omega(\tau^2))$.
\end{thm}
Note that although $(1/p) \Vb^\star \Vb^{\star \top} + \bSigma + \Bb_\CL$ is not necessarily a positive definite matrix, its rank-$p$ approximation is positive definite under the assumed conditions.

\begin{proof}[Proof of Theorem~\ref{thm: CL convergence ap}]
    From Proposition~\ref{prop: PCA CL ap}, we can rewrite the loss $\mL_\CL(\Vb)$ as
    \begin{align*}
        \mL_\CL(\Vb) &= \frac{\lambda}{2} \Big\| \Vb \Vb^\top - \frac{1}{\lambda} \PMIbbhat_\CL\Big\|_\F^2 + (\textnormal{constant}).
    \end{align*}
    Denote $s_j^{\text{E}}(A)$ by the $j$-th largest eigenvalue of a symmetric matrix $A$.
    Note that $\PMIbbhat_\CL$ is not necessarily positive definite, but a symmetric matrix. To prove that the minimizer of $\mL_\CL$ satisfies $\Vbhat_\CL \Vbhat_\CL^\top = (1/\lambda) \SVD_p(\PMIbbhat_\CL)$, we prove $s_p^{\text{E}}(\PMIbbhat_\CL) \geq |s_j^{\text{E}}(\PMIbbhat_\CL)|$ for all $j \geq p+1$.

    Define $\Sb \triangleq (1/p) \Vb^\star \Vb^{\star \top} + \bSigma + \Bb_\CL$.
    We first prove $s_p^{\text{E}}(\Sb) \geq |s_j^{\text{E}}(\Sb)|$ for all $j \geq p+1$.
    Since $(1/p) \Vb^\star \Vb^{\star \top}$ is a positive definite matrix, Weyl's inequality gives
    \begin{align*}
        s_p^{\text{E}}(\Sb) &\geq s_p^{\text{E}}\qty(\frac{1}{p} \Vb^\star \Vb^{\star \top} + \bSigma) - \|\Bb_\CL\| \geq s_p^{\text{E}}(\bSigma) - \|\Bb_\CL\|.
    \end{align*}
    From Assumptions~\ref{asm: signal covariance} and \ref{asm: noise covariance}, we obtain $s_p^{\text{E}}(\Sb) \gtrsim (1/p) > 0$. By a similar argument, we have
    \begin{align*}
        s_{p+1}^{\text{E}}(\Sb) &\leq s_{p+1}^{\text{E}}(\bSigma) + \frac{1}{p} \|\Vb^\star \Vb^{\star \top}\| + \|\Bb_\CL\|,\\
        s_d^{\text{E}}(\Sb) &\geq s_d^{\text{E}}\qty(\frac{1}{p} \Vb^\star \Vb^{\star \top} + \bSigma) - \|\Bb_\CL\| \geq - \|\Bb_\CL\|.
    \end{align*}
    From these inequalities, we have
    \begin{align*}
        s_p^{\text{E}}(\Sb) - |s_j^{\text{E}}(\Sb)| &\geq \qty(s_p^{\text{E}}(\bSigma) - s_{p+1}^{\text{E}}(\bSigma) - \frac{1}{p} \|\Vb^\star \Vb^{\star \top}\| - 2\|\Bb_\CL\|) \wedge (s_p^{\text{E}}(\Sb) - \|\Bb_\CL\|)\\
        &\geq \qty(s_p(\bSigma) - s_{p+1}(\bSigma) - \frac{1}{p} \|\Vb^\star \Vb^{\star \top}\| - 2\|\Bb_\CL\|) \wedge \qty(s_p(\bSigma) - 2 \|\Bb_\CL\|)\\
        &= \qty(\frac{s_p(\bSigma) - s_{p+1}(\bSigma)}{s_p(\bSigma)} s_p(\bSigma) - \frac{1}{p} \|\Vb^\star \Vb^{\star \top}\| - 2\|\Bb_\CL\|) \wedge \qty(s_p(\bSigma) - 2 \|\Bb_\CL\|)
    \end{align*}
    for $j \geq p+1$. Again from Assumptions~\ref{asm: signal covariance}, \ref{asm: noise covariance}, \ref{asm: noise covariance eigengap} and $\|\Bb_\CL\| \ll 1/p$, we obtain $s_p^{\text{E}}(\Sb) - |s_j^{\text{E}}(\Sb)| \gtrsim 1/p$ for $j \geq p+1$.
    
    Next, Theorems~\ref{thm: PMI decomposition ap} and \ref{thm: PMI estimation cl ap} imply that
    \begin{align}
        &\norm{\PMIbbhat_\CL - \Sb}_\F\\
        &\quad\lesssim \frac{(d_1 \vee d_2) d_1^2 \tau}{\sqrt{n} S_1^{(1) 1/2}} + \frac{(d_1 \vee d_2) d_2^2 \tau}{\sqrt{n} S_1^{(2) 1/2}} + \frac{(d_1 \vee d_2) d_1 d_2 \tau}{\sqrt{n} (S_1^{(1) 1/2} \wedge S_1^{(2) 1/2})} + \frac{\tau (d_1 \vee d_2)}{\sqrt{n}} + \frac{p^2 (d_1 \vee d_2) \tau}{(d_1 \wedge d_2)^2}\label{eq: PMI CL and S}
    \end{align}
    holds with high probability, and thus $\|\PMIbbhat_\CL - \Sb\| \ll 1/p$ holds by Assumption~\ref{asm: regime}. For $j \geq p+1$, Weyl's inequality yields
    \begin{align}
        s_p^{\text{E}}(\PMIbbhat_\CL) - |s_j^{\text{E}}(\PMIbbhat_\CL)| \geq s_p^{\text{E}}(\Sb) - |s_j^{\text{E}}(\Sb)| - 2\|\PMIbbhat_\CL - \Sb\| \gtrsim 1/p\label{eq: eigengap CL}
    \end{align}
    with high probability. Therefore, $\mL_\CL(\Vb)$ has the global minimizer $\Vbhat_\CL$ satisfying
    \begin{align}
        \Vbhat_\CL \Vbhat_\CL^\top = \frac{1}{\lambda} \SVD_p(\PMIbbhat_\CL)\label{eq: hat VV}
    \end{align}
    with high probability.

    To prove the claim, we apply Theorem 2 from \cite{yu2015useful} using \eqref{eq: eigengap CL} to obtain
    \begin{align*}
        \|\sin\Theta(\Pb_p(\Vbhat_\CL^\top), \Pb_p(\Sb))\|_\F &\lesssim \frac{1}{1/p} \norm{\PMIbbhat_\CL - \Sb}_\F
    \end{align*}
    with high probability. The conclusion follows from \eqref{eq: PMI CL and S}.
\end{proof}

\subsection{Concate and Diagonal-deletion PCA on Joint PMI Matrix}\label{sec: PCA ap}

Here we restate part of Theorem~\ref{thm: CL and PCA convergence}. We also provide the results for diagonal-deletion PCA \citep{florescu2016spectral,zhang2022heteroskedastic} applied to the joint PMI matrix, which applies PCA to the joint PMI matrix, with diagonal entries set to $0$. Diagonal-deletion PCA is shown to be effective at removing coordinate-wise uncorrelated noise \citep{zhang2022heteroskedastic}.

\begin{thm}[Restatement of Theorem~\ref{thm: CL and PCA convergence}: Concate Part]\label{thm: PCA DDPCA convergence ap}
    Suppose that Assumptions~\ref{asm: signal covariance}-\ref{asm: T moments} hold.
    Let $\Ubhat_\PCA = \Pb_p(\PMIbbhat)$ and $\Ubhat_\DDPCA = \Pb_p(\Delta(\PMIbbhat))$.
    If $\|\Bb_\PCA\| \ll 1/p$, then,
    \begin{align*}
        &\norm{\sin\Theta\qty(\Ubhat_\PCA, \Pb_p\qty(\frac{1}{p} \Vb^\star \Vb^{\star \top} + \bSigma + \Bb_\PCA))}_\F\\
        &\quad\lesssim \frac{p (d_1 \vee d_2) d_1^2 \tau}{\sqrt{n} S_1^{(1) 1/2}} + \frac{p (d_1 \vee d_2) d_2^2 \tau}{\sqrt{n} S_1^{(2) 1/2}} + \frac{p (d_1 \vee d_2) d_1 d_2 \tau}{\sqrt{n} (S_1^{(1) 1/2} \wedge S_1^{(2) 1/2})} + \frac{p (d_1 \vee d_2) \tau}{\sqrt{n}} + \frac{p^3 (d_1 \vee d_2) \tau}{(d_1 \wedge d_2)^2}
    \end{align*}
    holds with probability $1 - \exp(-\Omega(\tau^2))$.
    Furthermore, if $(d_1 \vee d_2)^{1/2} \|D(\Bb_\PCA)\| \ll 1/p$, then,
    \begin{align*}
        &\norm{\sin\Theta\qty(\Ubhat_\DDPCA, \Pb_p\qty(\frac{1}{p} \Vb^\star \Vb^{\star \top} + \bSigma + \Bb_\PCA))}_\F\\
        &\quad\lesssim \frac{p (d_1 \vee d_2) d_1^2 \tau}{\sqrt{n} S_1^{(1) 1/2}} + \frac{p (d_1 \vee d_2) d_2^2 \tau}{\sqrt{n} S_1^{(2) 1/2}} + \frac{p (d_1 \vee d_2) d_1 d_2 \tau}{\sqrt{n} (S_1^{(1) 1/2} \wedge S_1^{(2) 1/2})} + \frac{p (d_1 \vee d_2) \tau}{\sqrt{n}}\\
        &\quad\quad+ \frac{p^3 (d_1 \vee d_2) \tau}{(d_1 \wedge d_2)^2} + \frac{p^2 (d_1 \vee d_2)^{1/2}}{d_1 \wedge d_2} + p (d_1 \vee d_2)^{1/2} \|D(\Bb_\PCA)\|
    \end{align*}
    holds with probability $1 - \exp(-\Omega(\tau^2))$.
\end{thm}

\begin{proof}[Proof of Theorem~\ref{thm: PCA DDPCA convergence ap}]
    Define $\Sb \triangleq (1/p) \Vb^\star \Vb^{\star \top} + \bSigma + \Bb_\PCA$.
    From Theorems~\ref{thm: PMI decomposition ap} and \ref{thm: PMI estimation ap}, combined with $\|\Delta(\bA)\|_\F \leq \|\bA\|_\F$, we have
    \begin{align}
        &\norm{\PMIbbhat - \Sb}_\F \vee \norm{\Delta(\PMIbbhat) - \Delta(\Sb)}_\F\nonumber\\
        &\quad\lesssim \frac{(d_1 \vee d_2) d_1^2 \tau}{\sqrt{n} S_1^{(1) 1/2}} + \frac{(d_1 \vee d_2) d_2^2 \tau}{\sqrt{n} S_1^{(2) 1/2}} + \frac{(d_1 \vee d_2) d_1 d_2 \tau}{\sqrt{n} (S_1^{(1) 1/2} \wedge S_1^{(2) 1/2})} + \frac{(d_1 \vee d_2) \tau}{\sqrt{n}} + \frac{p^2 (d_1 \vee d_2) \tau}{(d_1 \wedge d_2)^2}\label{eq: PMI hat and S tilde}
    \end{align}
    with high probability.
    We first derive the bound for $\Ubhat_\PCA$.
    By a similar argument as in the proof of Theorem~\ref{thm: CL convergence ap}, and $\|\Bb_\PCA\| \ll 1/p$ by assumption,
    we have $s_p(\Sb) - s_{p+1}(\Sb) \gtrsim 1/p$.
    Using Theorem 2 from \cite{yu2015useful}, we obtain
    \begin{align*}
        \|\sin\Theta(\Ubhat_\PCA, \Pb_p(\Sb))\|_\F &\lesssim \frac{1}{p^{-1}} \norm{\PMIbbhat - \Sb}_\F.
    \end{align*}
    with high probability. The first claim follows from \eqref{eq: PMI hat and S tilde}.
    
    Next we derive the bound for $\Ubhat_\DDPCA$. Note that
    \begin{align*}
        \|\Delta(\Sb) - \Sb\|_\F &= \|D(\Sb)\|_\F \leq \sqrt{d_1 + d_2} \|D(\Sb)\|\\
        &\leq \sqrt{d_1 + d_2} \qty(\frac{1}{p} \norm{\Vb^\star}_{2,\infty}^2 + \|\diag(\bSigma)\|_{\max} + \|D(\Bb_\PCA)\|)\\
        &\lesssim \frac{p (d_1 \vee d_2)^{1/2}}{d_1 \wedge d_2} + (d_1 \vee d_2)^{1/2} \|D(\Bb_\PCA)\|,
    \end{align*}
    where the last inequality follows from Assumptions~\ref{asm: signal covariance} and \ref{asm: noise covariance}.
    Thus
    \begin{align*}
        \norm{\Delta(\PMIbbhat) - \Sb}_\F &\leq \norm{\Delta(\PMIbbhat) - \Delta(\Sb)}_\F + \norm{\Delta(\Sb) - \Sb}_\F\\
        &\lesssim \norm{\Delta(\PMIbbhat) - \Delta(\Sb)}_\F + \frac{p (d_1 \vee d_2)^{1/2}}{d_1 \wedge d_2} + (d_1 \vee d_2)^{1/2} \|D(\Bb_\PCA)\|
    \end{align*}
    holds with high probability. A similar argument as above combined with the additional assumption $(d_1 \vee d_2)^{1/2} \|D(\Bb_\PCA)\| \ll 1/p$ concludes the proof.
\end{proof}

\subsection{Superiority of CLAIME over CL, Concate and DDPCA}\label{sec: superiority}

In this section, we show the inability of simple concatenation, Contrastive Learning and Diagonal-deletion PCA to learn the representations.
\begin{thm}[Restatement of Theorem~\ref{thm: PCA DDPCA MMCL}]\label{thm: PCA DDPCA MMCL ap}
    Suppose that Assumptions~\ref{asm: signal covariance}-\ref{asm: noise covariance incoherence 1} hold.
    If $\|\Bb_\CL\| \vee \|\Bb_\PCA\| \ll 1/p$, then,
    \begin{align*}
        \|\sin\Theta(\Ubhat_\PCA, \Ub^\star)\|_\F \wedge \|\sin\Theta(\Ubhat_\CL, \Ub^\star)\|_\F &\gtrsim \sqrt{p}
     \end{align*}
    holds with probability $1 - \exp(-\Omega(\tau^2))$.
    Furthermore, if $\|\Bb_\CL\| \vee (d_1 \vee d_2)^{1/2} \|\Bb_\PCA\| \ll 1/p$, then,
    \begin{align*}
        \|\sin\Theta(\Ubhat_\DDPCA, \Ub^\star)\|_\F &\gtrsim \sqrt{p}
    \end{align*}
    holds with probability $1 - \exp(-\Omega(\tau^2))$.
\end{thm}

\begin{proof}[Proof of Theorem~\ref{thm: PCA DDPCA MMCL ap}]
    Recall that $\Ub^\star = (1/\sqrt{2})[\Ub_1^{\star \top}; \Ub_2^{\star \top}]$.
    We first bound $\|\sin\Theta(\Ubhat_\PCA, \Ub^\star)\|_\F$ from below.
    Note that $\|\sin\Theta(\Ub^\star, \Pb_p(\bSigma))\|_\F^2 = p - \|\Ub^{\star \top} \Pb_p(\bSigma)\|_\F^2$.
    Since the top-$p$ column space of $\bSigma$ is contained in a subspace spanned by column vectors of $[\Pb_p(\bSigma_1)^\top; \ZERO]^\top$ and $[\ZERO; \Pb_p(\bSigma_2)^\top]^\top$, we can write $\Pb_p(\bSigma)$ as 
    \begin{align*}
        \Pb_p(\bSigma) = \begin{pmatrix}
            \Pb_p(\bSigma_1)\\
            \ZERO
        \end{pmatrix} \Hb_1 + 
        \begin{pmatrix}
            \ZERO\\
            \Pb_p(\bSigma_2)
        \end{pmatrix} \Hb_2,
    \end{align*}
    where $\Hb_1, \Hb_2 \in \R^{p \times p}$ satisfy $\|\Hb_1\| \vee \|\Hb_2\| \leq 1$. Thus,
    \begin{align*}
        \|\Ub^{\star \top} \Pb_p(\bSigma)\|_\F^2 &= \|\Ub_1^{\star \top} \Pb_p(\bSigma_1)\|_\F^2 + \|\Ub_2^{\star \top} \Pb_p(\bSigma_2)\|_\F^2\\
        &= (p - \|\sin\Theta(\Ub^\star_1, \Pb_p(\bSigma_1))\|_\F^2) + (p - \|\sin\Theta(\Ub^\star_2, \Pb_p(\bSigma_2))\|_\F^2)\\
        &\leq \frac{2p}{3},
    \end{align*}
    where the last inequality follows from Assumption~\ref{asm: noise covariance incoherence 1}.
    This implies $\|\sin\Theta(\Ub^\star, \Pb_p(\bSigma))\|_\F^2 \geq (1/3) p$.
    From Theorem~\ref{thm: PCA DDPCA convergence ap} and Assumption~\ref{asm: regime}, we obtain
    \begin{align*}
        \|\sin\Theta(\Ubhat_\PCA, \Pb_p(\Sb))\|_\F \ll 1,
    \end{align*}
    where $\Sb \triangleq (1/p) \Vb^\star \Vb^{\star \top} + \bSigma + \Bb_\PCA$.
    Theorem 2 from \cite{yu2015useful} yields
    \begin{small}
    \begin{align*}
        \|\sin\Theta(\Pb_p(\Sb), \Pb_p(\bSigma))\|_\F \lesssim \frac{\sqrt{p}}{s_p(\bSigma) - s_{p+1}(\bSigma)} \norm{\Sb - \bSigma} = \frac{\sqrt{p}}{s_p(\bSigma) - s_{p+1}(\bSigma)} \norm{\frac{1}{p} \Vb^\star\Vb^{\star \top} + \Bb_\PCA} \ll \sqrt{p},
    \end{align*}
    \end{small}\noindent
    where we used Theorem~\ref{thm: PMI decomposition ap}, Assumptions~\ref{asm: noise covariance}, \ref{asm: noise covariance eigengap} and $\|\Bb_\PCA\| \ll 1/p$.
    By the triangle inequality,
    \begin{small}
    \begin{align*}
        \|\sin\Theta(\Ubhat_\PCA, \Ub^\star)\|_\F
        &\geq \|\sin\Theta(\Pb_p(\bSigma), \Ub^\star)\|_\F - \|\sin\Theta(\Ubhat_\PCA, \Pb_p(\Sb))\|_\F - \|\sin\Theta(\Pb_p(\Sb), \Pb_p(\bSigma))\|_\F\\
         &\geq \sqrt{\frac{p}{3}} - c - c' \sqrt{p} \gtrsim \sqrt{p},
    \end{align*}
    \end{small}\noindent
    where $c, c' > 0$ are sufficiently small constants. 
    The bound for $\Ubhat_\DDPCA$ follows similarly. Using Theorem~\ref{thm: CL convergence ap}, we can also obtain the bound for $\Ubhat_\CL$.
\end{proof}
\subsection{Bias from CL and Concate}\label{sec: bias}

In this subsection, we bound the bias terms $\Bb_\PCA$ and $\Bb_\CL$ introduced in Sections \ref{sec: PMI estimation ap} and \ref{sec: CL ap}.
\begin{lem}[Restatement of Lemma~\ref{lem: bias}]\label{lem: bias ap}
    Suppose that Assumptions~\ref{asm: signal covariance}, \ref{asm: regime} \ref{asm: noise covariance}, and \ref{asm: T moments} hold. Then,
    \begin{align*}
        \beta_\PCA^{(1,1)} \vee \beta_\PCA^{(2,2)} \vee \beta_\PCA^{(1,2)} &\lesssim 1,\\
        \beta_{\CL,1}^{(1,1)} \vee \beta_{\CL,1}^{(2,2)} \vee \beta_{\CL,1}^{(1,2)} &\lesssim 1,\\
        \beta_{\CL,2}^{(1,1)} \vee \beta_{\CL,2}^{(2,2)} \vee \beta_{\CL,2}^{(1,2)} &\lesssim 1.
    \end{align*}
    Moreover, if $T_i^{(1)} = T_i^{(2)}$, then,
    \begin{align*}
        \|\Bb_\PCA\| \vee \|\Bb_\CL\| \lesssim (d_1 \vee d_2) \frac{S_1^{(1)}}{S_2^{(1) 2}}.
    \end{align*}
    Furthermore, if $T_i^{(1)} = T_i^{(2)} \equiv T$ for all $i$, then
    \begin{align*}
        \|\Bb_\PCA\| \vee \|\Bb_\CL\| \lesssim \frac{d_1 \vee d_2}{T}.
    \end{align*}
\end{lem}

\begin{proof}
    For the term $\beta_\PCA^{(M,M)}$, note that under Assumption~\ref{asm: T moments},
    \begin{align}
        \frac{ \sum_{i=1}^n T_i (T_i-1) \sum_{i=1}^n T_i^{(M)} (T_i^{(M)}-1)  }{ \big ( \sum_{i=1}^n T_i^{(M)} (T_i -1) \big)^2 } &\lesssim \frac{(S_2^{(1) 2} + S_2^{(2) 2} + 2 S_2^{(1,2) 2}) S_2^{(M) 2}}{(S_2^{(M) 2} + S_2^{(1,2) 2})^2}\nonumber\\
        &\lesssim \frac{(S_2^{(1) 2} + S_2^{(2) 2}) S_2^{(M) 2}}{(S_2^{(M) 2} + S_1^{(1)} S_1^{(2)})^2}\nonumber\\
        &\lesssim \frac{(S_2^{(1) 2} + S_2^{(2) 2}) S_2^{(M) 2}}{S_2^{(M) 4} + S_2^{(1) 2} S_2^{(2) 2}} = 1,\label{eq: beta PCA 2 MM}
    \end{align}
    where we used $S_2^{(1,2) 2} \gtrsim S_1^{(1)} S_1^{(2)}$ in the second inequality, and $S_1^{(M')} \gtrsim S_2^{(M')}$ for $M' \in \{1,2\}$ in the last inequality. This yields $\beta_\PCA^{(M,M)} \lesssim 1$.
    For the term $\beta_\PCA^{(1,2)}$, a similar argument gives
    \begin{align}
        \frac{ \sum_{i=1}^n T_i (T_i-1) \sum_{i=1}^n T_i^{(1)} T_i^{(2)} }{ \qty( \sum_{i=1}^n T_i^{(1)} (T_i -1) ) \qty( \sum_{i=1}^n T_i^{(2)} (T_i -1) ) } &\lesssim \frac{(S_2^{(1) 2} + 2 S_2^{(1,2) 2} + S_2^{(2) 2}) S_2^{(1,2) 2}}{(S_2^{(1) 2} + S_2^{(1,2) 2}) (S_2^{(2) 2} + S_2^{(1,2) 2})}\nonumber\\
        &\lesssim \qty(\frac{1}{S_2^{(1) 2} + S_2^{(1,2) 2}} + \frac{1}{S_2^{(2) 2} + S_2^{(1,2) 2}}) S_2^{(1,2) 2}\nonumber\\
        &\lesssim 1.\label{eq: beta PCA 2 12}
    \end{align}
    Thus $\beta_\PCA^{(1,2)} \lesssim 1$.
    For the term $\beta_{\CL,1}^{(M,M)}$, a similar argument gives
    \begin{align*}
        \frac{\sum_{i \in [n]} T_i (T_i - 1)}{\sum_{i \in [n]} T_i^{(M)} (T_i^{(M)} - 1)} \frac{\sum_{i,j: i \neq j} T_i^{(M)} T_j^{(M)}}{\sum_{i,j: i \neq j} T_i T_j} &\lesssim \frac{(S_2^{(1) 2} + 2 S_2^{(1,2) 2} + S_2^{(2) 2}) S_1^{(M) 2}}{S_2^{(M) 2} (S_1^{(1) 2} + 2 S_1^{(1)} S_1^{(2)} + S_1^{(2) 2})} \lesssim 1,
    \end{align*}
    and hence $\beta_{\CL,1}^{(M,M)} \lesssim 1$.
    For the term $\beta_{\CL,1}^{(1,2)}$, a similar argument gives
    \begin{align*}
        \frac{\sum_{i \in [n]} T_i (T_i - 1)}{\sum_{i \in [n]} T_i^{(1)} T_i^{(2)}} \frac{\sum_{i,j: i \neq j} T_i^{(1)} T_j^{(2)}}{\sum_{i,j: i \neq j} T_i T_j} &\lesssim \frac{(S_2^{(1) 2} + 2 S_2^{(1,2) 2} + S_2^{(2) 2}) S_1^{(1)} S_1^{(2)}}{S_2^{(1,2) 2} (S_1^{(1) 2} + 2 S_1^{(1)} S_1^{(2)} + S_1^{(2) 2})}\\
        &\lesssim \frac{(S_1^{(1) 2} + S_1^{(2) 2}) S_1^{(1)} S_1^{(2)}}{S_2^{(1,2) 2} (S_1^{(1) 2} + S_1^{(2) 2})}\\
        &\lesssim 1,
    \end{align*}
    where we used $S_2^{(M')} \lesssim S_1^{(M')}$ by assumption and Cauchy-Schwarz inequality in the second inequality, and $S_2^{(1,2) 2} \gtrsim S_1^{(1)} S_1^{(2)}$ in the last inequality. This yields $\beta_{\CL,1}^{(1,2)} \lesssim 1$.
    For the term $\beta_{\CL,2}^{(M,M)}$, observe that by \eqref{eq: beta PCA 2 MM},
    \begin{align*}
        \qty(\frac{\sum_{i \in [n]} T_i^{(M)} (T_i - 1)}{\sum_{i \in [n]} T_i (T_i - 1)})^{-2} \frac{\sum_{i \in [n]} T_i^{(M)} (T_i^{(M)} - 1)}{\sum_{i \in [n]} T_i (T_i - 1)} &= \frac{ \sum_{i=1}^n T_i (T_i-1) \sum_{i=1}^n T_i^{(M)} (T_i^{(M)}-1)  }{ \big ( \sum_{i=1}^n T_i^{(M)} (T_i -1) \big)^2 } \lesssim 1.
    \end{align*}
    Hence $\beta_{\CL,2}^{(M,M)} \lesssim 1$. Similarly, \eqref{eq: beta PCA 2 12} gives $\beta_{\CL,2}^{(1,2)} \lesssim 1$
    
    If $T_i^{(1)} = T_i^{(2)}$, we have 
    \begin{align*}
        \beta_\PCA^{(M,M)} &= \log \frac{ \sum_{i=1}^n T_i (T_i-1) \sum_{i=1}^n T_i^{(M)} (T_i^{(M)}-1)  }{ \big ( \sum_{i=1}^n T_i^{(M)} (T_i -1) \big)^2 } \\
        &= \log \frac{ \sum_{i=1}^n 2T_i^{(M)} (2T_i^{(M)}-1) \sum_{i=1}^n T_i^{(M)} (T_i^{(M)}-1)  }{ \big ( \sum_{i=1}^n T_i^{(M)} (2 T_i^{(M)} - 1) \big)^2 } \\
        &= \log \frac{ 2 (2 S_2^{(M) 2} - S_1^{(M)}) (S_2^{(M) 2} - S_1^{(M)}) }{ (2 S_2^{(M) 2} - S_1^{(M)})^2 }\\
        &= \log(1 + \frac{S_1^{(M)}}{2 S_2^{(M) 2} - S_1^{(M)}})\\
        &= \log(1 + \frac{S_1^{(1)}}{2 S_2^{(1) 2} - S_1^{(1)}}).
    \end{align*} 
    Using Assumption~\ref{asm: T moments} and $\log (1 + x) \leq x$ for $x \geq 0$, we have $\log(1 + S_1^{(1)}/(2 S_2^{(1) 2} - S_1^{(1)})) = O(S_1^{(1)} / S_2^{(1) 2})$. 
    This yields $\|\Bb_\PCA^{(M,M)}\| \lesssim d_M S_1^{(1)}/S_2^{(1) 2}$, where we used $\|\1_{d_M} \1_{d_M}^\top\| = d_M$.
    Similarly, we obtain $\|\Bb_\PCA^{(1,2)}\| \lesssim \sqrt{d_1 d_2} S_1^{(1)}/S_2^{(1) 2}$.
    Hence $\|\Bb_\PCA\| \leq \|\Bb_\PCA^{(1,1)}\| + 2\|\Bb_\PCA^{(1,2)}\| + \|\Bb_\PCA^{(2,2)}\| \lesssim (d_1 \vee d_2) S_1^{(1)} / S_2^{(1) 2}$.
    Furthermore, if $T_i^{(1)} = T_i^{(2)} \equiv T$, then $\|\Bb_\PCA\| \lesssim (d_1 \vee d_2)/T$ since $S_2^{(1)} = S_1^{(1)} \equiv T$.

    A similar argument gives that if $T_i^{(1)} = T_i^{(2)}$, $\beta_{\CL,2}^{(M,M)} \lesssim S_1^{(1)} / S_2^{(1) 2}$ and $\beta_{\CL,2}^{(1,2)} \lesssim S_1^{(1)} / S_2^{(1) 2}$ follow. Also,
    \begin{align*}
        \beta_{\CL,1}^{(M,M)} &= 1 - \frac{\sum_{i \in [n]} T_i (T_i - 1)}{\sum_{i \in [n]} T_i^{(M)} (T_i^{(M)} - 1)} \frac{\sum_{i,j: i \neq j} T_i^{(M)} T_j^{(M)}}{\sum_{i,j: i \neq j} T_i T_j}\\
        &= 1 - \frac{1}{2} \frac{\sum_{i \in [n]} T_i^{(1)} (2T_i^{(1)} - 1)}{\sum_{i \in [n]} T_i^{(1)} (T_i^{(1)} - 1)}\\
        &\lesssim \frac{S_1^{(1)}}{S_2^{(1) 2}}.
    \end{align*}
    By a similar argument, we obtain $\beta_{\CL,1}^{(1,2)} \lesssim S_1^{(1)} / S_2^{(1) 2}$.
    From Theorem~\ref{thm: PMI decomposition ap} and Assumptions~\ref{asm: signal covariance} and \ref{asm: noise covariance}, $\|\PMIbb\| \lesssim 1$. This yields
    \begin{align*}
        \|\Bb_\CL^{(M,M)}\| \leq \beta_{\CL,1}^{(M,M)} \|\1_{d_M} \1_{d_M}^\top\| + \beta_{\CL,2}^{(M,M)} (\|\PMIbb^{(M,M)}\| + \beta_{\CL,1}^{(M,M)} \|\1_{d_M} \1_{d_M}^\top\|) \lesssim d_M \frac{S_1^{(1)}}{S_2^{(1) 2}}.
    \end{align*}
    A similar argument gives $\|\Bb_\CL^{(1,2)}\| \lesssim \sqrt{d_1 d_2} S_1^{(1)} / S_2^{(1) 2}$. Hence $\|\Bb_\CL\| \lesssim (d_1 \vee d_2) S_1^{(1)} / S_2^{(1) 2}$.
    Furthermore, if $T_i^{(1)} = T_i^{(2)} \equiv T$, a similar argument as above gives $\|\Bb_\CL\| \lesssim (d_1 \vee d_2) / T$. This concludes the proof.
\end{proof}

\section{Auxiliary Results}

In this section, we provide lemmas used to prove our main results. We first extend Lemma A.1 from \cite{xu2023inference} to incorporate a noisy data-generating process.
\begin{lem}[Modification to Lemma A.1 from \cite{xu2023inference}]\label{lem: A1}
    Suppose that Assumptions~\ref{asm: signal covariance} and \ref{asm: noise covariance} hold. 
    For $M \in \{1, 2\}$, define $\bq^{(M)} = (q_w)_{w \in \calW^{(M)}}$, where
    \begin{align*}
        q_w \triangleq \frac{1}{Z^{(M)}} \exp(\frac{\|\v_w^\star\|^2}{2p} + \frac{\sigma_{w,w}}{2}).
    \end{align*}
    Then, we have
    \begin{align*}
        \norm{\bq^{(M)} - \frac{1}{d_M}\boldsymbol{1}_{d_{M}}} \lesssim \frac{p}{d_M^{3/2}}.
    \end{align*}
\end{lem}

\begin{proof}[Proof of Lemma~\ref{lem: A1}]
    Without loss of generality, we focus on $M=1$.
    We omit the superscript $(\cdot)^{(1)}$ from $\calW^{(1)}$ and $Z^{(1)}$.
    Let $\b_w \triangleq \bSigma_1^{1/2} \be_w$ for $w \in \calW$.
    From Assumptions~\ref{asm: signal covariance} and \ref{asm: noise covariance}, we have that for all $w\in \calW$, $\|\v_w^\star\|^2/(2p) + \|\b_w\|^2/2 = O(1)$, and thus
    \begin{align*}
        \exp(\|\v_w^\star\|^2/(2p) + \|\b_w\|^2/2) &= 1 + \frac{\|\v_w^\star\|^2}{2p} + \frac{\|\b_w\|^2}{2} + O\qty(\frac{\|\Vb_1^\star\|_{2,\infty}^4}{p^4} + \max_{w \in \calW} \|\b_w\|^4)\\
        &= 1 + \frac{\|\v_w^\star\|^2}{2p} + \frac{\|\b_w\|^2}{2} + O\qty(\frac{p^2}{d_1^2}),
    \end{align*}
    where we used $\|\b_w\|^2 = \sigma_{w,w}$.
    It follows that
    \begin{equation*}
        \begin{split}
            \max_{w \in \calW} \abs{q_w - \frac{1}{d_1}} &= \max_{w \in \calW} \frac{|d_1 \exp(\|\v_w^\star\|^2/(2p) + \|\b_w\|^2/2) - Z|}{d_1 Z}\\
            &\leq \max_{w \in \calW} \frac{|d_1 \|\v_w^\star\|^2/(2p) + d_1 \|\b_w\|^2/2 - \sum_{w'} (\|\v_{w'}^\star\|^2/(2p) + \|\b_w\|^2/2)|}{d_1  Z} + O\qty(\frac{p^2}{d_1^3})\\
            &\leq \frac{\|\Vb^\star\|_{2,\infty}^2}{p Z} + \frac{\max_{w \in \calW} \sigma_{w,w}}{Z} + O\qty(\frac{p^2}{d_1^3})\\
            &\lesssim \frac{1}{d_1^2} + \frac{p}{d_1^2} + \frac{p^2}{d_1^3}\\
            &\lesssim \frac{p}{d_1^2},
        \end{split}
    \end{equation*}
    where the third inequality follows from $Z\geq d_1$.
    This concludes the proof.
\end{proof}

Next we provide the following lemma to bound the normalizing constant appearing in model \ref{model}.
\begin{lem}[Modification to Lemma A.3 from \cite{xu2023inference}]\label{lem: A3}
    Suppose that Assumptions~\ref{asm: signal covariance}, \ref{asm: regime} and \ref{asm: noise covariance} hold.
    Fix $M \in \{1, 2\}$.
    Let $Z^{(M)}(\c, \beps) \triangleq \sum_{w \in \calW^{(M)}} \exp(\v_w^{\star \top} \c + \epsilon_w)$. Then,
    \begin{align*}
        \P((1 - \varepsilon_{d_M}) Z^{(M)} \leq Z^{(M)}(\c, \beps) \leq (1 + \varepsilon_{d_M}) Z^{(M)}) = 1 - \exp(-\Omega(\tau^2)),
    \end{align*}
    where $\varepsilon_{d_M} = 2p\tau/d_M$ and $Z^{(M)} \triangleq \E[Z^{(M)}(\c, \beps)]$.
\end{lem}

\begin{proof}[Proof of Lemma~\ref{lem: A3}]
    Without loss of generality, we focus on $M=1$.
    We omit the superscript $(\cdot)^{(1)}$ and subscript $(\cdot)_1$ from $\calW^{(1)}$, $Z^{(1)}$, and $\bSigma_1$.
    By assumption, $\beps^{(1)} = \bSigma^{1/2} \beps'$, where $\beps' \sim N(\zero, \Ib_{d_1})$ and $\1_{d_1}^\top \bSigma = \zero$. We write $\bc' = \sqrt{p}\bc$ and $\b_w \triangleq \bSigma^{1/2} \be_w$.
    Denote an event $\mathcal{F} = \{\|\bc'\| \leq (\sqrt{p}/2) \tau\} \cap \{\max_{w \in \calW} |\b_w^\top \beps'| \leq (1/2) \sqrt{\max_{w \in \calW} \sigma_{w,w}} \tau\}$, then $\P(\mathcal{F}) = 1 - \exp(-\Omega(\tau^2))$ by the union bound argument.
    Let 
    $$
        f(\bc', \beps') = \sum_{w\in \calW} \exp\left((1/\sqrt{p}) \v_w^{\star \top} \bc' + \b_w^\top \beps' \right)I\{\mathcal{F}\}.
    $$ 
    Note that $\|\b_w\|^2 = \sigma_{w,w}$. Then, we have
    \begin{equation}
        \begin{split}
            \|\nabla f\|^2 &\leq \frac{1}{p} \sum_{j=1}^p \qty(\sum_{w\in\calW} (\v_w^\star)_j \exp\left((1/\sqrt{p})\v_w^{\star \top} \bc' + \b_w^\top \beps'\right))^2\\
            &\quad+ \sum_{j=1}^{d_1} \qty(\sum_{w\in\calW} (\b_w)_j \exp\left((1/\sqrt{p})\v_w^{\star \top} \bc' + \b_w^\top\beps'\right))^2.
         \end{split}
    \end{equation}
    We first bound the second term on the right hand side. Observe that
    \begin{align}
        &\sum_{j=1}^{d_1} \qty(\sum_{w \in \calW} (\b_w)_j \exp((1/\sqrt{p})\v_w^{\star \top} \bc' + \b_w^\top\beps'))^2\nonumber\\
        &\quad= \sum_{j=1}^{d_1} \qty{\sum_{w \in \calW} (\b_w)_j \qty(1 + \frac{1}{\sqrt{p}} \v_w^{\star \top} \bc' + \b_w^\top\beps') + O\qty(\sum_{w \in \calW} (\b_w)_j \qty(\frac{1}{\sqrt{p}} \v_w^{\star \top} \bc' + \b_w^\top\beps')^2)}^2\nonumber\\
        &\quad= O\qty(\sum_{j=1}^{d_1} \qty{\sum_{w \in \calW} |(\b_w)_j| \abs{\frac{1}{\sqrt{p}} \v_w^{\star \top} \bc' + \b_w^\top\beps'}}^2)\nonumber
    \end{align}
    holds, where the first equality follows from $\sum_{w \in \calW} \b_w = \zero$ and Assumption~\ref{asm: signal covariance} and \ref{asm: noise covariance}. The second equality follows from the fact that the big-O notation in the first equality does not depend on $j$. Thus,
    \begin{align}
        &\sum_{j=1}^{d_1} \qty(\sum_{w \in \calW} (\b_w)_j \exp((1/\sqrt{p})\v_w^{\star \top} \bc' + \b_w^\top\beps'))^2\nonumber\\
        &\quad\lesssim \sum_{j=1}^{d_1} \qty(\sum_{w\in\calW} (\b_w)_j^2 ) \qty(\sum_{w\in\calW} \qty(\frac{1}{\sqrt{p}} \v_w^{\star \top} \bc' + \b_w^\top\beps')^2 )\nonumber\\
        &\quad\lesssim (d_1 \max_{w \in \calW} \|\b_w\|^2) \qty(d_1 \qty(\frac{1}{p} \max_{w \in \calW} \|\v_w^\star\|^2 \|\bc'\|^2 + \max_{w \in \calW} |\b_w^\top \beps'|^2))\nonumber\\
        &\quad\lesssim p^2 \tau^2\label{eq: f grad 1}
    \end{align}
    holds, where we used Assumptions~\ref{asm: signal covariance} and \ref{asm: noise covariance} again.
    Similarly,
    \begin{align}
        \frac{1}{p} \sum_{j=1}^{d_1} \qty(\sum_{w \in \calW} (\v_w^\star)_j \exp((1/\sqrt{p})\v_w^{\star \top} \bc' + \b_w^\top\beps'))^2 
         &\leq p \tau^2.\label{eq: f grad 2}
    \end{align}
    Combining \eqref{eq: f grad 1} and \eqref{eq: f grad 2}, we obtain
    \begin{align*}
        \|\nabla f\|^2 &\lesssim p^2 \tau^2.
    \end{align*}
    
    Note that
    \begin{equation*}
        \E[Z(\bc, \beps)^2] \leq d_1 \E\qty[\sum_{w\in\calW} \exp(2\v_w^{\star \top}\bc + 2\b_w^\top \beps')] = d_1 \sum_{w\in\calW} \exp(\frac{2\|\v_w^\star\|^2}{p} + 2 \|\b_w\|^2) \leq d_1^2\exp(C'\frac{p}{d_1}),
    \end{equation*}
    for some $C' > 0$.
    Using the Cauchy-Schwarz inequality and $d_1 \geq p$, we have
    $$
    0 \leq \E[Z(\bc, \beps)] - \E[Z(\bc, \beps)I\{\mathcal{F}\}] \leq \sqrt{\P(\mathcal{F}^c) d_1^2 \exp(C' \frac{p}{d_1})} = \exp (-\Omega(\tau^2)).
    $$
    By Jensen's inequality, we have $\E_{\bc, \beps}[\exp(\v_w^{\star \top}\bc + \b_w^\top \beps')]\geq 1$. 
    Note that $Z = \E_{\bc, \beps}[Z(\bc, \beps)] = \sum_{w\in\calW} \E[\exp(\v_w^{\star \top}\bc + \b_w^\top \beps')] \geq d_1$, and that $(\bc', \beps') \sim N(\zero,\Ib_{p+d_1})$. 
    Using Lemma F.1 from \cite{xxx2021}, we have
    \begin{small}
    \begin{equation}
        \begin{split}
            \P \Big(\Big|\frac{Z(\bc, \beps) - Z}{Z}\Big| \geq 2\varepsilon \Big)&= \P \Big(\Big|\frac{Z(\bc, \beps) - Z}{Z}\Big| \geq 2\varepsilon, \cF^c \Big) + \P \Big(\Big|\frac{Z(\bc, \beps) - Z}{Z}\Big| \geq 2\varepsilon, \cF \Big)\\
            &\leq  \P(\cF^c) + \P\qty(\left| \frac{Z(\bc, \beps)I\{\mathcal{F}\} - \E[Z(\bc, \beps)I\{\mathcal{F}\}]}{Z} \right| \geq 2\varepsilon - \left|\frac{Z - \E[Z(\bc, \beps)I\{\mathcal{F}\}]}{Z}\right|)\\
            &\leq \P(\mathcal{F}^c) + \P \left(\left| \frac{Z(\bc, \beps)I\{\mathcal{F}\} - \E[Z(\bc, \beps)I\{\mathcal{F}\}]}{Z} \right| \geq \varepsilon \right)\\
            &\leq \exp(-\Omega(\tau^2)) + 2\exp(-\frac{d_1^2 \varepsilon^2}{C p^2 \tau^2})\nonumber,
        \end{split}
    \end{equation}
    \end{small}\noindent
    with some universal constant $C > 0$.
    Set $\varepsilon_{d_1} = (p/d_1) \tau$. Then,
    $$
    \P \Big(\Big|\frac{Z(\bc, \beps) - Z}{Z}\Big| \geq 2\varepsilon_{d_1} \Big) \leq \exp(-\Omega(\tau^2)).
    $$
    This concludes the proof.
\end{proof}

\begin{lem}[Modification to Lemma A.4 from \cite{xu2023inference}]\label{lem: A4}
    Suppose that Assumptions~\ref{asm: signal covariance}, \ref{asm: regime} and \ref{asm: noise covariance} hold.
    For $w \in \calW$, $M, M' \in \{1, 2\}$, define
    \begin{align}
        \bar \bv^{(M)} &\triangleq \frac{1}{Z^{(M)}} \sum_{w' \in \calW^{(M)}} \exp(\frac{\|\v_{w'}^\star\|^2}{2p} + \frac{\sigma_{w',w'}}{2}) \v_{w'}^\star,\label{eq: dfn v1}\\
        \bar \sigma^{(M)}_w &\triangleq \frac{1}{Z^{(M)}} \sum_{w' \in \calW^{(M)}} \exp(\frac{\|\v_{w'}^\star\|^2}{2p} + \frac{\sigma_{w',w'}}{2}) \sigma_{w,w'},\nonumber\\
        \bar \sigma^{(M, M')} &\triangleq \frac{1}{Z^{(M)}} \sum_{w' \in \calW^{(M)}} \exp(\frac{\|\v_{w'}^\star\|^2}{2p} + \frac{\sigma_{w',w'}}{2}) \bar \sigma^{(M')}_{w'}.\nonumber
    \end{align}
    Then, for $w \in \calW^{(M)}$,
    \begin{align*}
        &\abs{\log p_w^{(M)} - \qty(\frac{1}{2p} \|\v_w^\star\|^2 + \frac{1}{2} \sigma_{w,w} - \frac{1}{p} \v_w^{\star \top} \bar \bv^{(M)} - \bar \sigma^{(M)}_w + \frac{1}{p} \|\bar \bv^{(M)}\|^2 + \bar \sigma^{(M, M)} - \log Z^{(M)})}\\
        &\quad\lesssim \frac{p^2}{d_M^2} \tau.
    \end{align*}
\end{lem}

\begin{proof}[Proof of Lemma~\ref{lem: A4}]
    Without loss of generality, we focus on $M=1$.
    We omit the superscript $(\cdot)^{(1)}$ from $p_w^{(1)}$, $\calW^{(1)}$ and $Z^{(1)}$.
    For $w \in \calW$, denote $A_w = \exp(\v_w^{\star \top}\bc + \b_w^\top \beps')$,
    where $\b_w \triangleq \bSigma_1^{1/2} \be_w$ and $\beps' \sim N(\zero, \Ib_{d_1})$.
    Let $Z(\c, \beps) \triangleq \sum_{w \in \calW} \exp(\v_w^{\star \top} \c + \epsilon_w)$.
    Define
    \begin{align*}
        f_4(\bX, \bZ) \triangleq \exp(\frac{\|\bX + \v_w^\star\|^2}{2p} + \frac{\|\bZ + \b_w\|^2}{2}).
    \end{align*}
    Also define
    \begin{align*}
        f_5(\bX, \bZ) \triangleq f_4(\bX, \bZ)/f_4(\zero, \zero) = \exp\left(\frac{\|\bX\|^2+2\bX^\top\v_w^\star}{2p} + \frac{\|\bZ\|^2+2\bZ^\top \b_w}{2} \right).
    \end{align*}

    Then we have
    \begin{align*}
        \E \left[A_w\right] &= f_4(\zero, \zero),\\
        \E \left[A_w Z(\bc, \beps) \right] &= \E\left[\sum_{w'}\exp\left((\v_w^\star + \v_{w'}^\star)^\top\bc + (\b_w + \b_{w'})^\top\beps \right) \right] = \sum_{w'}f_4(\v_{w'}^\star, \b_{w'}),\\
        \E \left[A_w Z^2(\bc, \beps) \right] &= \E\left[\sum_{w',w''}\exp\left(\left(\v_w^\star + \v_{w'}^\star + \v_{w''}^\star \right)^\top\bc + \left(\b_w + \b_{w'} + \b_{w''} \right)^\top\beps \right) \right]\\
        &= \sum_{w',w''}f_4(\v_{w'}^\star+\v_{w''}^\star, \b_{w'}+\b_{w''}).
    \end{align*}
    Denote an event $\cF = \{|Z(\bc, \beps)/Z-1| < \varepsilon_{d_1}\}$,
    where $\varepsilon_{d_1} = 2p\tau/d_1$. By Lemma~\ref{lem: A3}, we know $\P(\cF)=1-\exp (-\Omega(\tau^2)).$
    For any integer $m>0$, we have the following decomposition:
    \begin{equation}\label{eq:Z_decom}
        \frac{1}{Z(\bc, \beps)} = \sum_{k=1}^m\frac{(Z-Z(\bc, \beps))^{k-1}}{Z^k} + \frac{(Z-Z(\bc, \beps))^m}{Z(\bc, \beps)Z^m}.
    \end{equation}
    Denote 
    \begin{equation*}
        g_{m,w} = \sum_{k=1}^m\frac{\E \big[ A_w  (Z-Z(\bc, \beps))^{k-1} \big]}{Z^k}, \ \ 
        b_{m,w} = \E \Big[\frac{A_w(Z-Z(\bc, \beps))^m}{Z(\bc, \beps)Z^m} \Big], m \geq 1.
    \end{equation*}
    Then we have $p_w = g_{m,w}+b_{m,w}$ for any integer $m > 0$. When $m = 2k+1, k \geq 0$, we have
    \begin{equation*}
         \begin{split}
             b_{2k+1,w} &= \E \Big[\frac{A_w(Z-Z(\bc, \beps))^{2k+1}}{Z(\bc, \beps)Z^{2k+1}} I(F) \Big] + \E \Big[\frac{A_w(Z-Z(\bc, \beps))^{2k+1}}{Z(\bc, \beps)Z^{2k+1}}I(F^c) \Big]\\
             &\leq \E \Big[\frac{A_w \varepsilon_{d_1}^{2k+1}}{Z(1-\varepsilon_{d_1})} I(F) \Big] + \E \Big[\frac{A_w}{Z(\bc, \beps)}I(F^c) \Big] \leq \E \Big[\frac{A_w \varepsilon_{d_1}^{2k+1}}{Z(1-\varepsilon_{d_1})} \Big] + \P(\cF^c)\\
             &= \frac{g_{1,w}\varepsilon_{d_1}^{2k+1}}{1-\varepsilon_{d_1}} + \exp(-\Omega(\tau^2)),
         \end{split}
    \end{equation*}    
    where the first inequality comes from the definition of set $F$ and $(Z-Z(\bc, \beps))^{2k+1} \leq Z^{2k+1}$, and the second inequality is due to $A_w = \exp\big(\langle \v_w^\star, \bc \rangle + \langle \b_w, \beps \rangle\big) \leq Z(\bc, \beps)$ and $I(F) \leq 1.$ On the other side, since $Z - Z(\bc,\beps) \geq - Z(\bc, \beps)$, $-Z(\bc, \beps) \leq 0$ and $2k+1$ is an odd number, we obtain
    \begin{align*}
        \frac{A_w(Z-Z(\bc, \beps))^{2k+1}}{Z(\bc, \beps)Z^{2k+1}} \geq \frac{A_w(-Z(\bc, \beps))^{2k+1}}{Z(\bc, \beps)Z^{2k+1}} \geq \frac{(-Z(\bc, \beps))^{2k+1}}{Z^{2k+1}},
    \end{align*}
    where the second inequality follows from $A_w/Z(\bc, \beps) \leq 1$.
    Thus we have
    \begin{equation*}
            b_{2k+1,w} \geq \E \Big[\frac{A_w(Z-Z(\bc, \beps))^{2k+1}}{Z(\bc, \beps)Z^{2k+1}}I(F^c) \Big] \geq \E \Big[\frac{(-Z(\bc, \beps))^{2k+1}}{Z^{2k+1}}I(F^c) \Big].
    \end{equation*}
    By Cauchy-Schwarz inequality and $Z \geq d_1$, we have
    \begin{equation*}
            b_{2k+1,w} \geq - \frac{\sqrt{\E[(Z(\bc, \beps))^{4k+2}] \P(\cF^c)}}{Z^{2k+1}} \geq - \frac{\exp(-\Omega(\tau^2))\sqrt{\E[(Z(\bc, \beps))^{4k+2}]}}{d_1^{2k+1}}.
    \end{equation*}
    By H\"older's inequality, we have $(\sum_{j=1}^{d_1} x_j)^m \leq d_1^{m-1}\sum_{j=1}^{d_1} |x_j|^m$. Thus we have
    \begin{align*}
        \E[(Z(\bc, \beps))^{4k+2}] &\leq d_1^{4k+1}\sum_{w\in \calW} \E[\exp((4k+2) \langle \v_w^\star, \bc \rangle + (4k+2) \langle \b_w, \beps \rangle)]\\
        &= d_1^{4k+1}\sum_{w\in\calW} \exp \Big(\frac{(4k+2)^2\|\v_w^\star\|^2}{2p} + \frac{(4k+2)^2\|\b_w\|^2}{2}\Big).
    \end{align*}
    By Assumptions~\ref{asm: signal covariance} and \ref{asm: noise covariance}, we have
    \begin{equation*}
       \E[(Z(\bc, \beps))^{4k+2}] \lesssim d_1^{4k+2} \exp(C (4k+2)^2 \frac{p}{d_1})
    \end{equation*}
    for some constant $C > 0$.
    It yields that when $m^2 p \ll d_1 \tau^2$,
    \begin{equation*}
         b_{2k+1,w} \geq - \exp(-\Omega(\tau^2) + C (4k+2)^2 \frac{p}{d_1}) = - \exp(-\Omega(\tau^2)).
    \end{equation*}
    Combining the properties of $b_{m,w}$ when $m$ is odd, we have
    \begin{equation*}
        g_{m,w} - \exp(-\Omega(\tau^2)) \leq p_w \leq g_{m,w} + \frac{g_{1,w}\varepsilon_{d_1}^{m}}{1-\varepsilon_{d_1}} + \exp(-\Omega(\tau^2)).
    \end{equation*}
    Thus, for $m = 1$, it follows that
    \begin{equation*}
        |p_w - g_{1,w}| \lesssim g_{1,w} \frac{p\tau}{d_1} + \exp(-\Omega(\tau^2)) \ll g_{1,w},
    \end{equation*}
    where the last inequality follows from $d_1 \gg p \tau$ by Assumption~\ref{asm: regime} and
    \begin{align*}
        g_{1,w} = \frac{f_4(\zero, \zero)}{Z} \geq \frac{1}{Z} \geq \frac{1}{d_1 \exp((1/2p) \|\Vb_1^\star\|_{2,\infty}^2 + (1/2) \max_{w \in \calW} \sigma_{w,w})} \gtrsim \frac{1}{d_1} \gg \exp(-\Omega(\tau^2)).
    \end{align*}
    We set $m = m^{\star} := 2\lfloor \sqrt{\frac{d_1\tau^2}{p}} \rfloor + 1$, and correspondingly we have for any $m \leq m^{\star}$,
    \begin{equation}\label{ineq:pw_gmw}
        |p_w - g_{m,w}| \leq Cp_w\big(\varepsilon_{d_1}^{m}+\exp(-\Omega(\tau^2))\big) \triangleq C p_w \varepsilon_{d_1}'^m
    \end{equation}
    since $\min_{w \in \calW} p_w \geq (1/2) \min_{w \in \calW} g_{1,w} \gtrsim 1/d_1$. Then we have
    \begin{equation}\label{ineq:logpw_gmw}
        |\log p_w - \log g_{m,w}| \leq \max \big(\log(1+C\varepsilon_{d_1}'^{m}), -\log(1-C\varepsilon_{d_1}'^{m})\big) \leq 2 C \varepsilon_{d_1}'^{m}.
    \end{equation}
    Next we calculate $g_{m,w}$. Note that
    \begin{equation*}
        \sum_{k=1}^m\frac{(Z-Z(\bc, \beps))^{k-1}}{Z^k} = \frac{1}{Z(\bc, \beps)}\Big[ 1 - \Big(1 - \frac{Z(\bc, \beps)}{Z} \Big)^m\Big] = \sum_{k=1}^m \binom{m}{k}(-1)^{k-1}\frac{Z(\bc, \beps)^{k-1}}{Z^k}.
    \end{equation*}
    We have 
    \begin{equation*}
        \begin{split}
            g_{m,w} &= \sum_{k=1}^m \binom{m}{k}\frac{(-1)^{k-1}}{Z^k}\E\big[A_w Z(\bc, \beps)^{k-1} \big]\\
            &=\sum_{k=1}^m \binom{m}{k}\frac{(-1)^{k-1}}{Z^k} \sum_{w^{[1]},\cdots,w^{[k-1]}}f_4\Big(\sum_{j=1}^{k-1} \v_{w^{[j]}}, \sum_{j=1}^{k-1} \b_{w^{[j]}}\Big)\\
            &=\frac{f_4(\zero, \zero)}{Z} \Big[ \sum_{k=1}^m\binom{m}{k}\frac{(-1)^{k-1}}{Z^{k-1}}\sum_{w^{[1]},\cdots,w^{[k-1]}} f_5 \Big(\sum_{j=1}^{k-1} \v_{w^{[j]}}, \sum_{j=1}^{k-1} \b_{w^{[j]}}\Big)\Big].
        \end{split}
    \end{equation*}
    Note that $f_5(\zero, \zero) = 1$, and
    \begin{equation*}
        \sum_{k=1}^m\binom{m}{k}\frac{(-1)^{k-1}}{Z^{k-1}}Z^{k-1} = \sum_{k=1}^m\binom{m}{k}(-1)^{k-1} = 1 -\sum_{k=0}^m\binom{m}{k}(-1)^{k} = 1.
    \end{equation*}
    We can further transform $g_{m,w}$ to
    \begin{equation}\label{eq:gmw_transform}
        g_{m,w}= \frac{f_4(\zero, \zero)}{Z} \Big[1 +  \sum_{k=2}^m\binom{m}{k}\frac{(-1)^{k}}{Z^{k-1}}\Big(Z^{k-1} - \sum_{w^{[1]},\cdots,w^{[k-1]}} f_5 \Big(\sum_{j=1}^{k-1} \v_{w^{[j]}}, \sum_{j=1}^{k-1} \b_{w^{[j]}}\Big)\Big)\Big].
    \end{equation}
    To make sure that the transformation of $\Vb$ preserves the low rank structure, we take $m = 3$.
    Combining \eqref{ineq:logpw_gmw} and \eqref{eq:gmw_transform}, we obtain
    \begin{small}
    \begin{equation}
        \abs{\log p_w - \log\frac{f_4(\zero, \zero)}{Z}- \log\Big(1 + 3\underbrace{\frac{Z - \sum_{w'}f_5(\v_{w'}^\star, \b_{w'})}{Z}}_{\triangleq u_1} + \underbrace{\frac{\sum_{w',w''}f_5(\v_{w'}^\star+\v_{w''}^\star, \b_{w'}+\b_{w''})-Z^2}{Z^2}}_{\triangleq u_2} \Big) } \leq C \varepsilon_{d_1}'^3.\label{eq: log p w taylor}
    \end{equation}
    \end{small}\noindent
    For the term $u_1$, by Taylor expansion, we have
    \begin{equation}
        \begin{split}
            \frac{Z - \sum_{w'}f_5(\v_{w'}^\star, \b_{w'})}{Z} &= -\frac{\sum_{w'}\exp(\|\v_{w'}^\star\|^2/2p + \|\b_{w'}\|^2/2)\big(\exp( \v_w^{\star \top}\v_{w'}^\star/p + \b_w^\top\b_{w'} )-1 \big)}{Z}\\
            &= -\sum_{w'}\frac{\exp(\|\v_{w'}^\star\|^2/2p + \|\b_{w'}\|^2/2)}{Z}\Big[\frac{\v_w^{\star \top}\v_{w'}^\star}{p} + \b_w^\top\b_{w'} + O\qty(\frac{p}{d_1^2} + \frac{p^2}{d_1^2} )\Big]\\
            &= -\frac{\v_w^{\star \top}\overline{\bv}^{(1)}}{p} - \bar \sigma_w^{(1)} + O\qty(\frac{p^2}{d_1^2}),
        \end{split}
    \end{equation}
    where the equality follows from $(\v_w^{\star \top}\v_{w'}^\star)^2/p = O(p/d_1^2)$ and $\max_{w \in \calW} \sigma_{w,w}^2 = O(p^2/d_1^2)$ from Assumptions~\ref{asm: signal covariance} and \ref{asm: noise covariance}.
    For the term $u_2$, we have
    \begin{small}
    \begin{equation}
        \begin{split}
            &\frac{\sum_{w',w''}f_5(\v_{w'}^\star+\v_{w''}^\star, \b_{w'}+\b_{w''})-Z^2}{Z^2}\\
            &\quad= \frac{\sum_{w',w''}\exp\left(\frac{\|\v_{w'}^\star\|^2+\|\v_{w''}^\star\|^2}{2p} + \frac{\|\b_{w'}\|^2+\|\b_{w''}\|^2}{2}\right)}{Z^2}\\
            &\quad\quad\times \left(\exp\left( \frac{\v_{w'}^{\star \top}\v_{w''}^\star+\v_w^{\star \top}\left(\v_{w'}^\star+\v_{w''}^\star\right)}{p} + \b_{w'}^\top\b_{w''}+\b_w^\top\left(\b_{w'}+\b_{w''}\right)\right)-1 \right)\\
            &\quad= \sum_{w',w''}\frac{\exp\left(\frac{\|\v_{w'}^\star\|^2+\|\v_{w''}^\star\|^2}{2p} + \frac{\|\b_{w'}\|^2+\|\b_{w''}\|^2}{2}\right)}{Z^2}\\
            &\quad\quad\times \qty(\frac{\v_{w'}^{\star \top}\v_{w''}^\star+\v_w^{\star \top}\left(\v_{w'}^\star+\v_{w''}^\star\right)}{p} + \b_{w'}^\top\b_{w''}+\b_w^\top\left(\b_{w'}+\b_{w''}\right)) + O\qty(\frac{p^2}{d_1^2})\\
            &\quad= \frac{\overline{\bv}^{(1) \top}\overline{\bv}^{(1)}+2\v_w^{\star \top}\overline{\bv}^{(1)}}{p} + \bar\sigma^{(1)}+2\bar \sigma_w^{(1)} + O\qty(\frac{p^2}{d_1^2}).
        \end{split}
    \end{equation}
    \end{small}\noindent
    Again using Taylor expansion, we have
    \begin{align*}
        \left|\log p_w -  \left(\frac{\|\v_w^\star\|^2}{2p} + \frac{\sigma_{w,w}}{2} - \log Z + \frac{\overline{\bv}^{(1) \top}\overline{\bv}^{(1)}-\v_w^{\star \top}\overline{\bv}^{(1)}}{p} + \bar\sigma^{(1)} -\bar\sigma_w^{(1)} \right) \right| &\leq 4\varepsilon_{d_1}'^3 + O\qty(\frac{p^2}{d_1^2})\\
        &\lesssim \frac{p^3}{d_1^3} \tau^3 + \frac{p^2}{d_1^2}\\
        &\lesssim \frac{p^2}{d_1^2} \tau.
    \end{align*}
    Here the last inequality follows from $d_1 \gg p \tau^2$ by Assumption~\ref{asm: regime}.
\end{proof}

\begin{lem}[Modification to Lemma A.5 from \cite{xu2023inference}]\label{lem: A5}
    Suppose that Assumptions~\ref{asm: signal covariance}, \ref{asm: regime} and \ref{asm: noise covariance} hold.
    Let $\bar \bv^{(M)}$, $\bar\sigma_w^{(M)}$ and $\bar\sigma^{(M, M')}$ be the quantities defined in Lemma~\ref{lem: A4}.
    Then, for $w \in \calW^{(M)}$ and $w' \in \calW^{(M')}$,
    \begin{small}
    \begin{align*}
            &\biggl|\log p_{w, w'}^{(M,M')} - \biggl(\frac{\|\v_w^\star\|^2 + \|\v_{w'}^\star\|^2 + 2\v_w^{\star \top}\v_{w'}^\star}{2p} + \frac{\sigma_{w,w} + \sigma_{w',w'} + 2\sigma_{w,w'}}{2} - \log Z^{(M)} - \log Z^{(M')}\\\
            &\quad\quad+ \frac{1}{p} \bar \bv^{(M) \top} \bar \bv^{(M)} + \frac{1}{p} \bar \bv^{(M') \top} \bar \bv^{(M')} - \frac{1}{p} \v_w^{\star \top} \bar \bv^{(M)} - \frac{1}{p} \v_{w'}^{\star \top} \bar \bv^{(M')} - \frac{1}{p} \v_w^{\star \top} \bar \bv^{(M')} - \frac{1}{p} \v_{w'}^{\star \top} \bar \bv^{(M)} + \frac{1}{p} \bar \bv^{(M) \top} \bar \bv^{(M')}\\
            &\quad\quad+ \bar \sigma^{(M, M)} + \bar \sigma^{(M', M')} - \bar \sigma_w^{(M)} - \bar \sigma_{w'}^{(M')} - \bar \sigma_w^{(M')} - \bar \sigma_{w'}^{(M)} + \bar \sigma^{(M, M')}\biggr)\biggr|\\
            &\quad\lesssim \frac{p^2}{(d_M \wedge d_{M'})^2} \tau.\label{eq: cross co-occurrence probability approximation}
    \end{align*}
    \end{small}\noindent
\end{lem}

\begin{proof}
    The proof is an analogy to the proof of Lemma~\ref{lem: A4}. Let $\b_w \triangleq \bSigma^{1/2} \be_w$. 
    Fix $w \in \calW^{(M)}, w' \in \calW^{(M')}$. Define $f_1$, $f_2$, and $f_3$ as
    \begin{align*}
        f_1(\bX, \bY, \bZ, \bW) &\triangleq \exp(\frac{\|\v_w^\star + \bX\|^2 + \|\v_{w'}^\star + \bY\|^2 + 2(\v_w^\star + \bX)^\top (\v_{w'}^\star + \bY)}{2p})\\
        &\quad\times \exp(\frac{\|\b_w + \bZ\|^2 + \|\b_{w'} + \bW\|^2 + 2 (\b_w + \bZ)^\top (\b_{w'} + \bZ)}{2}),\\
        f_2(\bX, \bY, \bZ, \bW) &\triangleq f_1(\bX, \bY, \bZ, \bW) / f_1(\zero, \zero, \zero, \zero),\\
        f_3(\bX, \bY, \bZ, \bW) &\triangleq \exp(\frac{2 \v_w^{\star \top} \bX + 2 \v_{w'}^{\star \top} \bY + 2 (\v_w^{\star \top} \bY + \bX^\top \v_{w'}^\star + \bX^\top \bY)}{2p})\\
        &\quad\times \exp(\frac{2 \b_w^\top \bZ + 2 \b_{w'}^\top \bW + 2 (\b_w^\top \bW + \bZ^\top \b_{w'} + \bZ^\top \bW)}{2}).
    \end{align*}
    Denote $A_w \triangleq \exp(\v_w^{\star \top}\bc + \b_w^\top\beps)$ and $Z^{(M)}(\c, \beps) \triangleq \sum_{w \in \calW^{(M)}} A_w$.
    Then we have
    \begin{equation*}
        \begin{split}
            \E[A_w A_{w'}] &= \mathbb{E}\Big[\exp\left(\left(\v_w^\star + \v_{w'}^\star \right)^\top\bc + \left(\b_w + \b_{w'} \right)^\top\beps \right)\Big]\\
            &= \exp\Big( \frac{\|\v_w^\star\|^2 + \|\v_{w'}^\star\|^2 + 2\v_w^{\star \top}\v_{w'}^\star}{2p} + \frac{\|\b_w\|^2 + \|\b_{w'}\|^2 + 2\b_w^\top\b_{w'}}{2} \Big) = f_1(\zero,\zero,\zero,\zero)
        \end{split}
    \end{equation*}
    and
    \begin{small}
    \begin{align*}
        \E[A_w A_{w'} Z^{(M)}(\bc, \beps) Z^{(M')}(\bc, \beps)] &= \sum_{\substack{w^{[1]} \in \calW^{(M)}\\w'^{[1]} \in \calW^{(M')}}} f_1(\v_{w^{[1]}}^\star,\v_{w'^{[1]}}^\star,\b_{w^{[1]}},\b_{w'^{[1]}}),\\
        \E[A_w A_{w'}Z^{(M) 2}(\bc, \beps)Z^{(M') 2}(\bc, \beps)] &= \sum_{\substack{w^{[1]},w^{[2]} \in \calW^{(M)}\\w'^{[1]},w'^{[2]} \in \calW^{(M')}}} f_1(\v_{w^{[1]}}^\star+\v_{w^{[2]}}^\star, \v_{w'^{[1]}}^\star+\v_{w'^{[2]}}^\star, \b_{w^{[1]}}+\b_{w^{[2]}}, \b_{w'^{[1]}}+\b_{w'^{[2]}}).
    \end{align*}
    \end{small}\noindent
    By Lemma~\ref{lem: A3}, we know that $\P(\cF)= 1 - \exp(-\Omega(\tau^2))$, where 
    $$
    \cF=\left\{\left|\frac{Z^{(M)}(\boldsymbol{c}, \beps)}{Z^{(M)}}-1\right|<\varepsilon_{d_M},\left|\frac{Z^{(M')}\left( \bc, \beps \right)}{Z^{(M')}}-1\right|<\varepsilon_{d_{M'}}\right\}.
    $$
    Here $\varepsilon_{d_M} = 2 (p/d_M) \tau$. Similar to \eqref{eq:Z_decom} in Lemma~\ref{lem: A3}, we have the following decomposition:
    \begin{align*}
       \frac{1}{Z^{(M)}(\bc, \beps)Z^{(M')}(\bc, \beps)} &= \sum_{k=1}^m \biggl\{ \frac{(Z^{(M)} Z^{(M')} - Z^{(M)}(\bc, \beps) Z^{(M')}(\bc, \beps))^{k-1}}{Z^{(M) k} Z^{(M') k}}\\
       &\quad\quad+ \frac{(Z^{(M)} Z^{(M')} - Z^{(M)}(\bc, \beps) Z^{(M')}(\bc, \beps))^m}{Z^{(M)}(\bc, \beps) Z^{(M')}(\bc, \beps)Z^{(M) m} Z^{(M') m}} \biggr\}.
    \end{align*}
    To simplify the expression, we denote 
    \begin{align*}
        g_{m,w,w'} &= \sum_{k=1}^m\frac{\E \big[ A_w A_{w'} (Z^{(M)} Z^{(M')}-Z^{(M)}(\bc, \beps)Z^{(M')}(\bc, \beps))^{k-1} \big]}{Z^{(M) k} Z^{(M') k}},\\
        b_{m,w,w'} &= \E \Big[\frac{A_w A_{w'}(Z^{(M)} Z^{(M')} - Z^{(M)}(\bc, \beps) Z^{(M')}(\bc, \beps))^m}{Z^{(M)}(\bc, \beps) Z^{(M')}(\bc, \beps)Z^{(M) m} Z^{(M') m}} \Big], \ \ m \geq 1.
    \end{align*}
    Then we have $p_{w,w'}^{(M,M')} = g_{m,w,w'}+b_{m,w,w'}$ for any integer $m > 0$. When $m = 2k+1, k \geq 0$, following the same procedure in Lemma~\ref{lem: A4}, we have
    \begin{equation*}
        -\exp(-\Omega(\tau^2)) \leq b_{m,w,w'} \leq \frac{g_{1,w,w'}(3\varepsilon)^{m}}{(1-\varepsilon_{d_1})^2} + \exp(-\Omega(\tau^2)),
    \end{equation*}
    where $\varepsilon = \varepsilon_{d_M} + \varepsilon_{d_{M'}} + \varepsilon_{d_M} \varepsilon_{d_{M'}}$.
    By a similar argument as in the proof of Lemma~\ref{lem: A4}, we have
    \begin{equation}\label{ineq:p_ww'u_two_side_bound}
        |p_{w,w'}^{(M,M')} - g_{m,w,w'}| \leq 2g_{1,w,w'} (3\varepsilon)^{m} + \exp(-\Omega(\tau^2)) \triangleq 2 g_{1,w,w'} \varepsilon'^m
    \end{equation}
    since both $p_{w,w'}^{(M,M')}$ and $g_{1,w,w'}$ are of order $1/(d_M d_{M'})$.  Then, we have
    \begin{equation}\label{ineq:logpww_gmww}
        |\log p_{w,w'}^{(M,M')} - \log g_{m,w,w'}| \leq \max \big(\log(1+C\varepsilon'^{m}), -\log(1-C\varepsilon'^{m})\big) \leq 2C\varepsilon'^{m}. 
    \end{equation}
    Next we calculate $g_{m,w}$. Note that
    \begin{equation*}
        \sum_{k=1}^m \frac{(Z^{(M)} Z^{(M')} - Z^{(M)}(\bc, \beps) Z^{(M')}(\bc, \beps))^{k-1}}{Z^{(M) k} Z^{(M') k}} = \sum_{k=1}^m \binom{m}{k}(-1)^{k-1}\frac{(Z^{(M)}(\bc, \beps)Z^{(M')}(\bc, \beps))^{k-1}}{Z^{(M) k} Z^{(M') k}}.
    \end{equation*}
    Then we have
    \begin{small}
    \begin{equation*}
        \begin{split}
            g_{m,w,w'} &= \sum_{k=1}^m \binom{m}{k}\frac{(-1)^{k-1}}{Z^{(M) k} Z^{(M') k}}\E\big[A_w A_{w'} (Z^{(M)}(\bc, \beps)Z^{(M')}(\bc, \beps))^{k-1} \big]\\
            &=\sum_{k=1}^m \binom{m}{k}\frac{(-1)^{k-1}}{Z^{(M) k} Z^{(M') k}}\\
            &\quad\times \sum_{\substack{w^{[1]},\cdots,w^{[k-1]} \in \calW^{(M)}\\ w'^{[1]},\cdots,w'^{[k-1]} \in \calW^{(M')}}} f_1\Big(\sum_{j=1}^{k-1} \v_{w^{[j]}},\sum_{j=1}^{k-1} \v_{w'^{[j]}}, \sum_{j=1}^{k-1} \b_{w^{[j]}},\sum_{j=1}^{k-1} \b_{w'^{[j]}}\Big)\\
            &=\frac{f_1(\zero,\zero,\zero,\zero)}{Z^{(M)} Z^{(M')}} \Big[ \sum_{k=1}^m\binom{m}{k}\frac{(-1)^{k-1}}{Z^{(M) k-1} Z^{(M') k-1}}\\
            &\quad\times \sum_{\substack{w^{[1]},\cdots,w^{[k-1]} \in \calW^{(M)}\\ w'^{[1]},\cdots,w'^{[k-1]} \in \calW^{(M')}}} f_2 \Big(\sum_{j=1}^{k-1} \v_{w^{[j]}}, \sum_{j=1}^{k-1} \v_{w'^{[j]}}, \sum_{j=1}^{k-1} \b_{w^{[j]}}, \sum_{j=1}^{k-1} \b_{w'^{[j]}}\Big)\Big].
        \end{split}
    \end{equation*}
    \end{small}\noindent
    Note that $f_2(\zero,\zero,\zero,\zero) = 1$, and
    \begin{equation*}
        \sum_{k=1}^m\binom{m}{k}\frac{(-1)^{k-1}}{Z^{(M) k-1} Z^{(M') k-1}} Z^{(M) k-1} Z^{(M') k-1} = \sum_{k=1}^m\binom{m}{k}(-1)^{k-1} = 1 -\sum_{k=0}^m\binom{m}{k}(-1)^{k} = 1.
    \end{equation*}
    We can further equivalently write $g_{m,w,w'}$ as
    \begin{align*}
        g_{m,w,w'} &= \frac{f_1(\zero,\zero,\zero,\zero)}{Z^{(M)} Z^{(M')}} \Big[1 + \sum_{k=2}^m\binom{m}{k}\frac{(-1)^{k}}{Z^{(M) k-1} Z^{(M') k-1}}\Big(Z^{(M) k-1} Z^{(M') k-1}\\
        &\quad- \sum_{\substack{w^{[1]},\cdots,w^{[k-1]} \in \calW^{(M)}\\ w'^{[1]},\cdots,w'^{[k-1]} \in \calW^{(M')}}} f_2 \Big(\sum_{j=1}^{k-1} \v_{w^{[j]}}, \sum_{j=1}^{k-1} \v_{w'^{[j]}}, \sum_{j=1}^{k-1} \b_{w^{[j]}}, \sum_{j=1}^{k-1} \b_{w'^{[j]}}\Big)  \Big) \Big].
    \end{align*}
    Here we let $m = 3$. 
    Then \eqref{ineq:logpww_gmww} yields that
    \begin{equation}\label{eq:pww'_leading_term}
        \begin{split}
           &\left|\log p_{w,w'}^{(M,M')} - \log \frac{f_1(\zero,\zero,\zero,\zero)}{Z^{(M)} Z^{(M')}} - \log\Big(3 - 3\frac{\sum_{w^{[1]} \in \calW^{(M)},w'^{[1]} \in \calW^{(M')}} f_2(\v_{w^{[1]}}^\star,\v_{w'^{[1]}}^\star,\b_{w^{[1]}},\b_{w'^{[1]}})}{Z^{(M)} Z^{(M')}}\right.\\
           &\quad+ \left.\frac{\sum_{w^{[1]},w^{[2]}  \in \calW^{(M)},w'^{[1]},w'^{[2]} \in \calW^{(M')}}f_2(\v_{w^{[1]}}^\star+\v_{w^{[2]}}^\star, \v_{w'^{[1]}}^\star+\v_{w'^{[2]}}^\star,\b_{w^{[1]}}+\b_{w^{[2]}}, \b_{w'^{[1]}}+\b_{w'^{[2]}})}{Z^{(M) 2} Z^{(M') 2}} \Big) \right| \leq C \varepsilon'^3
        \end{split}
    \end{equation}
    for sufficiently small $\varepsilon'$. 
    Note that $Z^{(M)} Z^{(M')} = \sum_{w^{[1]},w'^{[1]}}\exp\left(\frac{\|\v_{w^{[1]}}^\star\|^2 + \|\v_{w'^{[1]}}^\star\|^2}{2p} + \frac{\|\b_{w^{[1]}}\|^2 + \|\b_{w'^{[1]}}\|^2}{2}\right)$. By Taylor expansion,
    \begin{equation}\label{eq:taylor_1}
        \begin{split}
            &\frac{Z^{(M)} Z^{(M')} - \sum_{w^{[1]} \in \calW^{(M)},w'^{[1]} \in \calW^{(M')}} f_2(\v_{w^{[1]}}^\star,\v_{w'^{[1]}}^\star,\b_{w^{[1]}},\b_{w'^{[1]}}) }{Z^{(M)} Z^{(M')}}\\
            &\quad= -\frac{\sum_{w^{[1]} \in \calW^{(M)},w'^{[1]} \in \calW^{(M')}} \exp\big(\frac{\|\v_{w^{[1]}}^\star\|^2 + \|\v_{w'^{[1]}}^\star\|^2}{2p} + \frac{\|\b_{w^{[1]}}\|^2 + \|\b_{w'^{[1]}}\|^2}{2}\big) }{Z^{(M)} Z^{(M')}}\\
            &\quad\quad\times \left( f_3(\v_{w^{[1]}}^\star,\v_{w'^{[1]}}^\star,\b_{w^{[1]}},\b_{w'^{[1]}}) - 1\right)\\
            &\quad= -\sum_{w^{[1]} \in \calW^{(M)},w'^{[1]} \in \calW^{(M')}}\frac{\exp\left(\frac{\|\v_{w^{[1]}}^\star\|^2 + \|\v_{w'^{[1]}}^\star\|^2}{2p} + \frac{\|\b_{w^{[1]}}\|^2 + \|\b_{w'^{[1]}}\|^2}{2}\right)}{Z^{(M)} Z^{(M')}}\\
            &\quad\quad\times \qty(\frac{ 2\v_w^{\star \top}\v_{w^{[1]}}^\star+ 2\v_{w'}^{\star \top}\v_{w'^{[1]}}^\star}{2p} + \frac{ 2\b_w^\top\b_{w^{[1]}}+ 2\b_{w'}^\top\b_{w'^{[1]}}}{2} + S(w,w'))\\
            &\quad\quad+ O\qty(\frac{\|\Vb^\star\|_{2,\infty}^4}{p^2} + \max_{w \in \calW} \sigma_{w,w}^2)\\
            &\quad= -\frac{1}{p} \v_w^{\star \top} \bar \bv^{(M)} - \frac{1}{p} \v_{w'}^{\star \top} \bar \bv^{(M')} - \frac{1}{p} \v_w^{\star \top} \bar \bv^{(M')} - \frac{1}{p} \v_{w'}^{\star \top} \bar \bv^{(M)} - \frac{1}{p} \bar \bv^{(M) \top} \bar \bv^{(M')}\\
            &\quad\quad- \bar\sigma_w^{(M)} - \bar\sigma_{w'}^{(M')} - \bar\sigma_{w'}^{(M)} - \bar\sigma_{w}^{(M')} - \bar\sigma^{(M, M')} + O\qty(\frac{p^2}{(d_M \wedge d_{M'})^2}),
        \end{split}
    \end{equation}
    where we used Assumptions~\ref{asm: signal covariance} and \ref{asm: noise covariance}, and $S_{w,w'}(w^{[1]},w'^{[1]}) \triangleq \v_w^{\star \top}\v_{w'^{[1]}}^\star/p + \v_{w'}^{\star \top}\v_{w^{[1]}}^\star/p + \v_{w^{[1]}}^{\star \top}\v_{w'^{[1]}}^\star/p + \b_w^\top\b_{w'^{[1]}} + \b_{w'}^\top\b_{w^{[1]}} + \b_{w^{[1]}}^\top\b_{w'^{[1]}}$.
    Denote
    \begin{small}
    \begin{equation*}
        \begin{split}
           A_w^{(4)} &= \frac{1}{p} \|\v_{w^{[1]}}^\star\|^2+\frac{1}{p} \|\v_{w^{[2]}}^\star\|^2+\frac{1}{p} \|\v_{w'^{[1]}}^\star\|^2+\frac{1}{p} \|\v_{w'^{[2]}}^\star\|^2 +\|\b_{w^{[1]}}\|^2+\|\b_{w^{[2]}}\|^2+\|\b_{w'^{[1]}}\|^2+\|\b_{w'^{[2]}}\|^2,\\
           B_w^{(4)} &= \frac{1}{p} \langle\v_{w^{[1]}}^\star, \v_{w^{[2]}}^\star\rangle + \frac{1}{p} \langle\v_{w'^{[1]}}^\star, \v_{w'^{[2]}}^\star\rangle + \langle\b_{w^{[1]}}, \b_{w^{[2]}}\rangle + \langle\b_{w'^{[1]}}, \b_{w'^{[2]}}\rangle.
        \end{split}
    \end{equation*}
    \end{small}\noindent
    Then we transform the second term into
    \begin{small}
    \begin{equation}\label{eq:taylor_2}
        \begin{split}
            &\frac{1}{Z^{(M) 2} Z^{(M') 2}}\Big(\sum_{\substack{w^{[1]},w^{[2]} \in \calW^{(M)}\\w'^{[1]},w'^{[2]} \in \calW^{(M')}}} f_2(\v_{w^{[1]}}^\star+\v_{w^{[2]}}^\star, \v_{w'^{[1]}}^\star+\v_{w'^{[2]}}^\star,\b_{w^{[1]}}+\b_{w^{[2]}}, \b_{w'^{[1]}}+\b_{w'^{[2]}})-Z^{(M) 2} Z^{(M') 2}\Big)\\
            & = \sum_{\substack{w^{[1]},w^{[2]} \in \calW^{(M)}\\w'^{[1]},w'^{[2]} \in \calW^{(M')}}} \frac{\exp(A_w^{(4)}/2)}{Z^{(M) 2} Z^{(M') 2}}\cdot\Big( \exp(B_w^{(4)}) f_3(\v_{w^{[1]}}^\star+\v_{w^{[2]}}^\star, \v_{w'^{[1]}}^\star+\v_{w'^{[2]}}^\star, \b_{w^{[1]}}+\b_{w^{[2]}}, \b_{w'^{[1]}}+\b_{w'^{[2]}}) - 1\Big).
        \end{split}
    \end{equation}
    \end{small}\noindent
    Again by taking Taylor expansion, we have that
    \begin{small}
    \begin{align*}
        &\frac{1}{Z^{(M) 2} Z^{(M') 2}}\Big(\sum_{\substack{w^{[1]},w^{[2]} \in \calW^{(M)}\\w'^{[1]},w'^{[2]} \in \calW^{(M')}}} f_2(\v_{w^{[1]}}^\star+\v_{w^{[2]}}^\star, \v_{w'^{[1]}}^\star+\v_{w'^{[2]}}^\star, \b_{w^{[1]}}+\b_{w^{[2]}}, \b_{w'^{[1]}}+\b_{w'^{[2]}})-Z^{(M) 2} Z^{(M') 2}\Big)\\
        &= \sum_{\substack{w^{[1]},w^{[2]} \in \calW^{(M)}\\w'^{[1]},w'^{[2]} \in \calW^{(M')}}} \frac{\exp(A_w^{(4)}/2)}{Z^{(M) 2} Z^{(M') 2}}\cdot (B_w^{(4)} + \log f_3(\v_{w^{[1]}}^\star+\v_{w^{[2]}}^\star, \v_{w'^{[1]}}^\star+\v_{w'^{[2]}}^\star,\b_{w^{[1]}}+\b_{w^{[2]}}, \b_{w'^{[1]}}+\b_{w'^{[2]}}) \Big)\\
        &\quad+ O\qty(\frac{\|\Vb_1\|_{2,\infty}^4 \vee \|\Vb_2\|_{2,\infty}^4}{p^2} + \max_{w \in \calW} \sigma_{w,w}^2)\\
           &= \frac{1}{p} \bar \bv^{(M) \top} \bar \bv^{(M)} + \frac{1}{p} \bar \bv^{(M') \top} \bar \bv^{(M')} + \frac{2}{p} \v_w^{\star \top} \bar \bv^{(M)} + \frac{2}{p} \v_{w'}^{\star \top} \bar \bv^{(M')} + \frac{2}{p} \v_w^{\star \top} \bar \bv^{(M')} + \frac{2}{p} \v_{w'}^{\star \top} \bar \bv^{(M)} + \frac{4}{p} \bar \bv^{(M) \top} \bar \bv^{(M')}\\
        &\quad+ \bar \sigma^{(M, M)} + \bar \sigma^{(M', M')} + 2 \bar \sigma_w^{(M)} + 2 \bar \sigma_{w'}^{(M')} + 2 \bar \sigma_w^{(M')} + 2 \bar \sigma_{w'}^{(M)} + 4 \bar \sigma^{(M, M')} + O\qty(\frac{p^2}{(d_M \wedge d_{M'})^2}),
    \end{align*}
    \end{small}\noindent
    where we used Assumptions~\ref{asm: signal covariance} and \ref{asm: noise covariance}.
    Combining \eqref{eq:pww'_leading_term} ,\eqref{eq:taylor_1} and \eqref{eq:taylor_2}, we have
    \begin{equation}\label{bound_pwwu}
        \begin{split}
            &\biggl|\log p_{w, w'}^{(M,M')} - \biggl(\frac{\|\v_w^\star\|^2 + \|\v_{w'}^\star\|^2 + 2\v_w^{\star \top}\v_{w'}^\star}{2p} + \frac{\|\b_w\|^2 + \|\b_{w'}\|^2 + 2\b_w^\top\b_{w'}}{2} - \log Z^{(M)} - \log Z^{(M')}\\
            &\quad- 3 \biggl(\frac{1}{p} \v_w^{\star \top} \bar \bv^{(M)} + \frac{1}{p} \v_{w'}^{\star \top} \bar \bv^{(M')} + \frac{1}{p} \v_w^{\star \top} \bar \bv^{(M')} + \frac{1}{p} \v_{w'}^{\star \top} \bar \bv^{(M)} + \frac{1}{p} \bar \bv^{(M) \top} \bar \bv^{(M')}\\
            &\quad\quad+ \bar\sigma_w^{(M)} + \bar\sigma_{w'}^{(M')} + \bar\sigma_{w'}^{(M)} + \bar\sigma_{w}^{(M')} + \bar\sigma^{(M, M')}\biggr)\\
            &\quad+ \frac{1}{p} \bar \bv^{(M) \top} \bar \bv^{(M)} + \frac{1}{p} \bar \bv^{(M') \top} \bar \bv^{(M')} + \frac{2}{p} \v_w^{\star \top} \bar \bv^{(M)} + \frac{2}{p} \v_{w'}^{\star \top} \bar \bv^{(M')} + \frac{2}{p} \v_w^{\star \top} \bar \bv^{(M')} + \frac{2}{p} \v_{w'}^{\star \top} \bar \bv^{(M)} + \frac{4}{p} \bar \bv^{(M) \top} \bar \bv^{(M')}\\
            &\quad+ \bar \sigma^{(M, M)} + \bar \sigma^{(M', M')} + 2 \bar \sigma_w^{(M)} + 2 \bar \sigma_{w'}^{(M')} + 2 \bar \sigma_w^{(M')} + 2 \bar \sigma_{w'}^{(M)} + 4 \bar \sigma^{(M, M')}\biggr)\biggr| \\
            &\leq 56\varepsilon'^3 + O\qty(\frac{p^2}{(d_M \wedge d_{M'})^2}).
    \end{split}
    \end{equation}
    Here the last inequality uses Assumption~\ref{asm: noise covariance}. 
    After simplification, we obtain
    \begin{equation}
        \begin{split}
            &\biggl|\log p_{w, w'}^{(M,M')} - \biggl(\frac{\|\v_w^\star\|^2 + \|\v_{w'}^\star\|^2 + 2\v_w^{\star \top}\v_{w'}^\star}{2p} + \frac{\sigma_{w,w} + \sigma_{w',w'} + 2\sigma_{w,w'}}{2} - \log Z^{(M)} - \log Z^{(M')}\\\
            &\quad+ \frac{1}{p} \bar \bv^{(M) \top} \bar \bv^{(M)} + \frac{1}{p} \bar \bv^{(M') \top} \bar \bv^{(M')} - \frac{1}{p} \v_w^{\star \top} \bar \bv^{(M)} - \frac{1}{p} \v_{w'}^{\star \top} \bar \bv^{(M')} - \frac{1}{p} \v_w^{\star \top} \bar \bv^{(M')} - \frac{1}{p} \v_{w'}^{\star \top} \bar \bv^{(M)} + \frac{1}{p} \bar \bv^{(M) \top} \bar \bv^{(M')}\\
            &\quad+ \bar \sigma^{(M, M)} + \bar \sigma^{(M', M')} - \bar \sigma_w^{(M)} - \bar \sigma_{w'}^{(M')} - \bar \sigma_w^{(M')} - \bar \sigma_{w'}^{(M)} + \bar \sigma^{(M, M')}\biggr)\biggr|\\
            &\quad\leq 56\varepsilon'^3 + O\qty(\frac{p^2}{(d_1 \wedge d_2)^2})\\
            &\quad\lesssim \frac{p^3}{(d_M \wedge d_{M'})^3} \tau^3 + \frac{p^2}{(d_1 \wedge d_2)^2} \lesssim \frac{p^2}{(d_M \wedge d_{M'})^2} \tau,\label{eq: e4 final}
    \end{split}
    \end{equation}
    where we used $d_1 \wedge d_2 \gg p \tau^2$ from Assumption~\ref{asm: regime}.
    Note that the last inequality in \eqref{eq: e4 final} does not depend on $w$ and $w'$.
    This concludes the proof.
\end{proof}

\begin{lem}\label{lem: PMI decomposition in max norm}
    Suppose that Assumptions~\ref{asm: signal covariance}, \ref{asm: regime} and \ref{asm: noise covariance} hold.
    Then, for $M, M' \in \{1, 2\}$, 
    \begin{align}
        &\max_{w\in\calW^{(M)}, w'\in\calW^{(M')}} \Biggl|\log\frac{p_{w,w'}^{(M,M')}}{p_w^{(M)} p_{w'}^{(M')}}\nonumber\\
        &\quad- \biggl(\frac{\v_w^{\star \top}\v_{w'}^\star}{p} - \frac{1}{p} \v_w^{\star \top} \bar \bv^{(M')} - \frac{1}{p} \v_{w'}^{\star \top} \bar \bv^{(M)} + \frac{1}{p} \bar \bv^{(M) \top} \bar \bv^{(M')} + \sigma_{w,w'} - \bar \sigma_w^{(M')} - \bar \sigma_{w'}^{(M)} + \bar \sigma^{(M, M')}\biggr)\Biggr|\nonumber\\
        &\lesssim \frac{p^2 \tau}{(d_M \wedge d_{M'})^2},\label{eq: PMI decomposition 1}
    \end{align}
    and
    \begin{align}
        &\max_{w\in\calW^{(M)}, w'\in\calW^{(M')}} \Biggl|\frac{p_{w,w'}^{(M,M')}}{p_w^{(M)} p_{w'}^{(M')}} - 1\nonumber\\
        &\quad- \biggl(\frac{\v_w^{\star \top}\v_{w'}^\star}{p} - \frac{1}{p} \v_w^{\star \top} \bar \bv^{(M')} - \frac{1}{p} \v_{w'}^{\star \top} \bar \bv^{(M)} + \frac{1}{p} \bar \bv^{(M) \top} \bar \bv^{(M')} + \sigma_{w,w'} - \bar \sigma_w^{(M')} - \bar \sigma_{w'}^{(M)} + \bar \sigma^{(M, M')}\biggr)\Biggr|\nonumber\\
        &\lesssim \frac{p^2 \tau}{(d_M \wedge d_{M'})^2}.\label{eq: PMI decomposition 2}
    \end{align}
\end{lem}

\begin{proof}
    The first claim directly follows from Lemmas~\ref{lem: A4} and \ref{lem: A5}.
    Note that
    \begin{align*}
        &\max_{w\in\calW^{(M)}, w'\in\calW^{(M')}} \abs{\frac{\v_w^{\star \top}\v_{w'}^\star}{p} - \frac{1}{p} \v_w^{\star \top} \bar \bv^{(M')} - \frac{1}{p} \v_{w'}^{\star \top} \bar \bv^{(M)} + \frac{1}{p} \bar \bv^{(M) \top} \bar \bv^{(M')} + \sigma_{w,w'} - \bar \sigma_w^{(M')} - \bar \sigma_{w'}^{(M)} + \bar \sigma^{(M, M')}}\\
        &\quad\lesssim \begin{cases}
            \frac{1}{p} \|\Vb_M\|_{2,\infty}^2 + \max_{w \in \calW^{(M)}} \sigma_{w,w} & \text{ if $M = M'$},\\
            \frac{1}{p} \|\Vb_M\|_{2,\infty} \|\Vb_{M'}\|_{2,\infty} & \text{ if $M \neq M'$},\\
        \end{cases}\\
        &\quad\lesssim \frac{p}{\sqrt{d_M d_{M'}}},
    \end{align*}
    where we used $\bar \sigma_w^{(M')} = \bar \sigma_{w'}^{(M)} = \bar \sigma^{(M, M')} = 0$ for $M \neq M'$ in the first inequality, and Assumptions~\ref{asm: signal covariance} and \ref{asm: noise covariance} in the last inequality.
    The second claim follows by the fact that for any $x > 0$ and $z \in \R$ satisfying $|\log x - z| \leq 1/2$ and $|z| \leq 1/2$,
    \begin{align*}
        \abs{x - 1 - z} \lesssim \abs{\log x - z} + z^2.
    \end{align*}
    To see this, observe that
    \begin{align*}
        \abs{x - 1 - z} &\leq \abs{x - 1 - z - \log x + z} + \abs{\log x - z}\\
        &= \abs{\exp(\log x) - 1 - \log x} + \abs{\log x - z}\\
        &\leq \log^2 x + \abs{\log x - z}\\
        &\leq 2z^2 + 2 \abs{\log x - z}^2 + \abs{\log x - z}\\
        &\leq 2z^2 + 2 \abs{\log x - z},
    \end{align*}
    where we used $|e^{x'} - 1 - x'| \leq {x'}^2$ for any $|x'| \leq 1$ and $|\log x| \leq 1$ by assumption.
\end{proof}

\begin{lem}[Modification to Lemma A.11 from \cite{xu2023inference}]\label{lem: A11}
    Suppose that Assumptions~\ref{asm: signal covariance}, \ref{asm: regime}, and \ref{asm: noise covariance} hold. 
    Fix $M, M' \in \{1, 2\}$. Then, for any $\varepsilon > 0$,
    \begin{align}
        \abs{\frac{d_M}{n S_4^{(M) 2}} \sum_{i=1}^n T_i^{(M)} (T_i^{(M)} - 1) (p_{i,w}^{(M)} - p_w^{(M)})} \geq \varepsilon + \exp(-\Omega(\tau^2)),\label{eq: pw concentration}\\
        \abs{\frac{d_M}{n S_4^{(M, M') 2}} \sum_{i=1}^n T_i^{(M)} T_i^{(M')} (p_{i,w}^{(M)} - p_w^{(M)})} \geq \varepsilon + \exp(-\Omega(\tau^2)),\label{eq: pw concentration 1.5}\\
        \abs{\frac{d_M d_{M'}}{n S_4^{(M,M') 2}} \sum_{i=1}^n T_i^{(M)} T_i^{(M')} (p_{i,i,w,w'}^{(M,M')} - p_{w,w'}^{(M,M')})} \geq \varepsilon + \exp(-\Omega(\tau^2)),\label{eq: pww concentration 2}\\
        \abs{\frac{d_M^2}{n S_4^{(M) 2}} \sum_{i=1}^n T_i^{(M)} (T_i^{(M)} - 1) (p_{i,i,w,w'}^{(M,M)} - p_{w,w'}^{(M,M)})} \geq \varepsilon + \exp(-\Omega(\tau^2)),\label{eq: pww concentration 2.5}\\
        \abs{\frac{d_M d_{M'}}{n(n-1) S_2^{(M)} S_2^{(M')}} \sum_{i,j:i\neq j} T_i^{(M)} T_j^{(M')} (p_{i,j,w,w'}^{(M,M')} - p_w^{(M)} p_{w'}^{(M')})} \geq \varepsilon + \exp(-\Omega(\tau^2)),\label{eq: pww concentration 3}
    \end{align}
    hold with probability $\exp(-\Omega(n \varepsilon^2 \wedge \tau^2))$.
\end{lem}

\begin{proof}
    Denote an event 
    $$
        \mathcal{F}_i = \{\|\bc_i\| \leq (1/2) \tau\} \cap \{\max_{w \in \calW^{(M)}} |\be_w^\top \beps_i| \leq (1/2) \sqrt{\max_{w \in \calW} \sigma_{w,w}} \tau\},
    $$
    then $\P(\mathcal{F}_i) = 1 - \exp(-\Omega(\tau^2))$ by Gaussian concentration inequality. 
    Define
    \begin{align*}
        f_{i,w}^{(M)} \triangleq \frac{\exp(\v_w^{\star \top} \bc_i + \be_w^\top \beps_i)}{\sum_{w' \in \calW^{(M)}} \exp(\v_{w'}^{\star \top} \bc_i + \be_{w'}^\top \beps_i)} I\{\mathcal{F}_i\}.
    \end{align*}
    Note that $0 \leq f_{i,w}^{(M)} \leq 1$. Furthermore, since
    \begin{align*}
        \max_{w \in \calW^{(M)}} \abs{\v_w^{\star \top} \bc_i + \be_w^\top \beps_i} \lesssim \sqrt{\frac{p}{d_M}}
    \end{align*}
    holds on the event $\mathcal{F}_i$ by Assumptions~\ref{asm: signal covariance} and \ref{asm: noise covariance}, we have
    \begin{align*}
        0 \leq f_{i,w}^{(M)} = \frac{1 + O(\sqrt{p/d_M} \tau)}{d + d O(\sqrt{p/d_M} \tau)} &= \frac{1}{d_M} + O\qty(\frac{p^{1/2}}{d_M^{3/2}} \tau) = O\qty(\frac{1}{d_M}),
    \end{align*}
    where we used $d_1 \wedge d_2 \gg p \tau^2$ from Assumption~\ref{asm: regime}.
    Using Hoeffding's inequality combined with $0 \leq f_{i,w}^{(M)} \leq 1$, we obtain
    \begin{align*}
        \P\qty(\abs{\sum_{i=1}^n T_i^{(M)} (T_i^{(M)} - 1) (f_{i,w}^{(M)} - \E[f_{1,w}^{(M)}])} \geq \varepsilon) \leq 2\exp(-\frac{2 d_M^2 \varepsilon^2}{n S_4^{(M) 4}}),
    \end{align*}
    where $S_q^{(M)} \triangleq ((1/n) \sum_i T_i^{(M) q})^{1/q}$.
    Also note that 
    \begin{align*}
        \P(\max_{i \in [n]} |p_{i,w}^{(M)} - f_{i,w}^{(M)}| > 0) &\leq \sum_{i \in [n]} \P(\mathcal{F}_i^c) = n \exp(-\Omega(\tau^2)) = \exp(-\Omega(\tau^2)),\\
        |\sum_{i=1}^n T_i^{(M)} (T_i^{(M)} - 1) (\E[p_{1,w}^{(M)}] - \E[f_{1,w}^{(M)}])| &\leq S_2^{(M) 2} \E[I\{\mathcal{F}_1^c\}] = S_4^{(M) 2} \exp(-\Omega(\tau^2)).
    \end{align*}
    This gives
    \begin{align*}
        &\P\qty(\abs{\sum_{i=1}^n T_i^{(M)} (T_i^{(M)} - 1) (p_{i,w}^{(M)} - \E[p_{1,w}^{(M)}])} \geq \varepsilon + S_4^{(M) 2} \exp(-\Omega(\tau^2)))\\
        &\quad\leq 2\exp(-\frac{2 d_M^2 \varepsilon^2}{n S_4^{(M) 4}}) + \exp(-\Omega(\tau^2)),
    \end{align*}
    The claim \eqref{eq: pw concentration} holds by setting $\epsilon \leftarrow \epsilon n S_4^{(M) 2}/d_M$ and by the fact that $(d_M / n) \exp(-\Omega(\tau^2)) = \exp(-\Omega(\tau^2))$.
    We can similarly obtain \eqref{eq: pw concentration 1.5}, \eqref{eq: pww concentration 2} and \eqref{eq: pww concentration 2.5}.
    To prove \eqref{eq: pww concentration 3},
    the bounded difference inequality gives that
    \begin{align*}
        \P\qty(\abs{\sum_{i,j:i\neq j} T_i^{(M)} T_j^{(M')} (f_{i,w}^{(M)} f_{j,w'}^{(M')} - \E[f_{i,w}^{(M)}] \E[f_{i,w'}^{(M')}])} \geq \varepsilon)
        &\leq 2\exp(-\frac{2 d_M^2 d_{M'}^2 \varepsilon^2}{4 n^3 S_2^{(M) 2} S_2^{(M') 2}}).
    \end{align*}
    Since $p_{i,j,w,w'}^{(M,M')} = p_{i,w}^{(M)} p_{j,w'}^{(M')} = f_{i,w}^{(M)} f_{j,w'}^{(M')}$ holds with high probability, setting $\epsilon \leftarrow \epsilon n(n-1) S_2^{(M)} S_2^{(M')}/(d_M d_{M'})$ with a similar argument as above yields \eqref{eq: pww concentration 3}.
\end{proof}

\begin{lem}[Modification to Lemma A.7 from \cite{xu2023inference}]\label{lem: A7}
    Suppose that Assumptions~\ref{asm: signal covariance}, \ref{asm: regime}, \ref{asm: noise covariance} and \ref{asm: T moments} hold.
    Fix $M \in \{1, 2\}$.
    Then,
    \begin{align*}
        \P\qty(\max_{w \in \calW^{(M)}} \abs{\frac{\sum_{i=1}^{n} T_i^{(M)} (T_i^{(M)} - 1) p_{i,w}^{(M)}}{\sum_{i=1}^{n} T_i^{(M)} (T_i^{(M)} - 1) p_w^{(M)}}-1} \geq \frac{\tau}{\sqrt{n}}) &= \exp(-\Omega(\tau^2))\\
        \P\qty(\max_{w \in \calW^{(M)}} \abs{\frac{\sum_{i=1}^{n} T_i^{(M)} T_i^{(M')} p_{i,w}^{(M)}}{\sum_{i=1}^{n} T_i^{(M)} T_i^{(M')} p_w^{(M)}}-1} \geq \frac{\tau}{\sqrt{n}}) &= \exp(-\Omega(\tau^2)).
    \end{align*}
\end{lem}

\begin{proof}
    Fix $w \in \calW^{(M)}$.
    From Lemma~\ref{lem: A11}, we have
    \begin{align*}
         \P\qty(\abs{\frac{d_M}{n S_4^{(M) 2}} \sum_{i=1}^n T_i^{(M)} (T_i^{(M)}-1) (p_{i,w}^{(M)} - p_w^{(M)})} \geq \varepsilon + \exp(-\Omega(\tau^2))) &= \exp(- \Omega(n \varepsilon^2 \wedge \tau^2)).
    \end{align*}
    By Lemma~\ref{lem: A4}, we have that $\min_{w \in \calW^{(M)}} p_w^{(M)} \gtrsim 1/d_M$.
    Thus, setting $\varepsilon = (1/2) C^{-1} n^{-1/2} p_w^{(M)} d_M \tau$ (which is $\gg \exp(-\Omega(\tau^2))$ for a constant $C > 0$ to be chosen below) gives
    \begin{align*}
        &\P\qty(\abs{\frac{d_M}{n S_4^{(M) 2}} \sum_{i=1}^n T_i^{(M)} (T_i^{(M)} - 1) p_{i,w}^{(M)} - \frac{d_M}{n S_4^{(M) 2}} p_w^{(M)} \sum_{i=1}^n T_i^{(M)} (T_i^{(M)} - 1)} \geq \frac{p_w^{(M)} d_M \tau}{C \sqrt{n}})\\
        &= \P\qty(\abs{\frac{\sum_{i=1}^n T_i^{(M)} (T_i^{(M)} - 1) p_{i,w}^{(M)}}{\sum_{i=1}^n T_i^{(M)} (T_i^{(M)} - 1) p_w^{(M)}} - 1} \geq \frac{S_4^{(M) 2} \tau}{C \sqrt{n} (S_2^{(M) 2} - S_1^{(M)})})\\
        &= \exp(- \Omega(\tau^2)).
    \end{align*}
    Then, a union bound argument gives
    \begin{align*}
         \P\qty(\max_{w \in \calW^{(M)}} \abs{\frac{\sum_{i=1}^n T_i^{(M)} (T_i^{(M)} - 1) p_{i,w}^{(M)}}{\sum_{i=1}^n T_i^{(M)} (T_i^{(M)} - 1) p_w^{(M)}} - 1} \geq \frac{S_4^{(M) 2} \tau}{C \sqrt{n} (S_2^{(M) 2} - S_1^{(M)})}) &= d_M \exp(- \Omega(\tau^2))\\
         &= \exp(- \Omega(\tau^2)).
    \end{align*}
    From Assumption~\ref{asm: T moments},
    $S_4^{(M) 2} \lesssim S_2^{(M) 2} \lesssim S_2^{(M) 2} - S_1^{(M)}$. Setting a sufficiently large $C > 0$ satisfying $S_4^{(M) 2} \leq C (S_2^{(M) 2} - S_1^{(M)})$ gives the first claim. The second claim follows similarly.
\end{proof}

\begin{lem}[Modification to Lemma A.8 from \cite{xu2023inference}]\label{lem: A8}
    Suppose that Assumptions~\ref{asm: signal covariance}, \ref{asm: regime}, \ref{asm: noise covariance} and \ref{asm: T moments} hold.
    Fix $M, M' \in \{1, 2\}$.
    Then,
    \begin{align*}
        \P\qty(\max_{w \in \calW^{(M)},w' \in \calW^{(M')}} \abs{\frac{\sum_{i=1}^{n} T_i^{(M)} T_i^{(M')} p_{i,i,w,w'}^{(M,M')}}{p_{w,w'}^{(M,M')} \sum_{i=1}^{n} T_i^{(M)} T_i^{(M')}}-1} \geq \frac{\tau}{\sqrt{n}}) &= \exp(-\Omega(\tau^2)),\\
        \P\qty(\max_{w ,w' \in \calW^{(M)}} \abs{\frac{\sum_{i=1}^{n} T_i^{(M)} (T_i^{(M)} - 1) p_{i,i,w,w'}^{(M,M)}}{p_{w,w'}^{(M,M)} \sum_{i=1}^{n} T_i^{(M)} (T_i^{(M)} - 1)}-1} \geq \frac{\tau}{\sqrt{n}}) &= \exp(-\Omega(\tau^2)),\\
        \P\qty(\max_{w \in \calW^{(M)},w' \in \calW^{(M')}} \abs{\frac{\sum_{i,j:i\neq j} T_i^{(M)} T_j^{(M')} p^{(M,M')}_{i,j,w,w'}}{p_w^{(M)} p_{w'}^{(M')} \sum_{i,j:i\neq j} T_i^{(M)} T_j^{(M')}}-1} \geq \frac{\tau}{\sqrt{n}}) &= \exp(-\Omega(\tau^2)).
    \end{align*}
\end{lem}
Note that $p_{i,j,w,w'}^{(1,2)} = p_{i,w}^{(1)} p_{j,w'}^{(2)}$ for $i \neq j$.

\begin{proof}
    Fix $w \in \calW^{(M)}$ and $w' \in \calW^{(M')}$. From Lemma~\ref{lem: A11}, we have
    \begin{align*}
         \P\qty(\abs{\frac{d_M d_{M'}}{n S_4^{(M,M') 2}} \sum_{i=1}^n T_i^{(M)} T_i^{(M')} (p_{i,i,w,w'}^{(M,M')} - p_{w,w'}^{(M,M')})} \geq \varepsilon + \exp(-\Omega(\tau^2))) &= \exp(- \Omega(n \varepsilon^2 \wedge \tau^2)).
    \end{align*}
    By Lemma~\ref{lem: A5}, we have that $\min_{w \in \calW^{(M)}, w' \in \calW^{(M')}} p_{w,w'}^{(M,M')} \gtrsim 1/(d_M d_{M'})$.
    From Assumption~\ref{asm: T moments}, we can take $C > 0$ such that $S_4^{(M,M')} \leq C S_2^{(M,M')}$ holds.
    Setting 
    $$
        \varepsilon = (1/2) C^{-2} n^{-1/2} p_{w,w'}^{(M,M')} d_M d_{M'} \tau
    $$ 
    gives $\varepsilon \gg \exp(-\Omega(\tau^2))$, and thus
    \begin{align*}
        &\P\qty(\abs{\frac{\sum_{i=1}^n T_i^{(M)} T_i^{(M')} p_{i,i,w,w'}^{(M,M')}}{p_{w,w'}^{(M,M')} \sum_{i=1}^n T_i^{(M)} T_i^{(M')} } - 1} \geq n^{-1/2} \tau)\\
        &\quad \leq \P\qty(\abs{\frac{\sum_{i=1}^n T_i^{(M)} T_i^{(M')} p_{i,i,w,w'}^{(M,M')}}{p_{w,w'}^{(M,M')} \sum_{i=1}^n T_i^{(M)} T_i^{(M')} } - 1} \geq \frac{S_4^{(M,M') 2}}{C^2 S_2^{(M,M') 2}} n^{-1/2} \tau) \leq 2 \exp(- \Omega(\tau^2)).
    \end{align*}
    A union bound argument gives the first claim.
    For the rest of claims, note that there exist constants $C', C'' > 0$ such that 
    \begin{align*}
        S_4^{(M) 2} &\leq C' (S_2^{(M) 2} - S_1^{(M)}),\\
        n(n-1) S_2^{(M)} S_2^{(M')} &\leq C'' (n^2 S_1^{(M)} S_1^{(M')} - n S_2^{(M,M') 2})
    \end{align*}
    hold by $S_4^{(M)} \lesssim S_2^{(M)} \lesssim S_1^{(M)}$, $S_1^{(M)} S_1^{(M')} \geq S_2^{(M,M') 2}$ and $1/n = o(1)$ from Assumptions~\ref{asm: regime} and \ref{asm: T moments}.
    A similar argument 
    combined with Lemma~\ref{lem: A11} gives
    \begin{small}
    \begin{align*}
        &\P\qty(\max_{w,w' \in \calW^{(M)}} \abs{\frac{\sum_{i\in [n]} T_i^{(M)} (T_i^{(M)} - 1) p^{(M,M)}_{i,i,w,w'}}{p_{w,w'}^{(M,M)} \sum_{i\in [n]} T_i^{(M)} (T_i^{(M)} - 1)}-1} \geq \frac{\tau}{\sqrt{n}})\\
        &\quad\leq \P\qty(\max_{w,w' \in \calW^{(M)}} \abs{\frac{\sum_{i\in [n]} T_i^{(M)} (T_i^{(M)} - 1) p^{(M,M)}_{i,i,w,w'}}{p_{w,w'}^{(M,M)} \sum_{i\in [n]} T_i^{(M)} (T_i^{(M)} - 1)}-1} \geq \frac{S_4^{(M) 2}}{C'(S_2^{(M) 2} - S_1^{(M)})} \frac{\tau}{\sqrt{n}})\\
        &\quad= \exp(-\Omega(\tau^2)),\\
        &\P\qty(\max_{w \in \calW^{(M)},w' \in \calW^{(M')}} \abs{\frac{\sum_{i,j:i\neq j} T_i^{(M)} T_j^{(M')} p^{(M,M')}_{i,j,w,w'}}{p_w^{(M)} p_{w'}^{(M')} \sum_{i,j:i\neq j} T_i^{(M)} T_j^{(M')}}-1} \geq \frac{\tau}{\sqrt{n}})\\
        &\quad\leq \P\qty(\max_{w \in \calW^{(M)},w' \in \calW^{(M')}} \abs{\frac{\sum_{i,j:i\neq j} T_i^{(M)} T_j^{(M')} p^{(M,M')}_{i,j,w,w'}}{p_w^{(M)} p_{w'}^{(M')} \sum_{i,j:i\neq j} T_i^{(M)} T_j^{(M')}}-1} \geq \frac{n(n-1) S_2^{(M)} S_2^{(M')}}{C''( n^2 S_1^{(M)} S_1^{(M')} - n S_2^{(M,M') 2})} \frac{\tau}{\sqrt{n}})\\
        &\quad= \exp(-\Omega(\tau^2)).
    \end{align*}
    \end{small}\noindent
    This concludes the proof.
\end{proof}

\begin{lem}[Modification to Lemma A.9 from \cite{xu2023inference}]\label{lem: A9}
    Suppose that Assumptions~\ref{asm: signal covariance}, \ref{asm: regime}, \ref{asm: noise covariance} and \ref{asm: T moments} hold.
    Fix $M \neq M' \in \{1, 2\}$. Then,
    \begin{align*}
        \P\qty(\max_{w \in \calW^{(M)}}\abs{\frac{\CC^{(M,M)}(w, \cdot)}{\sum_{i=1}^n T_i^{(M)}(T_i^{(M)}-1) p_{i,w}^{(M)}} - 1} \geq \frac{d_M \tau}{\sqrt{n} S_1^{(M) 1/2}}) &= \exp(-\Omega(\tau^2)),\\
        \P\qty(\max_{w \in \calW^{(M)}}\abs{\frac{\DD^{(M,M')}(w, \cdot)}{\sum_{i=1}^n T_i^{(M)} T_i^{(M')} p_{i,w}^{(M)}} - 1} \geq \frac{d_M \tau}{\sqrt{n} S_1^{(M) 1/2}}) &= \exp(-\Omega(\tau^2)).
    \end{align*}
\end{lem}

\begin{proof}
    Notice that given $(\bc_i, \beps_i)_{i \in [n]}$, we have $\sum_i T_i^{(M)}$ i.i.d. random variables $(X_{i,w}^{(M)}(t))_{i \in [n], t \in [T_i^{(M)}]}$.
    From bounded difference inequality, we have that
    \begin{small}
    \begin{align*}
        &\P\qty(\abs{\sum_{i=1}^{n} (T_i^{(M)} - 1) \sum_{t=1}^{T_i^{(M)}} X_{i,w}^{(M)}(t) - \sum_{i=1}^{n} T_i^{(M)} (T_i^{(M)} - 1) p_{i,w}^{(M)}} \geq \delta \sum_{i=1}^{n} T_i^{(M)} (T_i^{(M)} - 1) p_{i,w}^{(M)} \Biggm| (\bc_i, \beps_i)_{i=1}^n)\\
        &\leq 2 \exp(-\frac{2 \delta^{2} \qty(\sum_{i=1}^{n} T_i^{(M)} (T_i^{(M)} - 1) p_{i,w}^{(M)})^2}{\sum_{i \in [n]} T_i^{(M)} (T_i^{(M)} - 1)^2})
    \end{align*}
    \end{small}\noindent
    Define an event $\cF$ as
    \begin{equation*}
        \cF = \qty{\abs{\frac{\sum_{i=1}^n T_i^{(M)} (T_i^{(M)} - 1) p_{i,w}^{(M)}}{\sum_{i=1}^n T_i^{(M)} (T_i^{(M)} - 1) p_w^{(M)}} - 1} < \varepsilon },
    \end{equation*}
    where $\varepsilon = \tau / \sqrt{n}$.
    By Lemma~\ref{lem: A7}, $\P(\cF) = 1 - \exp(-\Omega(\tau^2))$. Therefore, we have
    \begin{align*}
        &\P\qty(\abs{\frac{\sum_{i=1}^{n} (T_i^{(M)} - 1) \sum_{t=1}^{T_i^{(M)}} X_{i,w}^{(M)}(t)}{\sum_{i=1}^{n} T_i^{(M)} (T_i^{(M)} - 1) p_{i,w}^{(M)}} - 1} \geq \delta)\\
        &\leq \P\qty(\abs{\frac{\sum_{i=1}^{n} (T_i^{(M)} - 1) \sum_{t=1}^{T_i^{(M)}} X_{i,w}^{(M)}(t)}{\sum_{i=1}^{n} T_i^{(M)} (T_i^{(M)} - 1) p_{i,w}^{(M)}} - 1} \geq \delta \Biggm| \cF) + \P\qty(\cF^c)\\
        &\leq 2\exp(-\frac{2 \delta^2 \qty(\sum_{i=1}^{n} T_i^{(M)} (T_i^{(M)} - 1) p_w^{(M)})^2 (1-\varepsilon)^2}{\sum_{i \in [n]} T_i^{(M)} (T_i^{(M)} - 1)^2}) + \exp(-\Omega(\tau^2)).
    \end{align*}
    Let $\delta = (S_3^{(M) 3} / n S_2^{(M) 4})^{1/2} d_M \tau$. 
    Since $n \gg \tau^2$, it follows that $\varepsilon = o(1)$.
    Using $\min_{w \in \calW^{(M)}} p_w^{(M)} \gtrsim 1/d_M$ from Lemma~\ref{lem: A4}, we have
    \begin{align*}
        \exp(-\frac{2 \delta^2 \qty(\sum_{i=1}^{n} T_i^{(M)} (T_i^{(M)} - 1) p_w^{(M)})^2 (1-\varepsilon)^2}{\sum_{i \in [n]} T_i^{(M)} (T_i^{(M)} - 1)^2}) &= \exp(-\Omega\qty(\frac{(S_2^{(M) 2} - S_1^{(M)})^2}{S_2^{(M) 4}} \tau^2))\\
        &= \exp(-\Omega(\tau^2)),
    \end{align*}
    where the last equality follows from Assumption~\ref{asm: T moments}.
    This yields
    \begin{align*}
        &\P\qty(\abs{\frac{\sum_{i=1}^{n} (T_i^{(M)} - 1) \sum_{t=1}^{T_i^{(M)}} X_{i,w}^{(M)}(t)}{\sum_{i=1}^{n} T_i^{(M)} (T_i^{(M)} - 1) p_{i,w}^{(M)}} - 1}\geq \frac{S_3^{(M) 3/2}}{S_2^{(M) 2}} \frac{d_M}{\sqrt{n}} \tau)\\
        &\quad\leq 2\exp(-\Omega(\tau^2)) + \exp(-\Omega(\tau^2)) = \exp(-\Omega(\tau^2)).
    \end{align*}
    By a union bound argument, we obtain
    \begin{align*}
        &\P\qty(\max_{w \in \calW^{(M)}} \abs{\frac{\sum_{i=1}^{n} (T_i^{(M)} - 1) \sum_{t=1}^{T_i^{(M)}} X_{i,w}^{(M)}(t)}{\sum_{i=1}^n T_i^{(M)} (T_i^{(M)} - 1) p_{i,w}^{(M)}} - 1} \geq \frac{S_3^{(M) 3/2}}{\sqrt{n} S_2^{(M) 2}} d_M \tau)\\
        &= d_M \exp(-\Omega(\tau^2)) = \exp(-\Omega(\tau^2)).
    \end{align*}
    Note that $S_3^{(M) 3/2}/S_2^{(M) 2} \leq S_4^{(M) 3/2}/S_1^{(M) 2} \lesssim 1/S_1^{(M) 1/2}$ follows from Assumption~\ref{asm: T moments}.
    Furthermore,
    \begin{align*}
        \CC^{(M,M)}(w, \cdot) &= \sum_{i \in [n]} \sum_{w' \in \calW^{(M)}} \sum_{t \neq s \in [T_i^{(M)}]} \I\{w^{(M)}_{i,t} = w, w^{(M)}_{i,s} = w'\}\\
        &= \sum_{i \in [n]} \sum_{t \in [T_i^{(M)}]} \I\{w^{(M)}_{i,t} = w\} \sum_{s \in [T_i^{(M)}] \setminus\{t\}} \sum_{w' \in \calW^{(M)}} \I\{w^{(M)}_{i,s} = w'\}\\
        &= \sum_{i \in [n]}  (T_i^{(M)}-1) \sum_{t \in [T_i^{(M)}]} X_{i,w}^{(M)}(t).
    \end{align*}
    This gives the first claim. 
    
    To see the second claim, we again use the bounded difference inequality:
    \begin{align*}
        &\P\qty(\abs{\sum_{i=1}^{n} T_i^{(M')} \sum_{t=1}^{T_i^{(M)}} X_{i,w}^{(M)}(t) - \sum_{i=1}^{n} T_i^{(M)} T_i^{(M')} p_{i,w}^{(M)}} \geq \delta \sum_{i=1}^{n} T_i^{(M)} T_i^{(M')} p_{i,w}^{(M)} \Biggm| (\bc_i, \beps_i)_{i=1}^n)\\
        &\leq 2 \exp(-\frac{2 \delta^{2} \qty(\sum_{i=1}^{n} T_i^{(M)} T_i^{(M')} p_{i,w}^{(M)})^2}{\sum_{i \in [n]} T_i^{(M)} T_i^{(M') 2}})
    \end{align*}    
    Define an event $\cF$ as
    \begin{equation*}
        \cF = \qty{\abs{\frac{\sum_{i=1}^n T_i^{(M)} T_i^{(M')} p_{i,w}^{(M)}}{\sum_{i=1}^n T_i^{(M)} T_i^{(M')} p_w^{(M)}} - 1} < \varepsilon },
    \end{equation*}
    where $\varepsilon = \tau / \sqrt{n}$.
    By Lemma~\ref{lem: A7}, $\P(\cF) = 1 - \exp(-\Omega(\tau^2))$.
    Therefore, we have
    \begin{align*}
        &\P\qty(\abs{\frac{\sum_{i=1}^{n} T_i^{(M')} \sum_{t=1}^{T_i^{(M)}} X_{i,w}^{(M)}(t)}{\sum_{i=1}^{n} T_i^{(M)} T_i^{(M')} p_{i,w}^{(M)}} - 1} \geq \delta)\\
        &\leq \P\qty(\abs{\frac{\sum_{i=1}^{n} T_i^{(M')} \sum_{t=1}^{T_i^{(M)}} X_{i,w}^{(M)}(t)}{\sum_{i=1}^{n} T_i^{(M)} T_i^{(M')} p_{i,w}^{(M)}} - 1} \geq \delta \Biggm| \cF) + \P\qty(\cF^c)\\
        &\leq 2\exp(-\frac{2 \delta^2 \qty(\sum_{i=1}^{n} T_i^{(M)} T_i^{(M')} p_w^{(M)})^2 (1-\varepsilon)^2}{\sum_{i \in [n]} T_i^{(M)} T_i^{(M') 2}}) + \exp(-\Omega(\tau^2)).
    \end{align*}
    Let $\delta = (n^{-1} \sum_i T_i^{(M)} T_i^{(M') 2} / n S_2^{(M,M') 4})^{1/2} d_M \tau$. 
    By a similar argument as above,
    \begin{align*}
        \exp(-\frac{2 \delta^2 \qty(\sum_{i=1}^{n} T_i^{(M)} T_i^{(M')} p_w^{(M)})^2 (1-\varepsilon)^2}{\sum_{i \in [n]} T_i^{(M)} T_i^{(M') 2}}) &= \exp(-\Omega(\tau^2)).
    \end{align*}
    By a union bound argument and the fact that $\DD^{(M,M')}(w,\cdot) = \sum_{i=1}^n T_i^{(M')} \sum_{t \in [T_i^{(M)}]} X_{i,w}^{(M)}(t)$, we obtain
    \begin{align*}
        &\P\qty(\max_{w \in \calW^{(M)}} \abs{\frac{\DD^{(M,M')}(w, \cdot)}{\sum_{i=1}^n T_i^{(M)} T_i^{(M')} p_{i,w}^{(M)}} - 1} \geq \frac{(n^{-1} \sum_{i \in [n]} T_i^{(M)} T_i^{(M') 2})^{1/2}}{\sqrt{n} S_2^{(M,M') 2}} d_M \tau)\\
        &= d_M \exp(-\Omega(\tau^2)) = \exp(-\Omega(\tau^2)).
    \end{align*}
    Note that
    \begin{align*}
        \frac{(n^{-1} \sum_{i \in [n]} T_i^{(M)} T_i^{(M') 2})^{1/2}}{S_2^{(1,2) 2}} \leq \frac{S_2^{(M) 1/2} S_4^{(M')}}{S_1^{(1)} S_1^{(2)}} \lesssim \frac{1}{S_1^{(M) 1/2}},
    \end{align*}
    where we used Cauchy-Schwarz inequality, $S_4^{(M)} \lesssim S_2^{(M)} \lesssim S_1^{(M)}$, and $S_1^{(1)} S_1^{(2)} \leq S_2^{(1,2) 2}$ from Assumption~\ref{asm: T moments}.
    This concludes the proof.
\end{proof}

\begin{lem}[Modification to Lemma A.10 from \cite{xu2023inference}]\label{lem: A10}
    Suppose that Assumptions~\ref{asm: signal covariance}, \ref{asm: regime}, \ref{asm: noise covariance} and \ref{asm: T moments} hold. Then,
    \begin{small}
    \begin{align}
        \P\qty(\max_{w, w' \in \calW^{(M)}} \abs{\frac{\CC^{(M,M)}(w, w')}{\sum_{i=1}^{n} T_i^{(M)}(T_i^{(M)}-1) p_{i,i,w,w'}^{(M,M)}} - 1} \geq \frac{d_M^2 \tau}{\sqrt{n} S_1^{(M) 1/2}}) &= \exp(-\Omega(\tau^2)),\label{eq: A10 1}\\
        \P\qty(\max_{w \in \calW^{(1)}, w' \in \calW^{(2)}} \abs{\frac{\DD^{(1,2)}(w, w')}{\sum_{i=1}^{n} T_i^{(1)} T_i^{(2)} p_{i,i,w,w'}^{(1,2)}} - 1} \geq \frac{d_1 d_2 \tau}{\sqrt{n} (S_1^{(1) 1/2} \wedge S_1^{(2) 1/2})}) &= \exp(-\Omega(\tau^2)),\label{eq: A10 2}\\
        \P\qty(\max_{w,w'\in\calW^{(M)}} \abs{\frac{\sum_{i,j: i \neq j} \DD^{(M,M)}_{i,j}(w,w')}{\sum_{i,j: i \neq j} T_i^{(M)} T_j^{(M)} p_{i,w}^{(M)} p_{j,w'}^{(M)}} - 1} \geq \frac{d_M^2 \tau}{\sqrt{n} S_1^{(M) 1/2}}) &= \exp(-\Omega(\tau^2)),\label{eq: A10 3}\\
        \P\qty(\max_{w\in\calW^{(1)},w'\in\calW^{(2)}} \abs{\frac{\sum_{i,j: i \neq j} \DD^{(1,2)}_{i,j}(w,w')}{\sum_{i,j: i \neq j} T_i^{(1)} T_j^{(2)} p_{i,w}^{(1)} p_{j,w'}^{(2)}} - 1} \geq \frac{d_1 d_2 \tau}{\sqrt{n} (S_1^{(1) 1/2} \wedge S_1^{(2) 1/2})}) &= \exp(-\Omega(\tau^2)).\label{eq: A10 4}
    \end{align}
    \end{small}\noindent
\end{lem}

\begin{proof}
    For $w, w' \in \calW$, we use the bounded difference inequality for $\sum_{i=1}^{n} \sum_{t\neq s \in [T_i^{(M)}]} X_{i,i,w,w'}^{(M,M)}(t, s)$ as a function of random variables $(X_{i,w}^{(M)}(t))_{i \in [n], t \in [T_i^{(M)}]}$ and $(X_{i,w'}^{(M)}(t))_{i \in [n], t \in [T_i^{(M)}]}$.
    \begin{align*}
        &\P\qty(\abs{\sum_{i=1}^{n} \sum_{t \neq s \in [T_i^{(M)}]} X_{i,i,w,w'}^{(M,M)}(t,s) - \sum_{i=1}^{n} T_i^{(M)} (T_i^{(M)} - 1) p_{i,i,w,w'}^{(M,M)}} \geq \delta \sum_{i=1}^{n} T_i^{(M)} (T_i^{(M)} - 1) p_{i,i,w,w'}^{(M)} \Biggm| (\bc_i, \beps_i)_{i=1}^n)\\
        &\leq 2 \exp(-\frac{2 \delta^{2} \qty(\sum_{i=1}^{n} T_i^{(M)} (T_i^{(M)} - 1) p_{i,i,w,w'}^{(M)})^2}{4 \sum_{i \in [n]} T_i^{(M)} (T_i^{(M)} - 1)^2}).
    \end{align*}
    By a similar argument as in the proof of Lemma~\ref{lem: A9} combined with Lemma~\ref{lem: A8}, 
    we obtain
    \begin{align*}
        \P\qty(\max_{w, w' \in \calW^{(M)}} \abs{\frac{\sum_{i=1}^{n} \sum_{t\neq s \in [T_i^{(M)}]} X_{i,i,w,w'}^{(M,M)}(t, s)}{\sum_{i=1}^n T_i^{(M)}(T_i^{(M)}-1) p_{i,i,w,w'}^{(M,M)}} - 1} \geq \frac{S_3^{(M) 3/2}}{\sqrt{n} S_2^{(M) 2}} d_M^2 \tau) = \exp(-\Omega(\tau^2)).
    \end{align*}
    Since $\CC_{i,i}^{(M,M)}(w, w') = \sum_{i=1}^{n} \sum_{t \neq s \in [T_i^{(M)}]} X_{i,i,w,w'}^{(M,M)}(t, s)$ and $S_3^{(M) 3/2} / S_2^{(M) 2} \lesssim 1/S_1^{(M) 1/2}$ from Assumption~\ref{asm: T moments}, this yields \eqref{eq: A10 1}.
    Similarly, using bounded difference inequality, we obtain
    \begin{align*}
        &\P\qty(\abs{\sum_{i=1}^{n} \sum_{t \in [T_i^{(1)}], s \in [T_i^{(2)}]} X_{i,i,w,w'}^{(1,2)}(t,s) - \sum_{i=1}^{n} T_i^{(1)} T_i^{(2)} p_{i,i,w,w'}^{(1,2)}} \geq \delta \sum_{i=1}^{n} T_i^{(1)} T_i^{(2)} p_{i,i,w,w'}^{(1,2)} \Biggm| (\bc_i, \beps_i)_{i=1}^n)\\
        &\leq 2 \exp(-\frac{2 \delta^{2} \qty(\sum_{i=1}^{n} T_i^{(1)} T_i^{(2)} p_{i,i,w,w'}^{(1,2)})^2}{\sum_{i \in [n]} (T_i^{(1) 2} T_i^{(2)} + T_i^{(1)} T_i^{(2) 2})}).
    \end{align*}
    From Cauchy-Schwarz inequality and Assumption~\ref{asm: T moments},
    \begin{align*}
        \frac{\sum_i (T_i^{(1) 2} T_i^{(2)} + T_i^{(1)} T_i^{(2) 2})}{(\sum_i T_i^{(1)} T_i^{(2)})^2}
        &\leq \frac{S_4^{(1,2) 2} (S_2^{(1)} + S_2^{(2)})}{S_2^{(1,2) 4}} \lesssim \frac{S_2^{(1,2) 2} (S_1^{(1)} + S_1^{(2)})}{S_2^{(1,2) 4}} \leq \frac{S_1^{(1)} + S_1^{(2)}}{S_1^{(1)} S_1^{(2)}}.
    \end{align*}
    By a similar argument as above, we obtain
    \begin{align*}
        &\P\qty(\max_{w \in \calW^{(1)}, w' \in \calW^{(2)}} \abs{\frac{\sum_{i=1}^{n} \sum_{t \in [T_i^{(1)}],s \in [T_i^{(2)}]} X_{i,i,w,w'}^{(1,2)}(t, s)}{\sum_{i=1}^n T_i^{(1)} T_i^{(2)} p_{i,i,w,w'}^{(1,2)}} - 1} \geq \frac{1}{2 \sqrt{n}} \frac{1}{(S_1^{(1) 1/2} \wedge S_1^{(2) 1/2})} d_1 d_2 \tau)\\
        &= \exp(-\Omega(\tau^2)).
    \end{align*}
    Since for $w \in \calW^{(1)}, w' \in \calW^{(2)}$, $\DD^{(1,2)}_{i,i}(w, w') = \sum_{t \in [T_i^{(1)}],s \in [T_i^{(2)}]} X_{i,i,w, w'}^{(1,2)}(t, s)$, this yields \eqref{eq: A10 2} using $S_1^{(1)} S_1^{(2)} / S_2^{(1,2) 2} \gg 1/n$ from Assumption~\ref{asm: T moments}.
    Furthermore, since $\sum_{i,j: i \neq j} \DD^{(M,M)}_{i,j}(w,w') = \sum_{i,j: i \neq j} \sum_{t \in [T_i^{(M)}], s \in [T_j^{(M)}]} X_{i,w}^{(M)}(t) X_{j,w'}^{(M)}(s)$, the bounded difference inequality gives that for any fixed $w, w' \in \calW^{(M)}$,
    \begin{align*}
        &\P\qty(\abs{\sum_{i,j: i \neq j} \DD^{(M,M)}_{i,j}(w,w') - \sum_{i,j: i \neq j} T_i^{(M)} T_j^{(M)} p_{i,w}^{(M)} p_{j,w'}^{(M)}} \geq \delta \sum_{i,j: i \neq j} T_i^{(M)} T_j^{(M)} p_{i,w}^{(M)} p_{j,w'}^{(M)} \Biggm| (\bc_i, \beps_i)_{i \in [n]} )\\
        &\leq 2\exp(- \frac{\delta^2 (\sum_{i,j: i \neq j} T_i^{(M)} T_j^{(M)} p_{i,w}^{(M)} p_{j,w'}^{(M)})^2}{n^3 S_1^{(M) 3}}).
    \end{align*}
    By a similar argument combined with $S_1^{(M) 2} / S_2^{(M) 2} \gg 1/n$ from Assumptions~\ref{asm: regime} and \ref{asm: T moments}, we obtain \eqref{eq: A10 3}.
    Finally, \eqref{eq: A10 4} follows by a similar argument combined with $S_1^{(1)} S_1^{(2)} / S_2^{(1,2) 2} \gg 1/n$ from Assumptions~\ref{asm: regime} and \ref{asm: T moments}.
\end{proof}

\begin{lem}\label{lem: rank p approx diff}
    Fix $q_1, q_2 > p$.
    For any $\Ab, \Ab' \in \R^{q_1 \times q_2}$, if $s_p(\Ab) > s_{p+1}(\Ab)$ and $(1/2) (s_p(\Ab) - s_{p+1}(\Ab)) \geq \|\Ab' - \Ab\|$ hold, then, 
    \begin{align*}
        \|\SVD_p(\Ab') - \SVD_p(\Ab)\|_\F &\lesssim \frac{\|\Ab\|}{s_p(\Ab) - s_{p+1}(\Ab)} \|\Ab' - \Ab\|_\F.
    \end{align*}
\end{lem}

\begin{proof}
    By triangle inequality,
    \begin{align*}
        \|\SVD_p(\Ab') - \SVD_p(\Ab)\|_\F &= \|\Ab' \Pb_p(\Ab') \Pb_p^\top(\Ab') - \Ab \Pb_p(\Ab) \Pb_p^\top(\Ab)\|_\F\\
        &\leq \|(\Ab' - \Ab) \Pb_p(\Ab') \Pb_p^\top(\Ab')\|_\F + \|\Ab (\Pb_p(\Ab') \Pb_p^\top(\Ab') - \Pb_p(\Ab) \Pb_p^\top(\Ab))\|_\F\\
        &\leq \|\Ab' - \Ab\|_\F + \|\Pb_p(\Ab') \Pb_p^\top(\Ab') - \Pb_p(\Ab) \Pb_p^\top(\Ab)\|_\F \|\Ab\|.
    \end{align*}
    Using Theorem 2.9 from \cite{chen2021spectral}, we have
    \begin{align*}
        \|\Pb_p(\Ab') \Pb_p^\top(\Ab') - \Pb_p(\Ab) \Pb_p^\top(\Ab)\|_\F &\leq \frac{\sqrt{2} \|\Ab' - \Ab\|_\F}{s_p(\Ab) - s_{p+1}(\Ab) - \|\Ab' - \Ab\|}\\
        &\leq \frac{2 \sqrt{2} \|\Ab' - \Ab\|_\F}{s_p(\Ab) - s_{p+1}(\Ab)},
    \end{align*}
    where the last inequality follows by assumption.
    Thus
    \begin{align*}
        \|\SVD_p(\Ab') - \SVD_p(\Ab)\|_\F &\lesssim \frac{\|\Ab\|}{s_p(\Ab) - s_{p+1}(\Ab)} \|\Ab' - \Ab\|_\F.
    \end{align*}
    This concludes the proof.
\end{proof}

\section{Extension of the CLAIME Algorithm to Multiple Modalities} \label{sec:claime_extension}

To extend CLAIME beyond two modalities, we generalize its PMI-based embedding framework. Suppose that there are $m$ modalities. Let $\widehat{\PMIbb}^{(M,M')}_\CLAIME$ denote the estimated cross-modal PMI matrix between modalities $M$ and $M'$, computed analogously to the two-modality case (see Proposition~\ref{prop: SVD}). Our goal is to learn modality-specific embeddings $\{\Vb_M\}_{M=1}^m$ by minimizing the following loss function:
\begin{equation}
\label{eq: CLAIME loss multi}
    \mathcal{L} \left(\{\Vb_M\}_{1 \leq M \leq m} \right) = \sum_{1 \leq M \neq M' \leq m} \left\|\widehat{\PMIbb}^{(M,M')}_\CLAIME -  \Vb_M \Vb_{M'}^\top \right\|_\F^2.
\end{equation}
We solve this problem using a scaled gradient descent approach similar to \cite{tong2021accelerating}. The update rule at iteration $t$ for modality $M$ is:
\[
    \Vb_M^{(t+1)} = \Vb_M^{(t)} - \eta \nabla_{\Vb_M} \mathcal{L}(\{\Vb_{M'}^{(t)}\}_{M'=1}^m) \left( \sum_{M' \neq M} \Vb_{M'}^{(t)\top} \Vb_{M'}^{(t)} \right)^{-1},
\]
where $\eta > 0$ is the learning rate. We initialize using the CL estimator and terminate when the loss change drops below 0.01.

\begin{algorithm}[h!]
\caption{CLAIME for $m$-Modality Embedding Learning}
\label{alg:claime_multi}
\begin{algorithmic}[1]
\State \textbf{Input:} Cross-modal PMI matrices $\{\widehat{\PMIbb}^{(M,M')}_\CLAIME\}_{1 \leq M \neq M' \leq m}$, rank $p$, learning rate $\eta$, and tolerance $\epsilon$
\State Initialize $\{\Vb_M^{(0)}\}_{M=1}^m$ using CL estimator
\Repeat
    \For{$M = 1$ to $m$}
        \State Compute gradient $\nabla_{\Vb_M} \mathcal{L}$
        \State Update:
        \[
            \Vb_M^{(t+1)} = \Vb_M^{(t)} - \eta \nabla_{\Vb_M} \mathcal{L} \left( \sum_{M' \neq M} \Vb_{M'}^{(t)\top} \Vb_{M'}^{(t)} \right)^{-1}
        \]
    \EndFor
\Until{loss change $< \epsilon$}
\State \textbf{Output:} $\{\widehat{\Vb}_M\}_{M=1}^m$
\end{algorithmic}
\end{algorithm}

\section{Additional Experiments}

\subsection{Additional Details on Real Data Experiment} \label{supp_subsec:llm_prompt}

This section first introduces the prompt used to query both GPT-4 and GPT-3.5 in Section~\ref{real} and then provides some detailed examples to show that compared to Concate and CL, CLAIME can better encode and preserve relations between these medical concepts and their associated features.

We adopted the following prompt to query both GPT-4 and GPT-3.5:

\begin{verbatim}
messages = [
    {"role": "system", "content": "Assume the role of a medical expert and evaluate
    the relatedness of the following two clinical concepts. This is for feature
    selection in a medical context, so the concepts are expected to have a degree
    of clinical or medical relationship. Please rate the relatedness on a scale
    of 0 to 1, where 0 indicates no relationship and 1 indicates a perfect match.
    Provide your rating as a single decimal number without any additional text."},
    {"role": "user", "content": f"{str1} & {str2}"},
]
\end{verbatim}

Here, ``str1'' and ``str2'' are placeholders for two clinical concepts being evaluated. We populated this template using a CSV file containing all concept pairs and submitted the resulting prompts to GPT-4 and GPT-3.5 for scoring. The output scores reflect the models’ assessments of clinical relatedness between each pair.

Some detailed examples from the results in Figure~\ref{fig:real} are given in Table~\ref{tab:real-1}. More specifically, leflunomide is a type of disease-modifying antirheumatic drug used to treat rheumatoid arthritis by reducing inflammation and permanent damage. Table~\ref{tab:real-1} shows that both Concate and CL fail to capture the close connection between leflunomide, RA, and DMARDs. Additionally, they overestimate similarity with unrelated drugs like hydrochlorothiazide, metformin, and citalopram. In the case of pneumococcal vaccine, it is a vaccine targeting Streptococcus pneumoniae infections and can potentially cause injection site reactions. It is related to protein antibodies as they all target the immune system \citep{von2014systemic}. Both Concate and CL struggle to capture this similarity and instead overestimate similarity with unrelated drugs like tramadol, cyanocobalamin, and citalopram, used for pain relief, vitamin B12 deficiency, and depressive disorders respectively.

\begin{table}[ht!]
\footnotesize
    \centering
    \resizebox{\columnwidth}{!}{%
    \begin{tabular}{ll|ccccc}
    \hline
        ID & Desc & Concate & CL & CLAIME & GPT3.5 & GPT4  \\ \hline
        \multicolumn{7}{c}{Relatedness or similarity with Leflunomide}  \\ \hline
        PheCode:714.1 & Rheumatoid arthritis & 0.35 & 0.35 & 0.80 & 0.9 & 0.9  \\ 
        PheCode:714.2 & Juvenile rheumatoid arthritis & 0.36 & 0.37 & 0.78 & 0.9 & 0.8  \\ 
        C0242708 & DMARDs & 0.36 & 0.36 & 0.81 & 0.9 & 1.0  \\
        RXNORM:5487 & Hydrochlorothiazide & 0.80 & 0.81 & 0.01 & 0.1 & 0.2  \\ 
        RXNORM:6809 & Metformin & 0.79 & 0.79 & 0.01 & 0.2 & 0.2  \\ 
        RXNORM:2556 & Citalopram & 0.78 & 0.78 & 0.05 & 0.1 & 0.1  \\ \hline
        \multicolumn{7}{c}{Relatedness or similarity with Pneumococcal vaccine }\\ \hline
        C0151735 & Injection site reaction NOS & 0.29 & 0.29 & 0.74 & 0.8 & 0.8  \\ 
        CCS:228 & Prophylactic vaccinations and inoculations & 0.29 & 0.29 & 0.64 & 0.9 & 1.0  \\
        C1445860 & Protein antibody & 0.04 & 0.04 & 0.61 & 0.6 & 0.7  \\ 
        RXNORM:10689 & Tramadol & 0.80 & 0.80 & 0.12 & 0.1 & 0.1  \\ 
        RXNORM:2556 & Citalopram & 0.75 & 0.75 & 0.15 & 0.1 & 0.1  \\ 
        RXNORM:11248 & Cyanocobalamin & 0.81 & 0.81 & 0.18 & 0.2 & 0.1 \\ \hline
    \end{tabular}}
    \caption{Cosine similarity between leflunomide, pneumococcal vaccine and some selected features.}
    \label{tab:real-1}
\end{table}

\subsection{Real-World Evaluation on Multimodal Biobank Dataset}
\label{supp:multimodal}

We evaluate the extended version of CLAIME (Algorithm~\ref{alg:claime_multi}) on a cohort of about $140,000$ patients from the Mass General Brigham (MGB) Biobank.  Unlike the setup in Section~\ref{real}, we treat different types of EHR codes as separate modalities to extend CLAIME beyond two modalities and assess its performance. The dataset contains $d = 117,719$ features across five distinct clinical modalities: $d_1 = 1,863$ PheCodes (diseases), $d_2 = 248$ CCS codes (procedures), $d_3 = 1,625$ RxNorm codes (medications), $d_4 = 4,291$ LOINC codes (labs), and $d_5 = 109,692$ CUIs (NLP-derived features).

We construct co-occurrence matrices using a one-month time window, where two codes are considered co-occurring if they appear for the same patient within one month. We then apply Algorithm~\ref{alg:claime_multi} with learning rate $\eta = 0.01$, terminating when the change in loss falls below $\epsilon = 0.01$. We compare the learned embeddings $\widehat{\Vb}_M$ with those from baseline methods including Concate, CL, and KPCA.

For evaluation, we use a curated set of semantically related code–CUI pairs from UMLS, as in Section~\ref{real}. In this extended experiment, we adopt a more challenging evaluation protocol: for each true pair, we sample $10$ negative control pairs that share the same semantic type but are unrelated. We compute cosine similarities within each set, rank the pairs, and report precision as the fraction of sets where the true pair ranks highest.

\begin{figure}[ht!]
\includegraphics[scale=0.25]{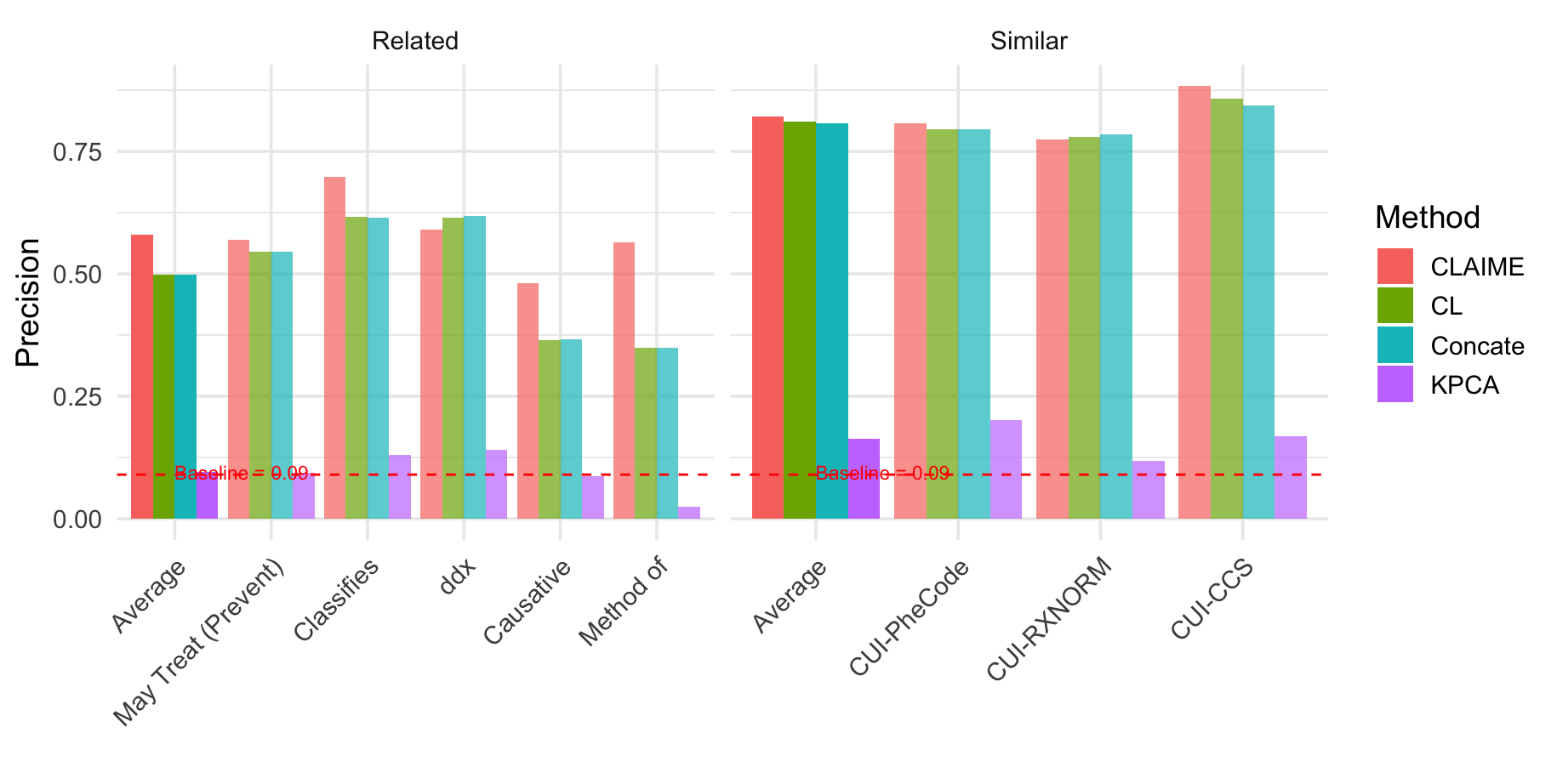}
\centering
\caption{Precision evaluation on the MGB Biobank dataset with five modalities: PheCode, CCS, RxNorm, LOINC, and CUI. For each semantically related (positive) code–CUI pair, we randomly sample $10$ unrelated (negative) pairs of the same semantic type. Precision is defined as the proportion of cases in which the positive pair has the highest cosine similarity among the $11$ total pairs. A random baseline achieves a precision of $0.09$.}
\label{fig:real_biobank}
\end{figure}

Figure~\ref{fig:real_biobank} summarizes the results. All methods perform better at capturing similarity than broader relatedness. KPCA performs poorly in this high-dimensional setting. CL and Concate perform similarly, while CLAIME consistently outperforms both methods, particularly in capturing nuanced relationships such as causality, classification hierarchies, and methodological associations.

\subsection{Additional Simulation Studies}

\subsubsection{Impact of Larger $d$} \label{supp_subsec:large_d}

In the following, we consider settings with larger $d$ to provide more empirical support for our real data analysis. Specifically, using the two data-generating processes detailed in the main text, we adopt the following setting. In the first experiment, we vary the sample sizes $n$ in $\{2\times 10^4,4\times 10^4,6\times 10^4,8\times 10^4,10^5\}$, while fixing $d=2d_1=2d_2=1000$ and $p=4$. In the second experiment, we vary $d$ in $\{200,400,600,800,1000\}$ and set $d_1=d_2=d/2$, while fixing $n=10^5$ and $p=4$. For the third experiment, we first fix $n=10^5$, $d=2d_1=2d_2=1000$ and $p=4$. Then, we vary $c$ equally spaced between $0.2$ and $1$ for Case 1, and we vary $\rho$ equally spaced between $0$ and $0.9$ for Case 2.  

\begin{figure}[h!]
\includegraphics[scale=0.52]{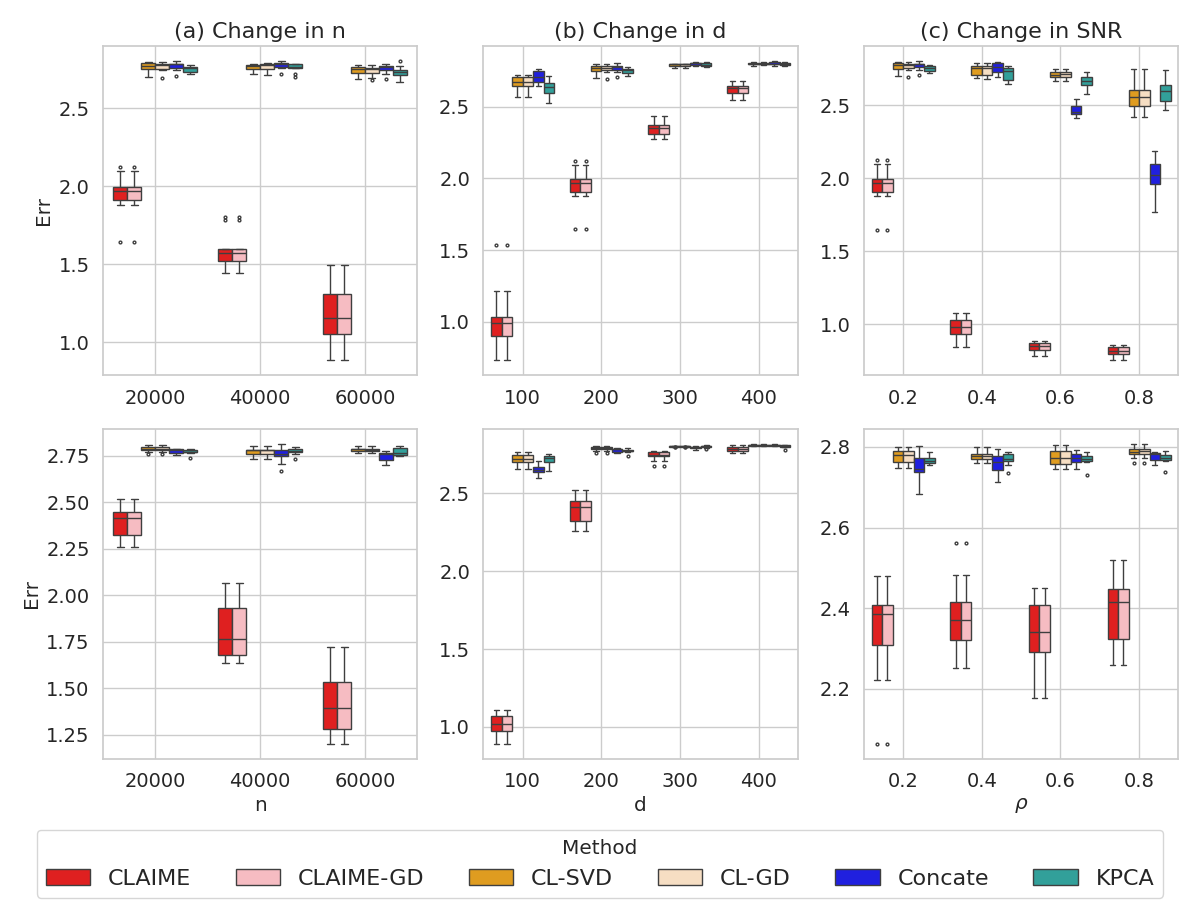}
\centering
\caption{In the top panel, the error metric is plotted against varying $n$, $d$ and $c$ for Case 1. In the bottom panel, the error metric is plotted against varying $n$, $d$ and $\rho$ for Case 2.}
\label{fig:simulation_larger_d}
\end{figure}

The error metrics $\text{Err}(\widehat{\mathbf{U}},\mathbf{U}^\star)$, averaged over 500 repetitions, are shown in Figure~\ref{fig:simulation_larger_d}. The curves for CL vs. CL-GD and CLAIME vs. CLAIME-GD closely align in Figures~\ref{fig:simulation_larger_d}(a)--(c), indicating that gradient-based methods on patient-level data perform comparably to SVD on summary-level PMI matrices. This aligns with Propositions~\ref{prop: SVD} and~\ref{prop: PCA CL}.

In Figures~\ref{fig:simulation_larger_d}(a) and (b), errors for CLAIME and CLAIME-GD decrease with increasing $n$ or decreasing $d$, consistent with the upper bounds in Theorems~\ref{thm: PMI decomposition}--\ref{thm: MMCL convergence}. In contrast, errors for Concate, CL, and CL-GD remain flat, as their lower bounds are independent of $n$ and $d$ (Theorem~\ref{thm: PCA DDPCA MMCL}). These trends mirror those observed in the small-$d$ setting discussed in Section~\ref{simulation} of the main paper.

Figure~\ref{fig:simulation_larger_d}(c, top) shows that all methods degrade as $c$ (i.e., SNR) decreases. Compared to the small-$d$ setting, CLAIME and CLAIME-GD are not generally superior in all signal regimes. Specifically, they are more robust only when $c \leq 0.5$ or when $\rho > 0.5$.

\subsubsection{Effect of Model Misspecification} \label{supp_subsec:model_misspecification}

\begin{figure}[h!]
\includegraphics[scale=0.52]{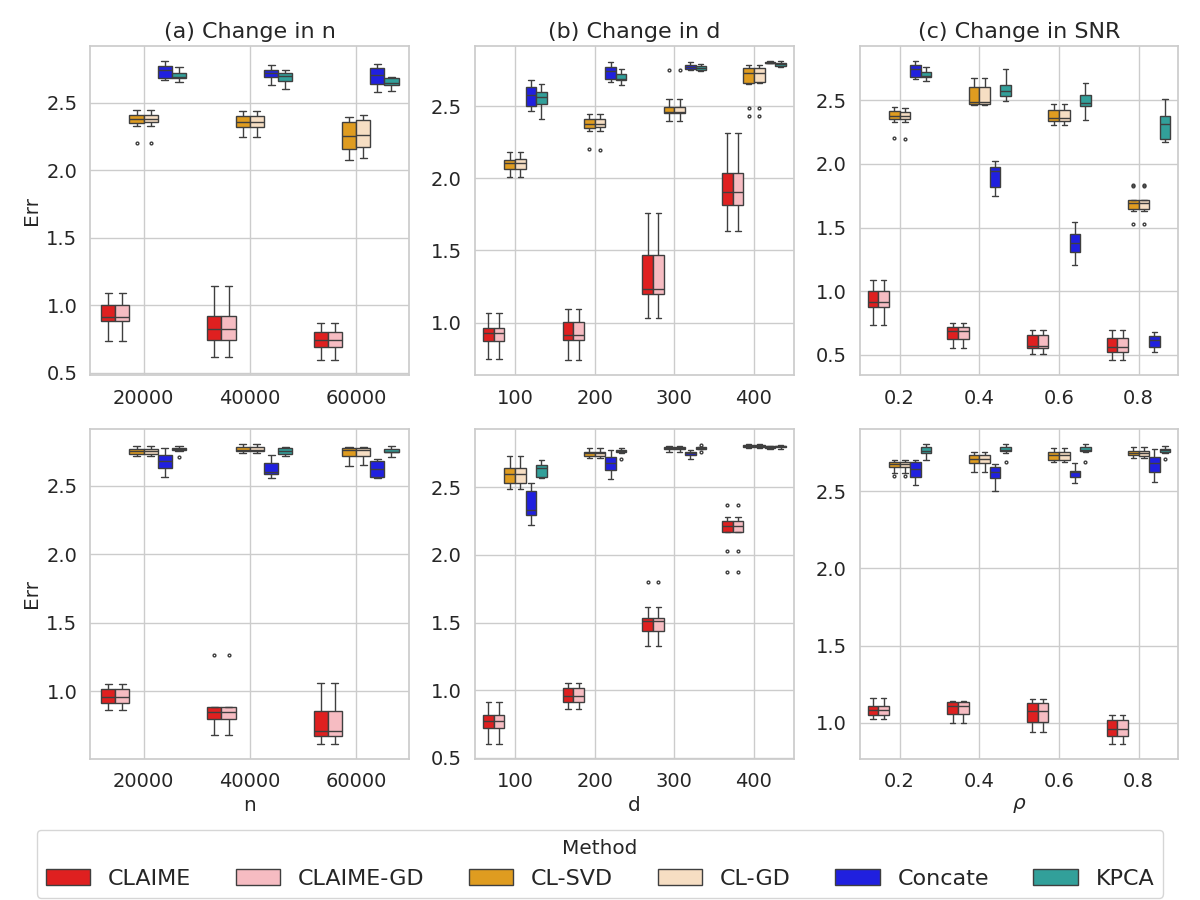}
\centering
\caption{In the top panel, the error metric is plotted against varying $n$, $d$ and $c$ for Case 1. In the bottom panel, the error metric is plotted against varying $n$, $d$ and $\rho$ for Case 2.}
\label{fig:simulation_misspecification}
\end{figure}

Although our theoretical results assume patient embeddings $\mathbf{c}_i \sim \mathcal{N}(0, \mathbf{I})$, we also tested the robustness of our method to distributional misspecification by generating $\mathbf{c}_i$ from a Student’s $t_3$ distribution, while keeping the other settings the same as in Section~\ref{simulation}. The results provided in Figure~\ref{fig:simulation_misspecification} show that our method maintains similar performance under this heavier-tailed distribution, further supporting its robustness.

\subsubsection{Debiasing Concate and CL} \label{supp_subsec:debias}

\begin{figure}[h!]
\includegraphics[scale=0.52]{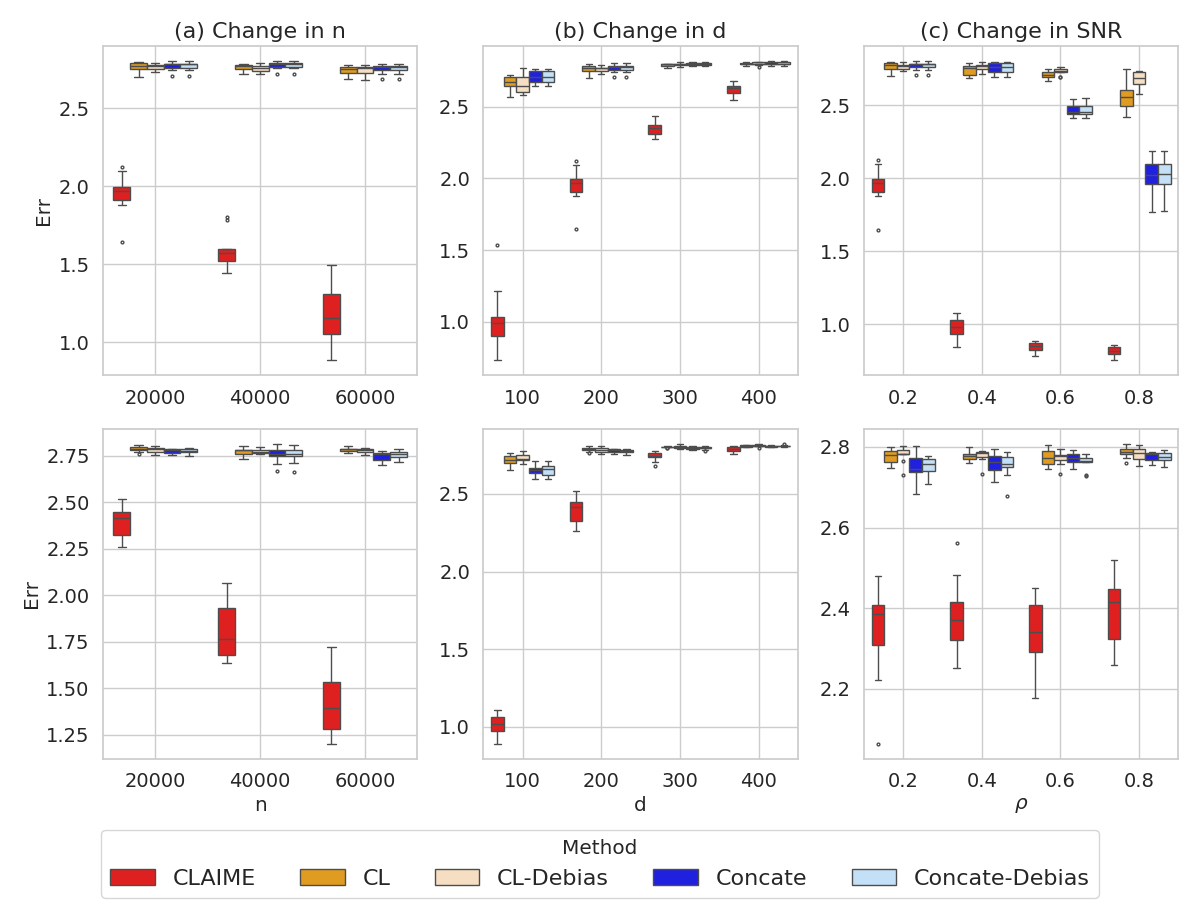}
\centering
\caption{In the top panel, the error metric is plotted against varying $n$, $d$ and $c$ for Case 1. In the bottom panel, the error metric is plotted against varying $n$, $d$ and $\rho$ for Case 2.}
\label{fig:simulation_bias}
\end{figure}

Since $\mathbf{B}_{\rm Concate}$ and part of $\mathbf{B}_{\rm CL}$ depend solely on $T_i^{(M)}$, we subtract the known bias to construct debiased versions of the Concate and CL estimators. We further include these two estimators into the first simulation experiment in Section~\ref{simulation} of the main paper and present their error metric $\text{Err}(\widehat{\mathbf{U}},\mathbf{U}^\star)$ along with that of CLAIME's. As shown in Figure~\ref{fig:simulation_bias}, the estimation errors for both debiased methods remain largely unchanged compared to their original counterparts. In contrast, CLAIME continues to outperform all other methods across settings, consistent with our theoretical analysis.

\clearpage
\newpage

\bibliographystyle{apalike}
\bibliography{ref}